\documentclass[10pt,twocolumn,letterpaper]{article}

\usepackage{cvpr}              
\usepackage{graphicx}
\usepackage{amsmath}
\usepackage{amssymb}
\usepackage{booktabs}
\usepackage{times}
\usepackage{epsfig}
\usepackage{graphics}
\usepackage{multirow}
\usepackage{booktabs}
\usepackage{caption}

\newcommand{\negvspace}{\vspace{-0.2cm}}
\usepackage[dvipsnames]{xcolor}

\definecolor{cvprblue}{rgb}{0.21,0.49,0.74}
\usepackage[pagebackref,breaklinks,colorlinks,citecolor=cvprblue]{hyperref}

\def\paperID{12527} 
\def\confName{CVPR}
\def\confYear{2024}

\title{Mind The Edge: Refining Depth Edges in Sparsely-Supervised Monocular Depth Estimation}
\author{Lior Talker$^1$~~~~~~~
Aviad Cohen$^1$~~~~~~~
Erez Yosef$^{1,2}$~~~~~~~
Alexandra Dana$^1$~~~~~~~ 
Michael Dinerstein$^1$ \\
~~~$^1$Samsung Israel R\&D Center, Tel Aviv, Israel~~~~~~~~~~~~~~~~~~~~~~~~$^2$Tel Aviv University, Israel \\
{\tt\small \{lior.talker,aviad.cohen,alex.dana,m.dinerstein\}@samsung.com}~~~~~~~~~~~~~{\tt\small erez.yo@gmail.com}~~~~~~~~~~~~~~~
}

\begin{document}
\twocolumn[{%
\renewcommand\twocolumn[1][]{#1}%
\maketitle
\begin{center}
    \centering
                           
    \negvspace
    \negvspace
    \negvspace
    \begin{tabular}{cccc}
   \includegraphics[height=2cm,width=0.18\linewidth]{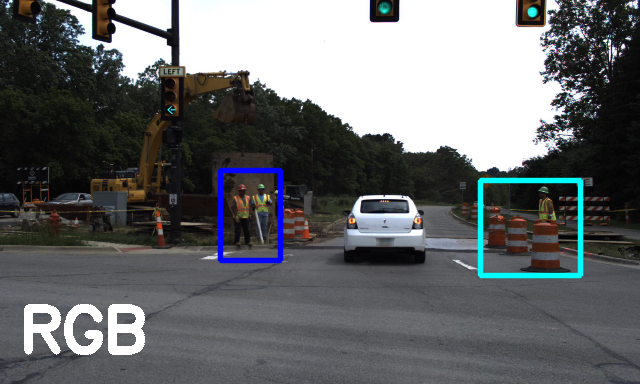}
   \includegraphics[height=2cm,width=0.18\linewidth]{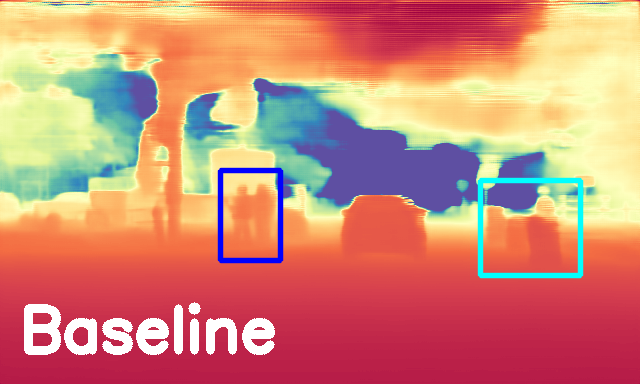}
   \includegraphics[height=2cm,width=0.18\linewidth]{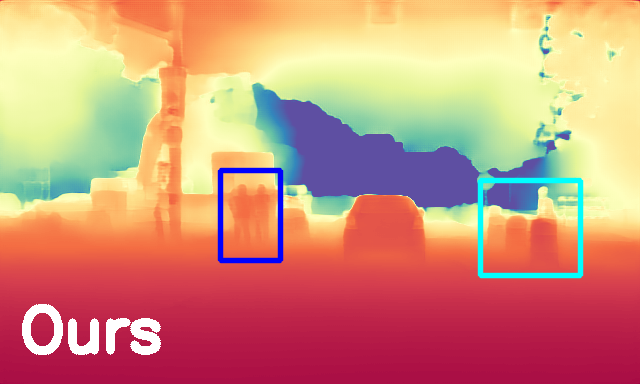}
   \includegraphics[height=2cm,width=0.42\linewidth]{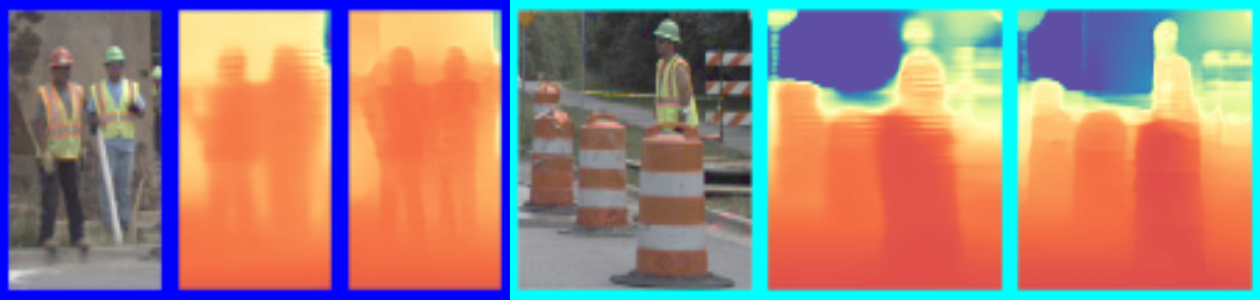} \\  
      ~~~~~~~~~~~~~~~~~~~~~~~~~~~~~~~~~~~~~~~~~~~~~~~~~~~~~~~~~~~~~~~~~~~~~~~~~~~~~~~~~~~~~~~~~\textbf{(a)}~~~~~~~~~~~
   RGB~~Baseline~Ours~~~~~~RGB~~~~~~Baseline~~~~~Ours \\
   \includegraphics[width=0.262\linewidth]{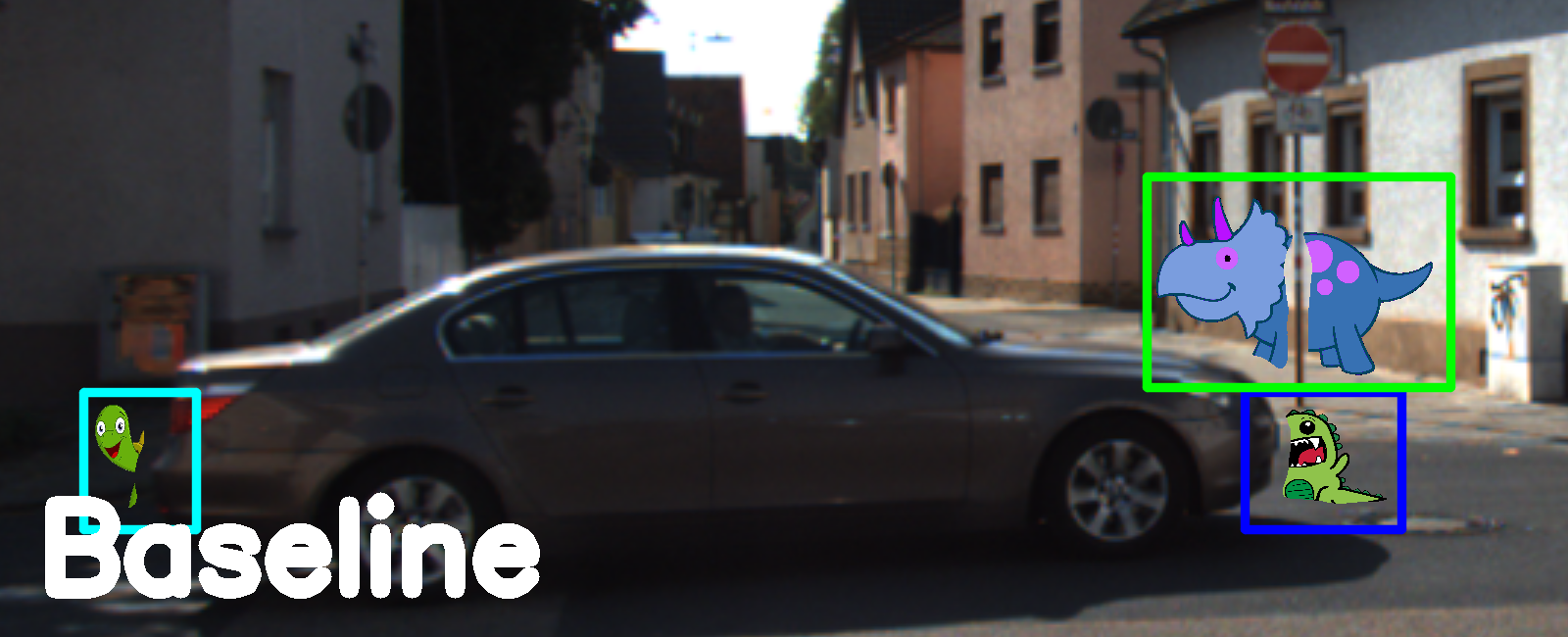}
   \includegraphics[width=0.262\linewidth]{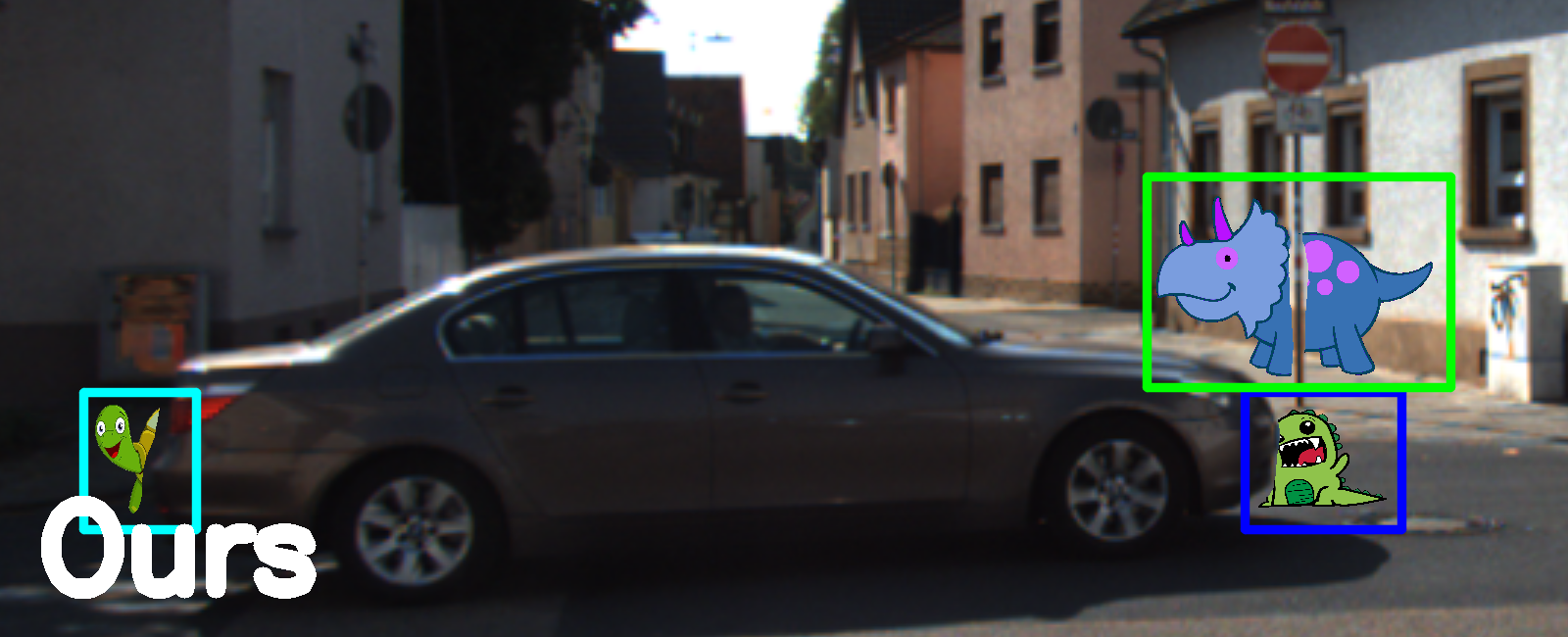} 
   \includegraphics[width=0.155\linewidth]{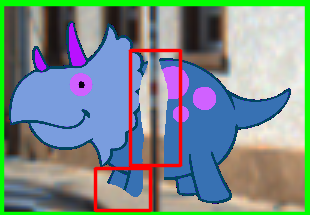}
   \includegraphics[width=0.155\linewidth]{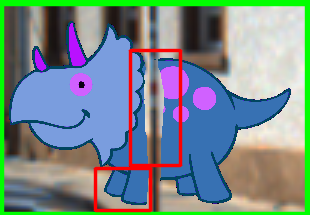}
   \includegraphics[width=0.119\linewidth]{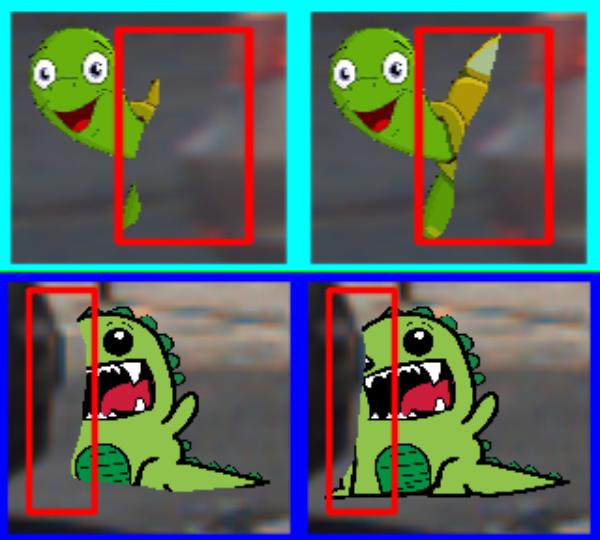} \\
   ~~~~~~~~~~~~~~~~~~~~~~~~~~~~~~~~~~~~~~~~~~~~~~~~~~~~~~~~~~~~~~~~~~~~~~~~~~~~~~~~~~~~~~~~~~~~~\textbf{(b)}~~~~~~~~~~~~~~~~Baseline
   ~~~~~~~~~~~~~~~~~~~~Ours
   ~~~~~~~~~~~Baseline~~Ours

	\end{tabular}
    \negvspace
    \captionof{figure}{
\textbf{Refining depth edges with our method when using Packnet-SAN \cite{guizilini2021sparse} as an MDE baseline.} (a) Depth estimation (DDAD dataset). Zoom-in on the crops (on the right) to see the improvement in the 2D localization of the depth edges between the baseline and our method. (b) Augmented reality. An example of virtual objects planted in a scene from the KITTI dataset for AR applications. Zoom-in and inspect the boundaries for the best impression of the depth edges accuracy.
    }
    \label{fig:fig1}
\end{center}%
}]

\begin{abstract}
\negvspace
Monocular Depth Estimation (MDE) is a fundamental problem in computer vision with numerous applications. Recently, LIDAR-supervised methods have achieved remarkable per-pixel depth accuracy in outdoor scenes. However, significant errors are typically found in the proximity of depth discontinuities, i.e., depth edges, which often hinder the performance of depth-dependent applications that are sensitive to such inaccuracies, e.g., novel view synthesis and augmented reality. Since direct supervision for the location of depth edges is typically unavailable in sparse LIDAR-based scenes, encouraging the MDE model to produce correct depth edges is not straightforward. To the best of our knowledge this paper is the first attempt to address the depth edges issue for LIDAR-supervised scenes.
In this work we propose to learn to detect the location of depth edges from densely-supervised synthetic data, and use it to generate supervision for the depth edges in the MDE training. 
To quantitatively evaluate our approach, and due to the lack of depth edges GT in LIDAR-based scenes, we manually annotated subsets of the KITTI and the DDAD datasets with depth edges ground truth. We demonstrate significant gains in the accuracy of the depth edges with comparable per-pixel depth accuracy on several challenging datasets. Code and datasets are available at \url{https://github.com/liortalker/MindTheEdge}.
\negvspace
\negvspace
\end{abstract}

\section{Introduction}
\label{sec:intro}

Monocular Depth Estimation (MDE) aims to recover the depth of each pixel in a single RGB image. It is used in many important applications, such as robotic navigation \cite{correa2012mobile}, novel view synthesis \cite{cao2022fwd} and Augmented Reality (AR) \cite{macedo2021occlusion}. Nonetheless, MDE is an ill-posed problem; that is, a single RGB image may be generated from many possible scenes. However, in recent years, many MDE methods, based on Convolutional Neural Networks (CNNs) have shown remarkable results. CNN-based supervised methods are trained with depth Ground Truth (GT) which is usually dense for indoor scenes, e.g., acquired using a RGBD camera \cite{silberman2012indoor}, and sparse for outdoor scenes, e.g., acquired using a LIDAR sensor \cite{geiger2012we}.

\begin{figure*}[t!]
\centering
\def\svgwidth{505pt}
\fontsize{10}{0}\selectfont
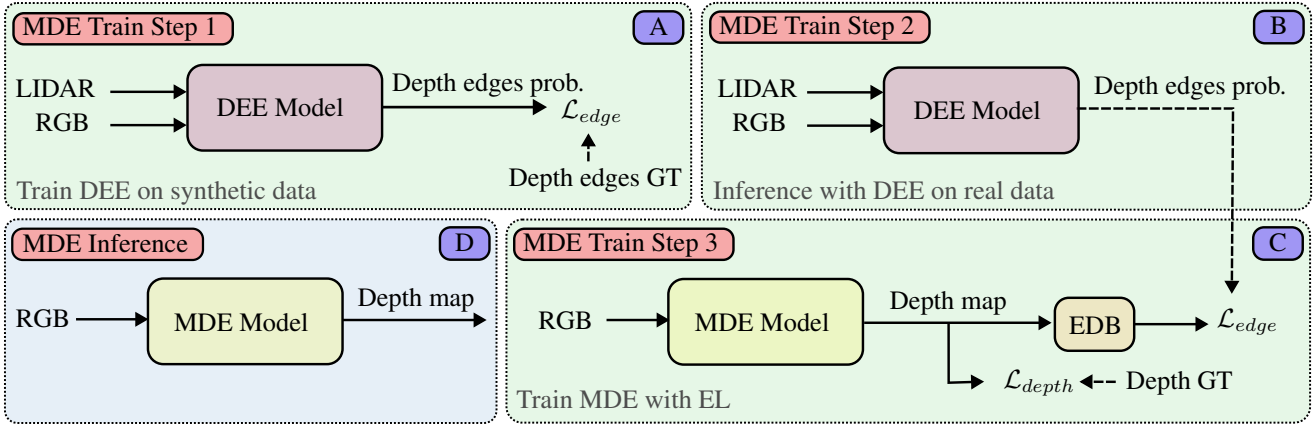
\caption{\textbf{Overview of our proposed MDE training method.} (A) Training the DEE model on synthetic data. (B) Inferring depth edges on the training set of the real data using the trained DEE model. (C) Training the MDE model on real data with the Edge Loss (EL) using the supervision from the previous step. (D) Inference using the MDE model on the real data. Solid and broken lines represent the dataflow and the GT used in loss functions, respectively.}
\label{fig:flow}
\vspace{-0.3cm}	
\end{figure*}


It was observed in several papers \cite{miangoleh2021boosting, xian2020structure, zhu2020edge} that the 2D locations of depth discontinuities, which we refer to as \textit{depth edges} in this paper, are often poorly estimated, resulting in thick smooth depth gradients or incorrectly-localized edges (see the baseline in Fig.~\ref{fig:fig1}). 
Some applications that use predicted depth maps are highly sensitive to errors in depth edges, which are often part of the silhouette of objects. One example of such an application is Novel View Synthesis (NVS) -- generating a new view of a scene captured from one or more views. In NVS methods that use depth explicitly \cite{cao2022fwd}, these type of localization errors in the 2D location of depth edges, 
may result in wrongly localized object parts in another newly generated view. Another important application that often uses predicted depth is virtual object rendering for AR \cite{macedo2021occlusion}, which computes for each pixel the closest occluding object from the point of view of the user. When relying on inaccurate depth edges in the computation of the occlusion, significant artifacts and unrealistic appearance may occur (see Fig.~\ref{fig:fig1}).

\begin{figure}[t]
\centering
\begin{tabular}{cc}
\includegraphics[width=0.80\linewidth]{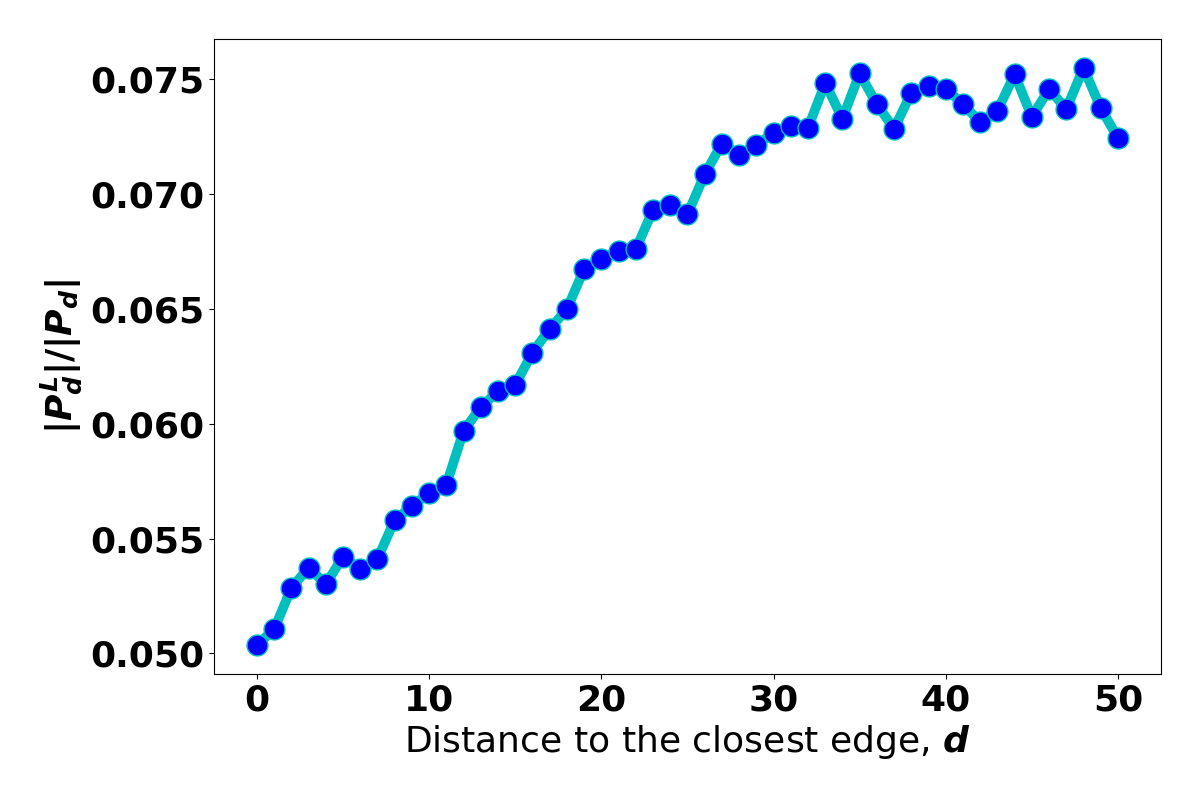} 
\includegraphics[width=0.136\linewidth]{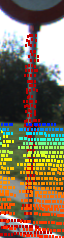} \\
~~~~~~~~~~~~~~~~~~~~~~~~~~~~~(a) ~~~~~~~~~~~~~~~~~~~~~~~~~~~~~~~~~~~~~~~~ (b)
\end{tabular}
\vspace{-0.3cm}
\caption{\textbf{The density of the LIDAR near edges in a partial set of the KITTI dataset (our proposed KITTI-DE dataset).} Denote the set of all pixels with a distance $d$ to the closest edge as $P_d$, and the set of pixels, $p$, in $p\in P_d$ with LIDAR measurement as $P_d^L$. (a) The ratio of LIDAR measurements, $|P_d^L|/|P_d|$, out of all pixels in $P_d$, as a function of $d$. (b) An example from the KITTI dataset of a gap in the LIDAR measurments (left of the pole) and an infiltration of LIDAR measurements from the background to the pole (right of the pole). For visualization purposes the LIDAR measurements are dilated.} 
\label{fig:lidar_near_edges}
\vspace{-0.4cm}
\end{figure}

Roughly speaking, the MDE network solves two subtasks: (i) depth edge estimation, and (ii) continuous surface depth estimation. Our method is based on the observation that MDE networks tend to focus on the latter much more than the former. The underlying reasons for this behaviour are probably twofold. The first is the small impact of the depth edges on the network's loss since they occupy only a tiny portion of the image. The second, which is partially discussed in \cite{cheng2019noise, xu2019depth}, is due to alignment errors between the RGB image and the LIDAR signal. Specifically, LIDAR measurements are often wrongly associated with objects although belonging to the background, or absent in occluded areas that are revealed in the time gap between the RGB and the LIDAR data acquisition (Fig.~\ref{fig:lidar_near_edges}b). This phenomenon is also depicted in Fig.~\ref{fig:lidar_near_edges}a for the LIDAR of the KITTI dataset \cite{geiger2012we}, where regions close to depth edges suffer from a lower density of LIDAR measurements.



In this paper we propose to improve the accuracy of the depth edges in MDE methods by directly encouraging the depth edges of the predicted depth map to be well-localized. Assuming that depth edges GT is available, a dedicated \textit{depth edges loss} that encourages the network to generate depth discontinuities in the correct locations may be used. However, due to the sparsity of the depth GT in typical outdoor scenes (e.g., KITTI), which are considered in this paper as our target domain, obtaining accurate depth edges GT is a challenging task. To this end, we are the first to propose a method to tackle this problem in sparsely-supervised MDE methods. We note that another aspect of depth edges accuracy, namely edge \emph{sharpness} \cite{chen2019over}, is not considered in this paper, and is complementary to our work.

To this end, we propose to train a Depth Edge Estimation (DEE) network (Sec.~\ref{sec:dee}), which predicts the probability that a pixel is located on a depth edge, using an accurate dense depth GT of a synthetic dataset \cite{hurl2019precise} (Fig.~\ref{fig:flow}A). Using the DEE network we infer a probabilistic map of depth edges on the training set of the target real domain (Fig.~\ref{fig:flow}B). These maps are then used to guide the proposed edge loss in the training of the MDE network (Fig.~\ref{fig:flow}C).
Although training on synthetic data and inferring on real data is known to lead to a performance decrease due to a 'domain gap' \cite{sankaranarayanan2018learning,pnvr2020sharingan}, our experiments (Sec.~\ref{sec:experiments}) demonstrate that the DEE network performs very well in practice. In particular, the DEE network's obtained depth edges are significantly more accurate than the depth edges 'naturally' obtained in the standard MDE training (Fig.~\ref{fig:prec_recall}). This is probably, as discussed above, due to the low attention paid to estimating the correct locations of the depth edges in the MDE training. Importantly, the MDE, which is trained only on the target dataset, benefits from the depth edges predictions of the DEE network despite their imperfection.




Due to the lack of depth edges GT for evaluation in LIDAR-supervised real-world datasets, we manually annotated depth edges in two evaluation sets of 102 and 50 images from the KITTI and DDAD datasets, respectively (see an example in the top of Fig.~\ref{fig:qualitive_ddad}). We show on these newly annotated datasets that our approach generates depth with significantly more accurate depth edges than the MDE baseline, while maintaining a similar depth accuracy.

Our main contributions are twofold:
\begin{itemize}
  \item A novel method to improve the localization of depth edges in MDE methods while preserving their per-pixel depth accuracy.
  \item A benchmark of human-annotated depth edges to evaluate their quality in sparsely-supervised MDE algorithms.
\end{itemize}

%

\section{Previous Work}
\label{sec:prev_work}
Depth estimators can be trained using full supervision with LiDAR or other absolute depth measurements \cite{lee2019big,guizilini2021sparse,bhat2021adabins,song2021monocular, agarwal2023attention, piccinelli2023idisc, ning2023trap}, by self-supervision using two or more RGB images \cite{zhou2017unsupervised,godard2019digging, gordon2019depth, guizilini20203d,watson2021temporal} or using semi-supervision \cite{kuznietsov2017semi,amiri2019semi,gurram2021monocular,wimbauer2021monorec}. The main objective of these methods is to reduce the overall mean absolute relative error (ARE), recently achieving state-of-the art (SOTA) result with ARE close to 
5\% on the KITTI dataset \cite{guizilini2021sparse, agarwal2023attention, piccinelli2023idisc}. However, specific applications such as AR require that the edges of the predicted depth are also accurate.

To achieve this, various depth estimation models were trained in a fully-supervised manner using dense depth GT in datasets such as NYU-2 \cite{silberman2012indoor}, IBims-1 \cite{koch2018evaluation} and Middlebury \cite{scharstein2014high}. Dense outdoor depth maps include the small scale ETH3D \cite{schops2017multi} dataset and large-scale DIODE \cite{vasiljevic2019diode}. SharpNet \cite{ramamonjisoa2019sharpnet} improved depth prediction on object edges in indoor scenes by predicting occluding contours. Their solution was first pretrained to predict occluding contours on synthetic data and then fine-tuned on the indoor NYUv2 dataset to constrain normal, depth and occluding contours. In their subsequent work \cite{ramamonjisoa2020predicting} the authors predicted displacement fields of pixels with poorly predicted depth values of dense depth maps. DCTNet \cite{zhao2022discrete} boosted the resolution of depth maps from low-resolution ones using an edge attention and high-resolution RGB images, where they assume co-occurrence between the texture edges of RGB images and depth edges. 
These methods require training on dense GT collected using expensive short-range laser scanners which are mostly ineffective for outdoor scenes.


Other works utilized the partial consistency between the edges of semantic classes and depth edges to improve the depth predictions on these regions. A dedicated loss was developed \cite{zhu2020edge} which slightly improved the ARE near the edges of the segmentation masks. However, the edges between semantic classes do not necesserily match the depth edges; for example, depth edges between the instances of the same class (e.g., buildings).
\cite{saeedan2021boosting} and \cite{xian2020structure} used stereo images in addition to panoptic segmentation maps to refine depth edges. Recently, \cite{chen2023self} attempted to solve a phenomenon of 'edge-fattening' by redesigning the triplet loss for depth estimation. However, the solution is designed for self-supervised approaches, and their per-pixel accuracy is low. In contrast, our work improves depth edges without using additional segmentation maps from the target domain, while achieving comparable ARE to SOTA methods.

In an attempt to overcome the sparsity limitation of ground-truth depth maps for outdoor scenes, disparity maps were generated from the 3D Movies Dataset \cite{ranftl2020towards}. However, stereo matching is a non-trivial task, which is especially inaccurate around depth edges \cite{ranftl2020towards}. Moreover, without knowing the exact baseline and other camera characteristics, the estimated depth maps units cannot be metric-reconstructed as needed for applications that require absolute depth. MiDAS \cite{ranftl2020towards} and DPT \cite{ranftl2021vision} were also trained on this dataset, achieving visually striking depth maps. However, the lack of absolute scale limits their usage.


In recent works, \cite{miangoleh2021boosting} and \cite{dai2023multi} introduced methods to merge two depth maps at different resolutions, trained on dense depth datasets. The accuracy of \cite{miangoleh2021boosting} was evaluated using order ranking around depth edges, but did not measure the absolute depth metrics on sparsely-supervised datasets, while \cite{dai2023multi} did measure absolute depth metrics but not the accuracy around depth edges. To the best of our knowledge we are the first to consider the problem of depth edges in \emph{sparse LIDAR-based} scenes; therefore, no 'natural' candidates for comparison exist. Nevertheless, due to their impressive depth edges and their similar goal, we compare our method to \cite{miangoleh2021boosting} and \cite{dai2023multi} (BoostingDepth and GradientFusion, respectively) and show their limitations (Sec.~\ref{sec:miangoleh}).


\section{Method}
\label{sec:depth_edges}

The flow of our MDE training method can be described in three steps as illustrated in Fig.~\ref{fig:flow}A-C. In the first step, the DEE network is trained on the source synthetic dataset with the depth edges GT, to predict a probabilistic map of depth edges for a given RGB image and the corresponding LIDAR measurement (Fig.~\ref{fig:flow}A and Sec.~\ref{sec:dee}). In the second step, inference is applied on the training set of the target real data using the trained DEE network, resulting in (approximate) depth edge labels, $\tilde{E}_{GT}$ (Fig.~\ref{fig:flow}B). In the last step we train the MDE network with an Edge Detection Block (EDB) to improve the localization of the depth edges (Fig.~\ref{fig:flow}C and Sec.~\ref{sec:edb}). The training is carried out using a straightforward supervised depth loss with LIDAR as GT, and the proposed edge loss with $\tilde{E}_{GT}$ as approximate GT. In the following sections we describe our method and the used steps in detail.



\subsection{Monocular Depth Estimation}
MDE models are commonly trained to regress per-pixel depth, $D(I)$, from an RGB image, $I$, given a depth GT, $D_{GT}$, in $S$ scales, using a loss function:
\begin{equation}
\mathcal{L}_{depth}^{ms}=\frac{1}{S}\sum_s\mathcal{L}_{depth}(D^s(I),D_{GT}^s),
\label{eq:mde}
\end{equation}
where $D^s(I)$ and $D_{GT}^s$ indicate the predicted depth and GT depth in scale $s$, respectively. For brevity we omit the scale indexes for the rest of the paper, although, unless explicitly stated, all losses are multi-scale. 
\subsection{Edge Detection Block (EDB)}
\label{sec:edb}
To encourage the model to produce depth edges at the correct locations, we use a differentiable layer that computes the per-pixel probability of depth edges, $\tilde{E}(D(I))$, from the predicted depth map, $D(I)$, where $I$ is the input RGB image. Given $D(I)$, the EDB computes the magnitude of the spatial image gradient, $\vert\nabla{D(I)}\vert$, and then transforms it into an 'edge-ness' probability score:
\begin{equation}
\tilde{E}(D(I)) = sigmoid(\vert\nabla{D(I)}\vert-t_{grad})
\end{equation}
by shifting it with the paramater $t_{grad}$ and passing it through a sigmoid function.

In practice, when using the standard image gradient 
\begin{equation*}
\vert\nabla{D(I)}\vert = \sqrt{ \left( \dfrac{dD(I)}{dx} \right) ^2 + \left( \dfrac{dD(I)}{dy} \right)  ^2}, 
\end{equation*}
some cyclic gradient patterns may emerge since both $dD(I)/dx$ and $dD(I)/dy$ are unconstrained (see depiction in the supplementary material). Therefore, we compute the gradient as a derivative in the direction perpendicular to the edge. To this end, the normal direction to the edge is first computed from the depth edge GT (estimated using the DEE network - Sec.~\ref{sec:dee}), $\tilde{E}_{GT}$, by: 
\begin{equation*}
\theta = atan2\left( \dfrac{d\tilde{E}_{GT}}{dy},\dfrac{d\tilde{E}_{GT}}{dx} \right)
\end{equation*}
The derivative in the direction perpendicular to the edge for every pixel $(x,y)$ is therefore given by:
\begin{equation*}
\begin{split}
\nabla_{\theta}{D_I(x,y)} = D_I(x+\cos\theta,y+\sin\theta) - \\D_I(x-\cos\theta,y-\sin\theta),
\end{split}
\end{equation*}
where $D_I=D(I)$ and the coordinates $x \pm \cos\theta$ and $y \pm \sin\theta$ are rounded in practice. 
The predicted edge probability, $\tilde{E}(D(I))$, is then used by the depth edges loss as described in the following section.

\subsection{Depth Edges Loss}
\label{sec:de_loss}
Given the depth edges GT, $\tilde{E}_{GT}$, and the output of the EDB, $\tilde{E}(D(I))$, we use the following loss to encourage depth discontinuity at $\tilde{E}_{GT}$:
\begin{equation}
\begin{split}
&\mathcal{L}_{edge}(\tilde{E}(D(I)),\tilde{E}_{GT}) = BBCE\Big(\tilde{E}(D(I)),\tilde{E}_{GT}\Big)
\label{eq:loss}
\end{split}
\end{equation}
where the $BBCE$ \cite{xie2015holistically} is the Balanced Binary Cross Entropy loss where the positives (edge pixels) and the negatives (non-edge pixels) are reweighted in a standard BCE loss so they have equal contribution.

\subsection{Total Loss}
\label{sec:total_loss}
Our MDE model is trained with a linear combination of the edge loss, $\mathcal{L}_{edge}$, and the standard depth loss, $\mathcal{L}_{depth}^{ms}$ (Eq.~\ref{eq:mde}). The total loss is given by
\begin{equation}
\begin{split}
\mathcal{L} & = \mathcal{L}_{depth}^{ms}(D(I),D_{GT}) + \alpha \mathcal{L}_{edge}(\tilde{E}(D(I)),\tilde{E}_{GT}),
\label{eq:total_loss}
\end{split}
\end{equation}
where $\alpha$ is a parameter to balance between the two losses, and $D_{GT}$ is the depth GT.

\subsection{Depth Edges Estimation (DEE)}
\label{sec:dee}
Since the actual depth edges GT is unavailable for the target real data we use the DEE network, which predicts depth edges, $\tilde{E}(I,D')$, from RGB, $I$, and LIDAR, $D'$. We note that the LIDAR measurements have significant impact on the performance of the DEE network when given as input in addition to the RGB (ablation study is presented in the supplementary material). In order to train the DEE network we use a synthetic dataset with dense depth and LIDAR that are available for each RGB image. To extract depth edges GT, $E_{GT}$, we use the Canny edge detector \cite{canny1986computational} on the dense depth GT. The DEE network is trained with the depth edges loss as presented in Eq.~\ref{eq:loss}; that is, $\mathcal{L}_{edge}(\tilde{E}(I,D'),E_{GT})$. 

The predicted depth edges, $\tilde{E}(I,D')$, obtained from the DEE network are dense with Gaussian-like distributions. To produce one-pixel wide thin depth edges, we use two standard edge-detection post-processing operations \cite{canny1986computational}: (a) Non-Maximum Suppression (NMS), and (b) hysteresis (with $0.85$ and $0.9$ as low and high parameters).



\section{Experiments}
\label{sec:experiments}

In the following we describe experiments that demonstrate the effectiveness of our suggested method and compare it to the SOTA and other relevant methods. 


\subsection{Datasets}
To train the DEE network we use the GTA-PreSIL dataset \cite{hurl2019precise}, which is rendered from the Grand Theft Auto (GTA) video game. It consists of $\sim$44K images for training and $\sim$7K images for validation and testing, where each image has a corresponding dense depth GT. To generate the GT for the depth edges, we use Canny edge detector \cite{canny1986computational} with low and high thresholds of $4$ and $5$ meters, respectively. We note that the DEE network could have been trained with Virtual KITTI \cite{gaidon2016virtual}, probably achieving better performance on KITTI (details below) due to a smaller domain gap between the datasets. However, since we aim to demonstrate a general-purpose DEE network, GTA-PreSIL was the better choice since its structures and styles are significantly different from the target real domains.

We demonstrate our method on three different datasets: Synscapes \cite{wrenninge2018synscapes}, KITTI \cite{Uhrig2017THREEDV} and  DDAD \cite{packnet}. The data split for the highly realistic Synscapes dataset, where each image has a corresponding dense depth, is similar to \cite{wang2022semi}. 
We have chosen this dataset since (1) it provides perfect depth edges GT to evaluate our method, and (2) it has dense depth that allows us to fully analyze the impact of our method on the per-pixel depth accuracy. For KITTI we use the Eigen split \cite{eigen2014depth}, 
and for the DDAD dataset we use the official split. 

\vspace{-0.3cm}
\paragraph{KITTI \& DDAD Depth Edges Evaluation Sets:} To enable direct evaluation of the depth edges of MDE networks on the KITTI and DDAD datasets, we introduce the KITTI Depth Edges (KITTI-DE) and the DDAD Depth Edges (DDAD-DE) evaluation sets which consist of depth edges annotations of 102 images from the KITTI dataset and 50 images from DDAD dataset, respectively. To ease the manual annotation process, we started from the corresponding images of the Semantic and Instance Segmentation evaluation benchmark of KITTI \cite{Alhaija2018IJCV} and the validation set of DDAD. To get an initial approximation of the depth edges we derive the edges of the instance segmentation masks for the relevant classes, and the semantic segmentation maps for the other classes. Then, false depth edges were manually removed (e.g. car wheels point of contact on the road), and missing depth edges were added (e.g., edges between building). See the supplementary material for additional details of the annotation process.
We note that the objective of the proposed depth edges evaluation sets is to complement the standard depth evaluation sets, not to replace it, thus allowing to inspect another important aspect of MDE.

\begin{figure}[t]
\centering
\begin{tabular}{c}

   \includegraphics[width=0.92\linewidth]{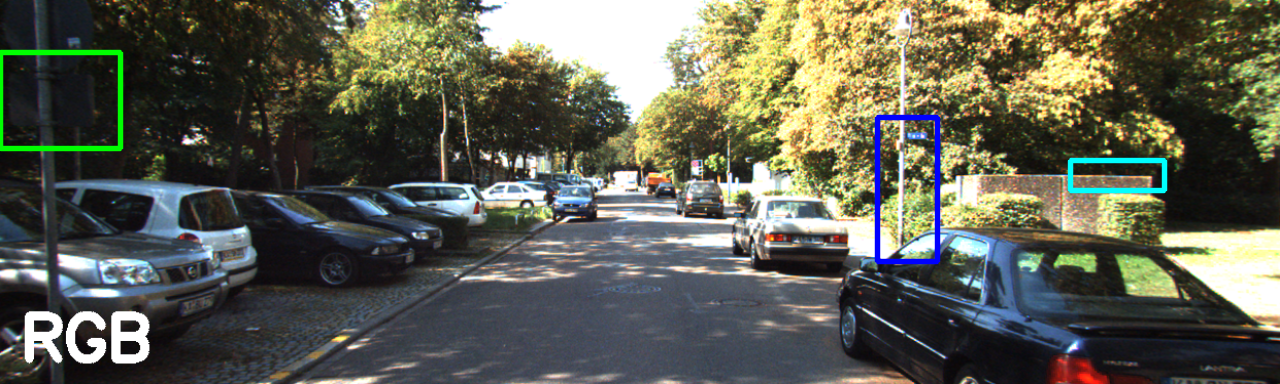} \\
   \includegraphics[width=0.92\linewidth]{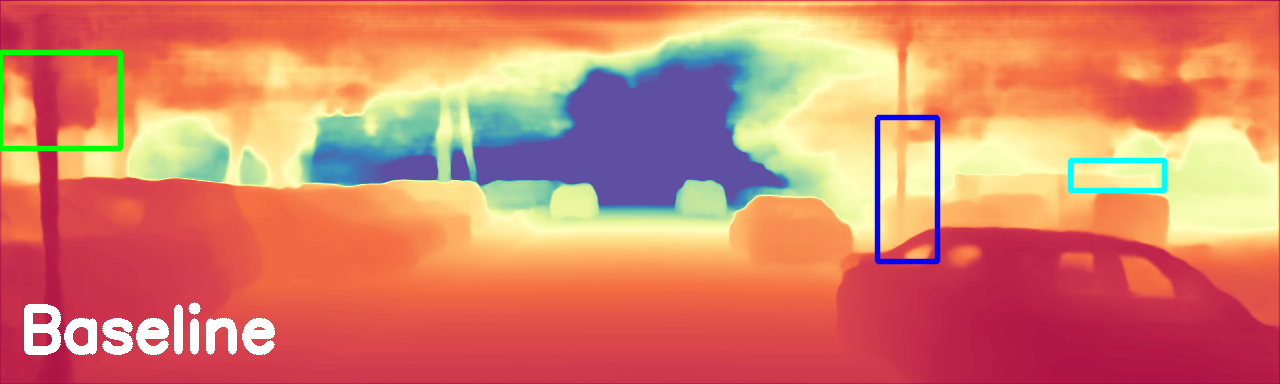} \\
   \includegraphics[width=0.92\linewidth]{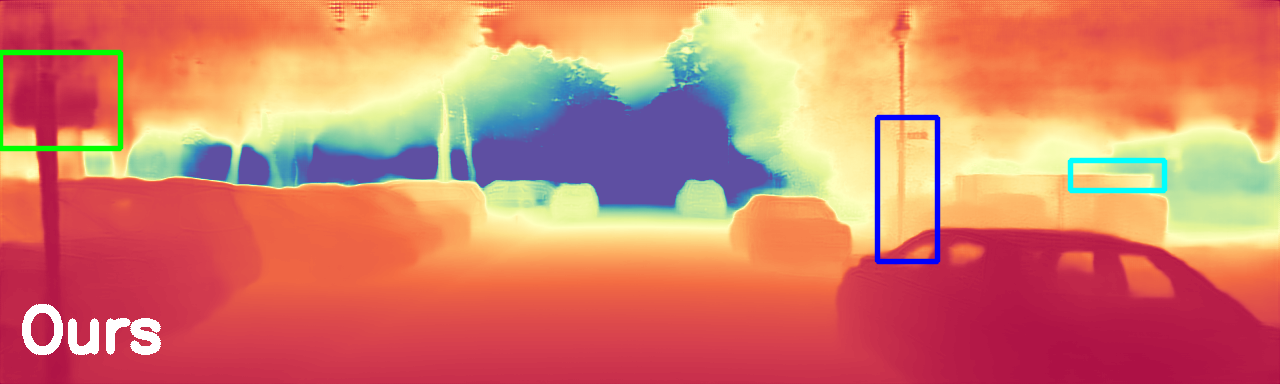}  \\
   \includegraphics[width=0.92\linewidth]{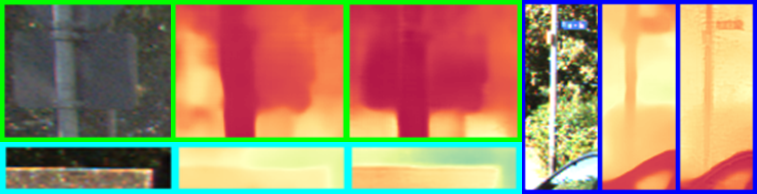} \\ ~~~~~~~RGB~~~~~~~~~~Baseline~~~~~~~~~~Ours~~~~~RGB~Baseline~Ours

\end{tabular}
\caption{\textbf{Examples of depth predictions in the KITTI-DE dataset.} The depth predictions for the baseline and our method correspond to Packnet-SAN and Packnet-SAN+EL, respectively.}
\label{fig:qualitive_kitti}
\vspace{-0.5cm}
\end{figure}

\vspace{-0.1cm}
\subsection{Metrics}
\vspace{-0.1cm}
To evaluate the performance of the MDE networks, we use some of the common per-pixel MDE depth metrics, as well as metrics to evaluate the quality of the edges of the predicted depth. 
Both depth and depth edges are evaluated in the bottom 60\% of the image where LiDAR is commonly available, as defined in Garg et al. \cite{garg2016unsupervised}, for all datasets.

\vspace{-0.4cm}	
\paragraph{Depth metrics:} We use the common Absolute Relative Error (ARE) to measure the per-pixel depth error, given by $ARE(\tilde{d},d)=|\tilde{d}-d|/d$, where $\tilde{d}$ and $d$ are the predicted depth and GT depth, respectively. Note that the ARE is computed only over the sparse LIDAR measurements in the KITTI and the DDAD datasets, which are very sparse near edges (Fig.~\ref{fig:lidar_near_edges}), making the ARE a poor metric to measure the quality of the 2D localization of depth edges.

\vspace{-0.4cm}	
\paragraph{Depth edges metrics:} To evaluate the edge quality of the predicted depth, one option is the ORD metric \cite{xian2020structure}, which measures the percentage of order disagreements between pairs of depth points in the predicted depth against the GT depth. We argue that this metric suffers from a similar problem as the ARE - since LIDAR measurements are sparse, many possible 2D edges can obtain the same ORD value. 

Therefore, in addition to the ORD metric, we use the most common metric from the edge detection literature, BSDS \cite{martin2004learning}. It consists of finding a bijective matching, $\mu(\tilde{E},E)$, between the edge pixels of the predicted depth, $\tilde{E}$, and the edge pixels of the GT, $E$. Each match $(\tilde{e},e)\in\mu(\tilde{E},E)$ between edge pixel, $\tilde{e}\in \tilde{E}$, and edge pixel GT, $e\in E$, is obtained such that the locations of the pixels are similar; that is, $||\tilde{e},e||_2\leq t_e$ where $t_e$ is a small threshold (we use $t_e=2$). The precision, $PR(\tilde{E},E)$, and recall, $RE(\tilde{E},E)$, are then given by $|\mu(\tilde{E},E)|/|\tilde{E}|$, and $|\mu(\tilde{E},E)|/|E|$, respectively. 

In practice, $\tilde{E}$ is obtained by extracting edges from the predicted depth $d$ using Canny edge detector with low and high thresholds of $th_{low}$ and $th_{high}$, respectively. Since there is an inherent tradeoff between the precision and recall, we invoke multiple runs of a Canny edge detector with different parameters to obtain a precision-recall curve. Finally, to quantify the depth edges quality in a single number, we compute the Area Under the Curve (AUC). Since a large part of the graph is uncovered by all MDE algorithms, we report both the AUC for a partial representative range and for all range ([0,1]).
Note that the edge metrics used in the I-BIMS1 dataset \cite{koch2018evaluation}, $\epsilon^{acc}_{DBE}$ and $\epsilon^{comp}_{DBE}$, define a matching between the predicted edges and the GT edges, similarly to our AUC metric. However, they suffer from a significant limitation in comparison to our AUC metric since their matching is not bijective and allows multiple-to-multiple matching. For example, broken and non-consecutive predicted edge can be perfectly matched to a consecutive GT edge, and yield the best score under these metrics.

\begin{figure*}[t]
\centering
\begin{tabular}{cc}
KITTI-DE & DDAD-DE \\
\includegraphics[width=0.48\linewidth]{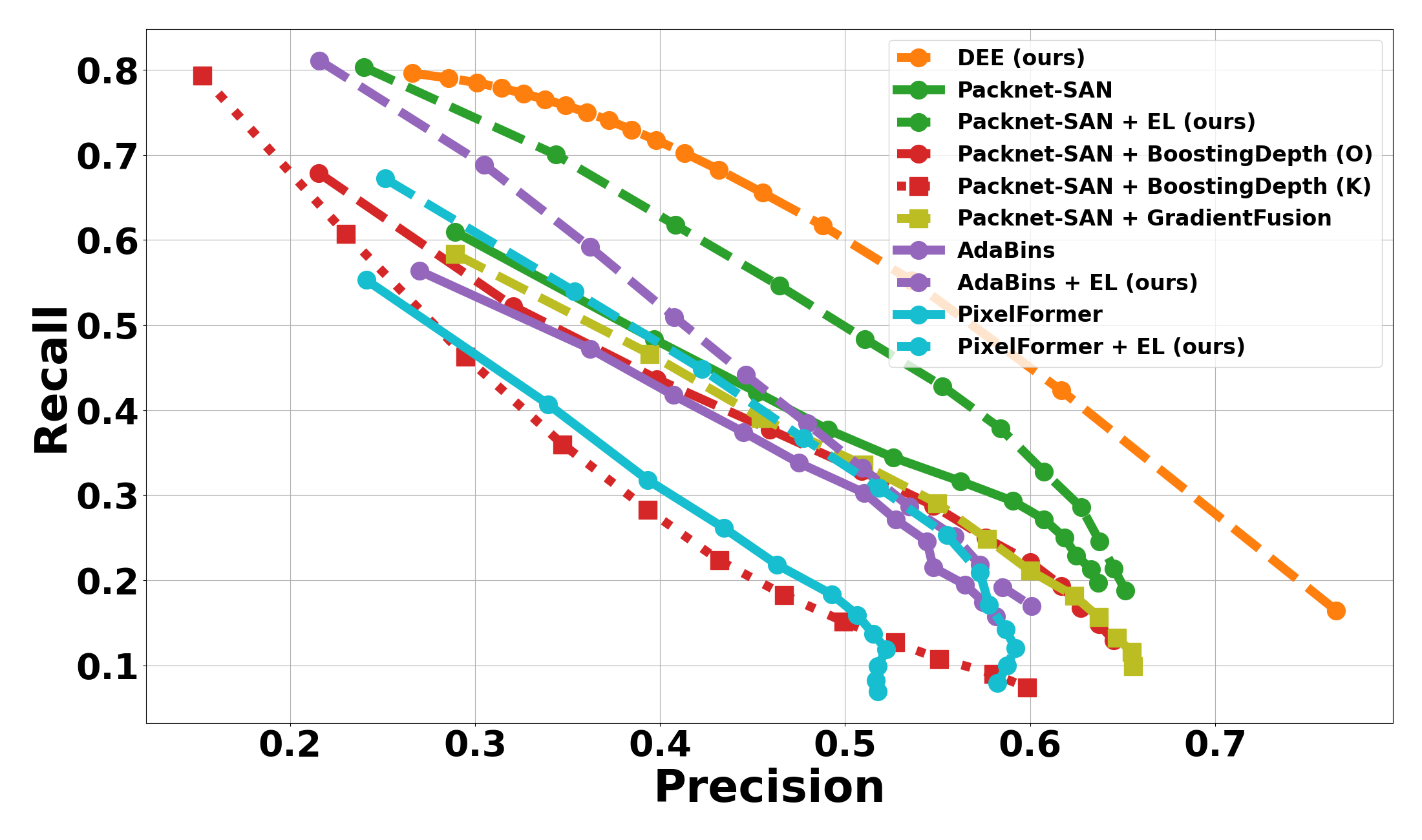} &
\includegraphics[height=4.95cm, width=0.48\linewidth]{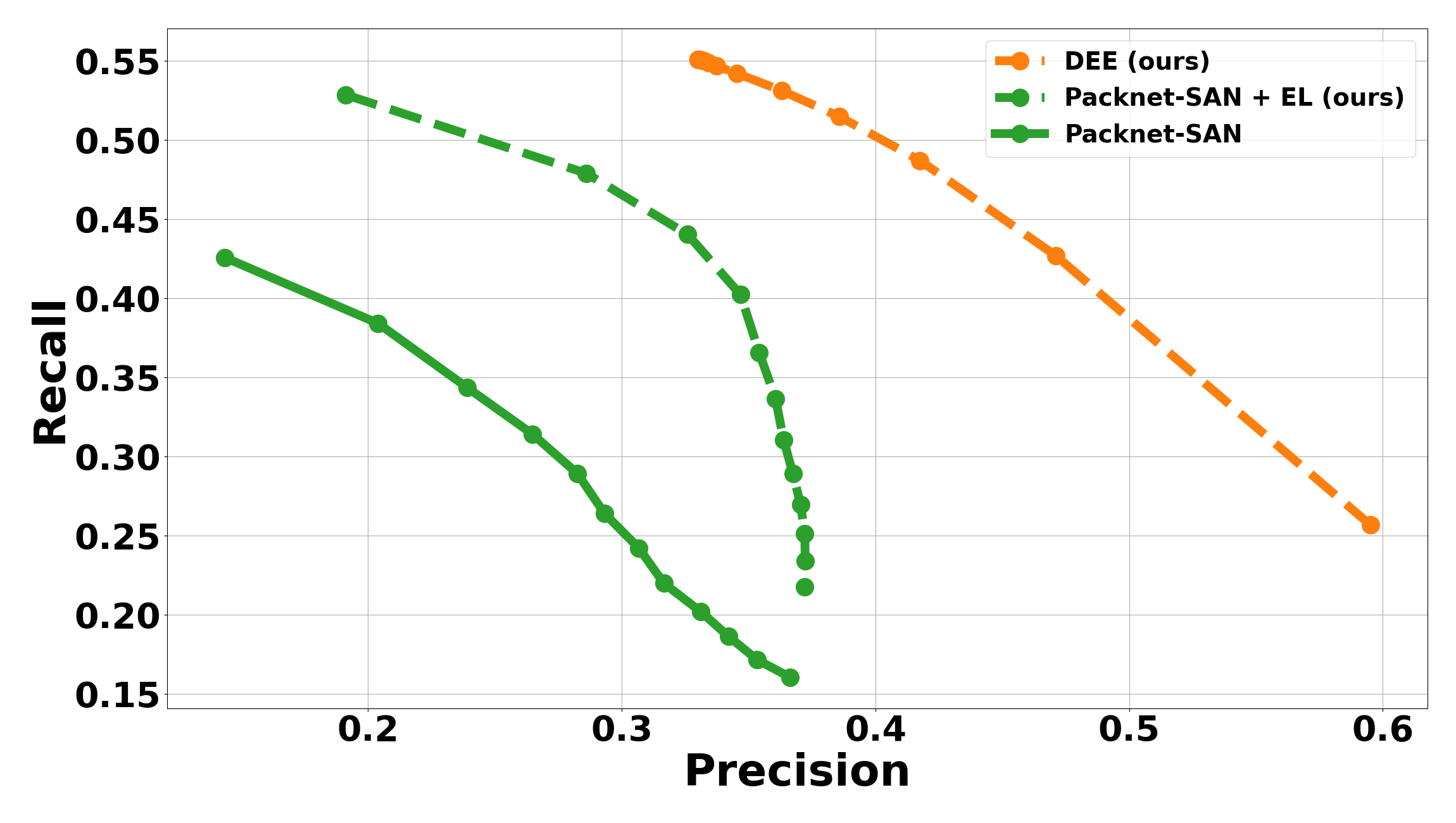}
\end{tabular}
\vspace{-0.5cm}
\caption{\textbf{Precision and recall of the depth edges on the KITTI-DE and DDAD-DE evaluation sets}. Each point on the graphs of the MDE methods is generated with different parameters of the Canny edge detector. Each of the points on the graphs that correspond to the DEE method is generated by thresholding the depth edge probability in the range $(0,1)$.}
\label{fig:prec_recall}
\end{figure*}

\renewcommand{\tabcolsep}{4pt}
{\begin{table*}[t]
	\begin{center}
    		\begin{tabular}{ | l | c | c | c | c | c | c | c |}
    		\hline
				 \textbf{Method} & \multicolumn{4}{c|}{\textbf{KITTI-DE}} & \multicolumn{3}{c|}{\textbf{KITTI test}} \\ \hline
			  & AUC (edges) $\uparrow$ & ORD $\downarrow$ & ARE $\downarrow$ & $\delta < 1.25 \uparrow$  & ORD $\downarrow$ & ARE $\downarrow$ & $\delta < 1.25 \uparrow $ \\ \hline \hline	
			\textbf{Packnet-SAN} & 47.56\% (39.40\%) & 7.68\% & 3.45\% & 98.66\% & 12.40\% & 6.17\% & 95.39\% \\ \hline
			\textbf{Packnet-SAN + BoostingDepth (O)} & 46.04\% (37.07\%) & 10.35\% & 9.32\% & 88.90\% & 12.63\% & 11.10\% & 86.41\% \\  \hline
			\textbf{Packnet-SAN + BoostingDepth (K)} & 36.19\% (31.27\%) & 9.47\% & 7.24\% & 93.62\% & 11.45\% & 8.33\% & 91.99\% \\  \hline
			\textbf{Packnet-SAN + GradientFusion} & 44.51\% (34.10\%) & 9.18\% & 5.93\% & 95.66\% & 11.15\% & 7.18\% & 94.17\% \\  \hline
			\textbf{Packnet-SAN + EL (ours)} & \textbf{61.87\% (49.02\%)} & 7.75\% & 3.61\% & 98.53\% & 12.48\% & 6.50\% & 95.06\% \\\hline \hline
			\textbf{AdaBins} & 41.23\% (34.11\%) & 7.69\% & 3.14\% & 98.78\% & 10.14\% & 6.28\% & 95.85\% \\ \hline 
			\textbf{AdaBins + EL (ours)} & \textbf{53.47\% (44.00\%)} & 7.64\% & 3.11\% & 98.79\% & 10.13\% & 6.21\% & 95.87\% \\  \hline \hline
			\textbf{PixelFormer} & 32.79\% (26.44\%) & 7.47\% & 3.00\% & 98.79\% & 7.56\% & 5.45\% & 96.98\% \\ \hline 
			\textbf{PixelFormer + EL (ours)} & \textbf{46.23\% (35.33\%)} & 7.53\% & 2.94\% & 98.80\% & 7.58\% & 5.59\% &  96.72\% \\  \hline
    		\end{tabular}
	\end{center}
	\vspace{-0.4cm}
\caption{\textbf{Results on the KITTI dataset.} The AUC is given for the range where at least one MDE method has valid measurement: [0.12,0.65]. In parentheses we also report the AUC of the full [0,1] range. In BoostingDepth, O is for the original training (dense data) by the authors, and K is for our training (KITTI data).}
\label{tab:results_kitti}
\vspace{-0.3cm}
\end{table*}
}


\subsection{Results}
We use our method with three SOTA sparsely-supervised MDE methods as baselines: Packnet-SAN \cite{guizilini2021sparse}, AdaBins \cite{bhat2021adabins} and PixelFormer \cite{agarwal2023attention}. For all methods we use the publicly available code and weights and resume training for ten more epochs with our proposed loss (Sec.~\ref{sec:de_loss}), and without it as a baseline for the comparison. In the following sections, we refer to the original methods by Packnet-SAN, AdaBins and PixelFormer, and refer to these methods with the addition of our edge loss (Sec.~\ref{sec:depth_edges}) as \{method name\} + EL (ours), where EL is abbreviation for 'Edge Loss'. We further show on the supplementary material that training simoultaneously (or sequentially) on both source (GTA-PreSIL) and target (KITTI or DDAD) datasets is inferior to our method. We note that, since Packnet-SAN exhibits better depth edges than both other methods, we use it as our main baseline in all experiments.

\vspace{-0.3cm}
\subsubsection{The KITTI dataset}
In Fig.~\ref{fig:qualitive_kitti} we present an image from the KITTI-DE dataset alongside the depth predictions of the baselines and our method. It can be seen, for example on the crops in the bottom row, that our method has considerably more accurate depth edges than the baseline in many parts of the scene.

The precision-recall curves for the KITTI-DE dataset, which are presented in the leftside of Fig.~\ref{fig:prec_recall}, are computed for the baselines and competing methods (solid curves) and for our method with the edge loss (non-red dashed curves). The precision-recall curves of our method in comparison to the corresponding baselines, when compared using the same precision (same $x$ coordinate), have a significantly higher recall. In Tab.~\ref{tab:results_kitti}, the AUC metric for the edge quality is presented, where our method achieves an AUC of $61.87\%$, $53.47\%$ and $46.23\%$ in comparison to the Packnet-SAN, AdaBins and PixelFormer baselines that achieve an AUC of $47.56\%$, $41.23\%$ and $32.79\%$, respectively, indicating a relative improvement between $30\%$ and $40\%$.

The quality of the per-pixel depth is also evaluated on the KITTI-DE dataset, and the ARE is presented in Tab.~\ref{tab:results_kitti}. The Packnet-SAN, AdaBins and PixelFormer baselines achieve an ARE of $3.45\%$, $3.14\%$ and $3.00\%$, respectively, in comparison to our Packnet-SAN + EL, AdaBins + EL and PixelFormer + EL that achieve an ARE of $3.61\%$, $3.11\%$ and $2.94\%$, respectively, which indicates a comparable performance. We also evaluate the methods on the standard KITTI test set (Tab.~\ref{tab:results_kitti}), where only the per-pixel depth quality can be estimated. The Packnet-SAN, AdaBins and PixelFormer baselines achieve an ARE of $6.17\%$, $6.28\%$ and $5.46\%$, respectively, in comparison to our Packnet-SAN + EL, AdaBins + EL and PixelFormer + EL that achieve an ARE of $6.50\%$, $6.21\%$ and $5.59\%$, respectively, which also indicates a comparable performance. We note that, interestingly, the quality of edges is not highly correlative to the per-pixel depth quality, which demonstrates the need for the depth edges annotated datasets we propose.

\begin{figure*}[t]
\centering
\begin{tabular}{ccc}
   \includegraphics[width=0.236\linewidth]{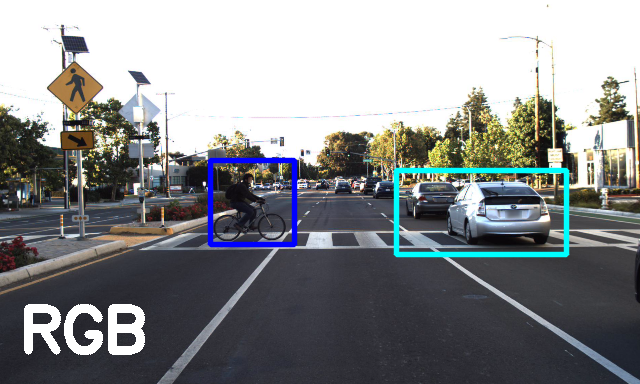}
   \includegraphics[width=0.236\linewidth]{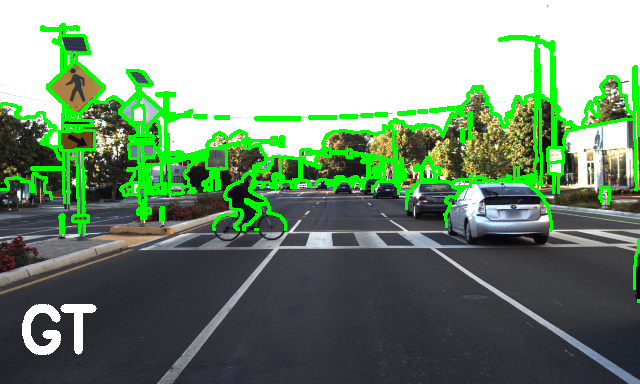} 
   \includegraphics[width=0.236\linewidth]{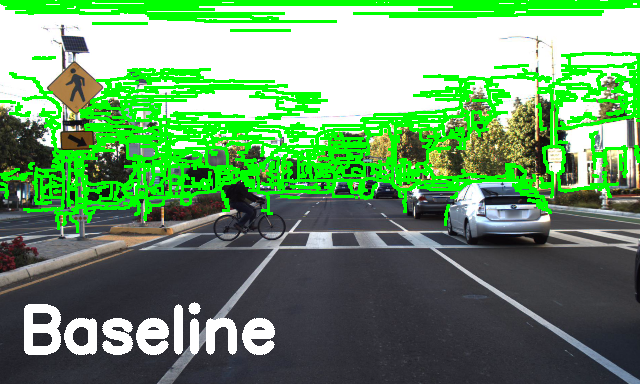} 
   \includegraphics[width=0.236\linewidth]{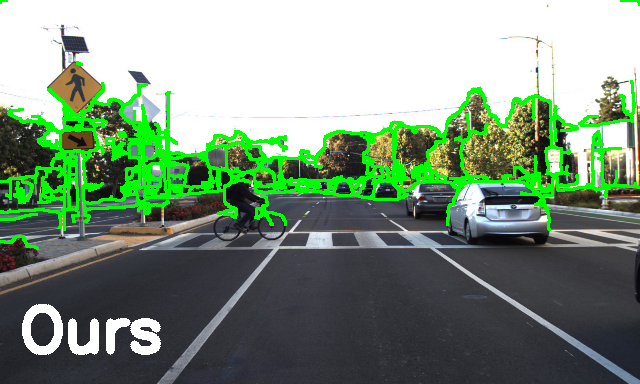} \\
   \includegraphics[width=0.297\linewidth]{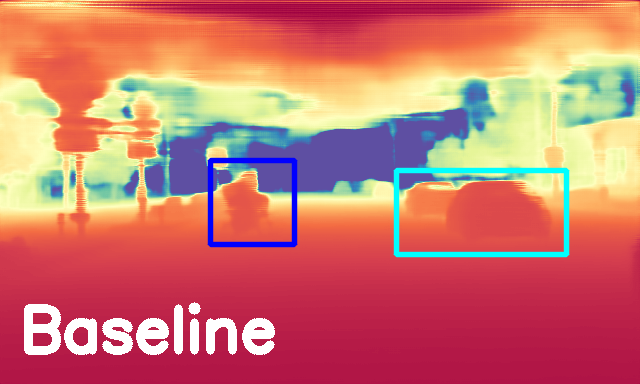} 
   \includegraphics[width=0.297\linewidth]{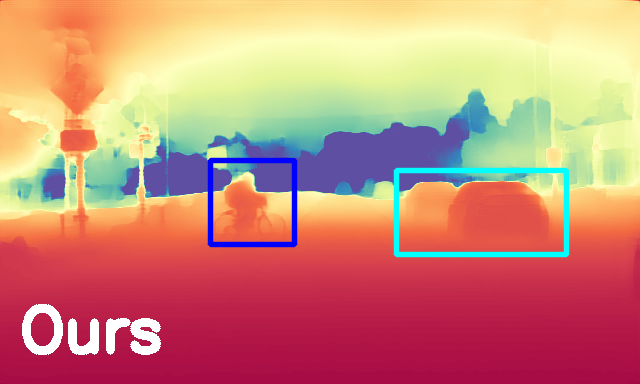}
   \includegraphics[height=3.1cm,width=0.355\linewidth]{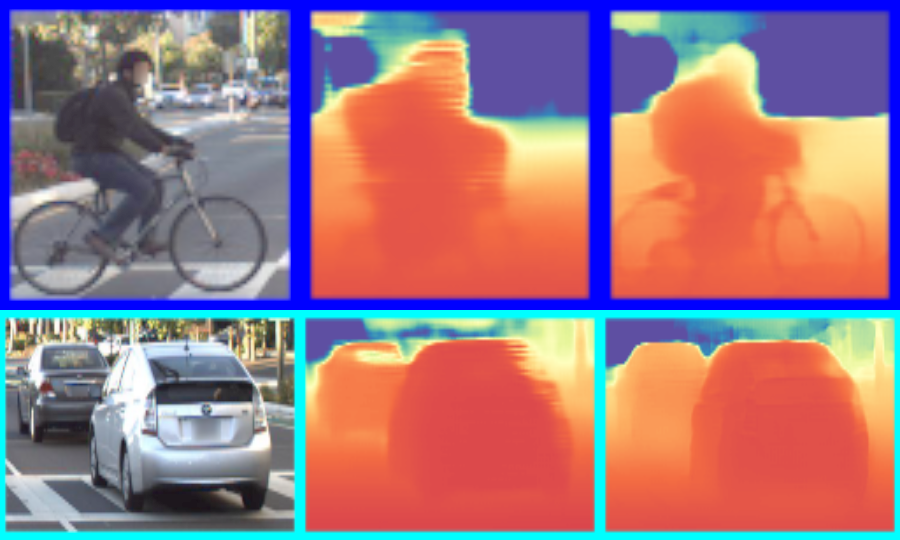} \\
~~~~~~~~~~~~~~~~~~~~~~~~~~~~~~~~~~~~~~~~~~~~~~~~~~~~~~~~~~~~~~~~~~~~~~~~~~~~~~~~~~~~~~~~~~~~~~~~~~~~~~~~~~~~~~~~~~~~~~~~~~RGB
   ~~~~~~~~~~~~Baseline
   ~~~~~~~~~~~~Ours \\

\end{tabular}
\vspace{-0.3cm}
\caption{\textbf{Examples of depth predictions of Packnet-SAN and Packnet-SAN + EL (ours) on images from the DDAD-DE dataset.} The top row, from left to right: RGB, depth edges GT (manually annotated), depth edges extracted (using Canny) from the baseline and our depth, respectively. The bottom row, from left to right: the predicted depth of the baseline and our method, respectively, followed by two zoom-in crops with the RGB, baseline and our predicted depth, respectively.}
\label{fig:qualitive_ddad}
\end{figure*}


{\begin{table*}[t]
	\begin{center}
    		\begin{tabular}{ | l | c | c | c | c | c | c | c | c | c | c | c |}
    		\hline
				 \textbf{Method} & \multicolumn{4}{c|}{\textbf{DDAD-DE}} & \multicolumn{3}{c|}{\textbf{DDAD test}} & \multicolumn{4}{c|}{\textbf{Synscapes}} \\ \hline
			  & AUC (edges) $\uparrow$ & ORD $\downarrow$ & ARE $\downarrow$ & $\delta_1 \uparrow$  & ORD & ARE & $\delta_1$ & AUC & ORD & ARE & $\delta_1$ \\ \hline \hline	
			\textbf{Packnet-SAN} & 31.52\% (23.32\%) & 8.03\% & 8.89\% & 91.6\% & 8.95\% & 9.49\% & 90.7\% & 61.17\% & 30.8\% & 5.43\% & 96.2\% \\ \hline
			\textbf{Packnet-SAN} & \textbf{48.32\% (32.29\%)} & 8.38\% & 8.99\% & 91.4\% & 9.43\% & 10.0\% & 89.5\% & \textbf{65.38\%} & 21.1\% & 4.85\% & 96.5\% \\
			\textbf{+ EL (ours)} & & & & & & & & & & & \\ \hline
    		\end{tabular}
	\end{center}
	\vspace{-0.5cm}
\caption{\textbf{Results on the DDAD and Synscapes datasets.} The AUC (for DDAD) is given for the range where at least one MDE method has valid measurement: [0.14,0.37]. In parentheses we also report the AUC of the full [0,1] range.}
\label{tab:results_ddad}
\vspace{-0.3cm}	
\end{table*}
}

\vspace{-0.3cm}
\subsubsection{The DDAD dataset}
In Fig.~\ref{fig:qualitive_ddad} we present an image from the DDAD-DE dataset, alongside the depth predictions of Packnet-SAN and our method with Packnet-SAN as baseline. Furthermore, the depth edges (extracted similarly to KITTI), which are laid on top of the RGB images, are presented in the top row. It can be seen that our method has considerably more accurate edges that fit more tightly on objects' silhouette in comparison to the baseline. Moreover, some missing edges are added and some clear false positives are discarded in our method in comparison to the baseline. 

The precision-recall curves for the DDAD-DE dataset are presented in the right side of Fig.~\ref{fig:prec_recall}, where our method, when compared with the same precision (same $x$ coordinate), has significantly higher recall than the Packnet-SAN baseline. In Tab.~\ref{tab:results_ddad}, the AUC metric for the edge quality is presented, where our method achieves an AUC of $48.32\%$ in comparison to Packnet-SAN that achieves an AUC of $31.52\%$, indicating around $50\%$ of relative improvement.

The quality of the per-pixel depth is also evaluated on the DDAD-DE dataset, and the ARE is presented in Tab.~\ref{tab:results_ddad}. The Packnet-SAN baseline and our methods achieve an ARE of $8.89\%$ and $8.99\%$, respectively, which indicates a comparable performance. On the standard DDAD evaluation set (Tab.~\ref{tab:results_ddad}), the ARE of the Packnet-SAN and our method are $9.49\%$ and $10.0\%$, respectively, indicating, a~$5\%$ decrease in per-pixel depth accuracy of our method.

\vspace{-0.2cm}
\subsubsection{The Synscapes dataset}
We present the results of an additional experiment on the Synscapes dataset in Tab.~\ref{tab:results_ddad} and in the supp.~material, where we show that an increase in the depth edges accuracy yields an increase in the per-pixel depth accuracy (e.g., ARE) for dense depth GT. It further suggests that the slight decrease in the per-pixel depth accuracy, which is sometimes observed in KITTI and DDAD, might be due to the sparseness of their depth GT, especially near depth edges. 


\subsection{Comparison to Merging Methods \cite{miangoleh2021boosting, dai2023multi}}
\label{sec:miangoleh}
\vspace{-0.2cm}
To the best of our knowledge there are no other methods to improve the depth edges of sparsely-supervised MDE methods; however, BoostingDepth \cite{miangoleh2021boosting} and GradientFusion \cite{dai2023multi} share a similar aim to ours, even though these methods were designed for indoor scenes with dense GT. To this end, we use the predictions of Packnet-SAN with BoostingDepth's original depth merger for comparison with our method. Additionally, we train the depth merger on the KITTI dataset for a fair comparison (see details in the supplementary material). The fusion-net in GradientFusion was trained solely on the high-res HR-WSI dataset \cite{xian2020structure}. Their method is based on a preprocessing of HR-WSI's images with a guided filter to produce high and low resolutions that allow the self-supervised training. We therefore use their original fusion-net that was used for all datasets and MDE methods. Two experiments are presented: one with LeRes \cite{yin2021learning}, which was the main backbone for GradientFusion (see supp.~material), and one with Packnet-SAN. 

For both BoostingDepth (with the original weights) and GradientFusion, the AUC metric presented in Tab.~\ref{tab:results_kitti} and Fig.~\ref{fig:prec_recall} shows a slightly worse performance than the Packnet-SAN baseline, and a significantly worse performance than Packnet-SAN with our method. One possible reason for the performance is probably due to the domain gap between the datasets used to train the depth mergers and KITTI. Additional reasons for BoostingDepth might be: (i) The depth merger can process only square images, which deviates significantly from KITTI's aspect ratio, and (ii) It is highly sensitive to the network's Receptive Field (RF), where Packnet-SAN has a large RF of 1028, close to the size of the image. The ARE performance is significantly worse than both other methods, probably due to the usage of the GAN-based pix2pix \cite{isola2017image} to train the depth merger which is known to create results that 'look' realistic but not necessarily accurate. The performance of GradientFusion for KITTI is low since self-supervision using the guided filter strongly depends on image resolution (as stated in their paper also), which is low in datasets like KITTI.

BoostingDepth with the depth merger that was trained on KITTI presents worse edges (AUC) than the original weights, but better ARE. We hypothesize that the basic assumption of BoostingDepth, in which an MDE network produces more fine details in high resolution (but worse overall shape) breaks for LIDAR-based scenes due to the lack of depth GT near depth edges. See supp.~material for a more detailed discussion, visual results and training details. 


\vspace{-0.1cm}
\subsection{Application: AR occlusions}
We present an experiment in rendering virtual objects for AR applications. We use the predicted depth of the Packnet-SAN baseline and our method to compute the occluded parts of the virtual objects from the point of view of the camera. In Fig.~\ref{fig:fig1}b three animated characters are planted in the scene. To observe the difference in the depth edges accuracy, zoom-in on the sides of the vehicle and on the 'no-entry' post. The occlusions of the character in our method seems much more realistic than the baseline. 

\vspace{-0.1cm}
\section{Conclusion}
\label{sec:discussion}
\vspace{-0.2cm}
We have presented a method to improve the localization of the depth edges in sparsely-supervised MDE methods, while preserving the per-pixel depth accuracy. The method is based on the observation that detecting the location of the depth edges can be learned effectively from synthetic data. 
To evaluate our method on real data, we introduced two depth edges evaluation sets for the KITTI and the DDAD datasets by manual annotation. 
The proposed method can be further improved by considering the images (and LIDAR) of the target dataset (e.g., KITTI) in the training procedure, possibly using methods from the unsupervised domain adaptation literature. 


\section{Supplementary Material for Mind The Edge: Refining Depth Edges in Sparsely-Supervised Monocular Depth Estimation}

In the following section we provide additional details, examples and results for the experiments described in the main paper.

\subsection{The Synscapes dataset}
To simulate a real oudoor dataset with LIDAR supervision, we sample pixels from the dense depth GT in a typical LIDAR pattern, similarly to the KITTI dataset, with 64 vertical beams and an horizontal beam density of 0.09 degrees between beams. Since this is a na\"ive approach we follow the LIDAR density, presented in Fig.~3 of the main paper, and randomly remove LIDAR samples such that the resulting distribution is similar to KITTI's. Note, however, that the true LIDAR spatial distribution in KITTI is more complex than our simulation since large continuous areas lack LIDAR samples in KITTI, in contrast to our simulation. We hypothesize that a more realistic simultion would result in worse depth edges of the baseline, increasing the depth edges performance gap from our method. 

\begin{figure}
\centering
\includegraphics[width=0.98\linewidth]{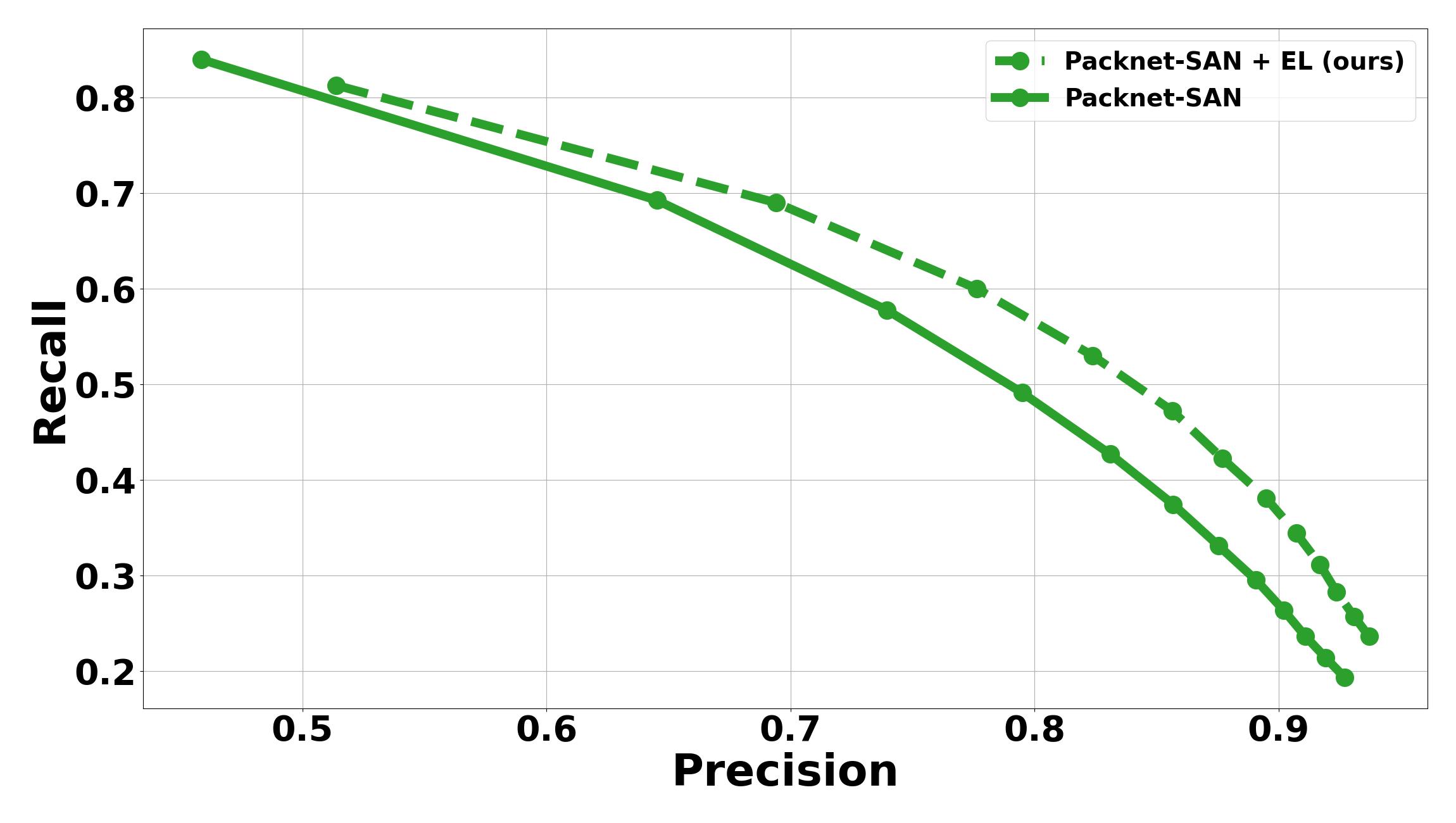}
\caption{Precision and recall of the depth edges of the baseline vs. our method for Packnet-SAN on the Synscapes dataset.}
\label{fig:prec_recall_synscapes}
\end{figure}

In Fig.~\ref{fig:qualitive_synscapes} we present two images from the Synscapes dataset alongside the depth prediciton of the baseline and our method (with Packnet-SAN). It can be seen that although the Packnet-SAN baseline is has mostly accurate edges, our method still outperforms it in some parts of the image. The quantitative results are presented in the main paper (Tab.~3) and in the bottom of Fig.~\ref{fig:prec_recall_synscapes}, which demonstrate a small, but consistent improvement in the depth edges quality. Importantly, as argued in the main paper, the per-pixel metrics, e.g., ARE, of our method is significantly better, which suggests that when dense depth is available, the improvement in the edge quality is translated into an improvement in the per-pixel depth metrics. 

\begin{figure*}
\centering
\begin{tabular}{cc}
~~~~~~~~~~~~~~~~~~~~~~~~~~~~~~~~~~~~~~~~~~~~~~~~~~~~~~~~~~~~~~~~~~~~~~~~~~~~~~~~~~~~~~~~~~~~~~~~~~~~~ RGB ~~~~~~~~~~~~~~ Standard Gradient ~~ Orthogonal Gradient \\
   \includegraphics[width=0.44\linewidth]{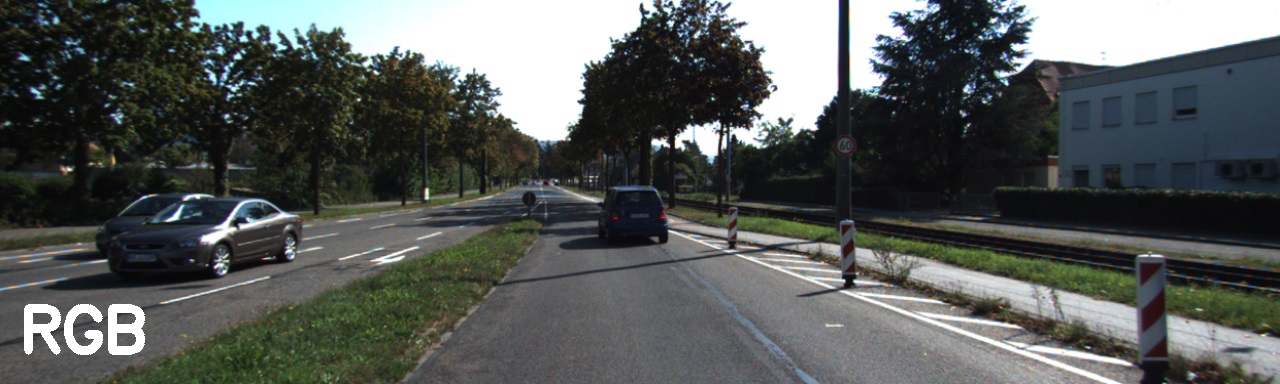} 
   \includegraphics[width=0.17\linewidth]{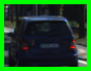} 
   \includegraphics[width=0.17\linewidth]{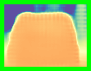} 
   \includegraphics[width=0.17\linewidth]{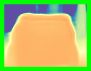}  \\
   \includegraphics[width=0.48\linewidth]{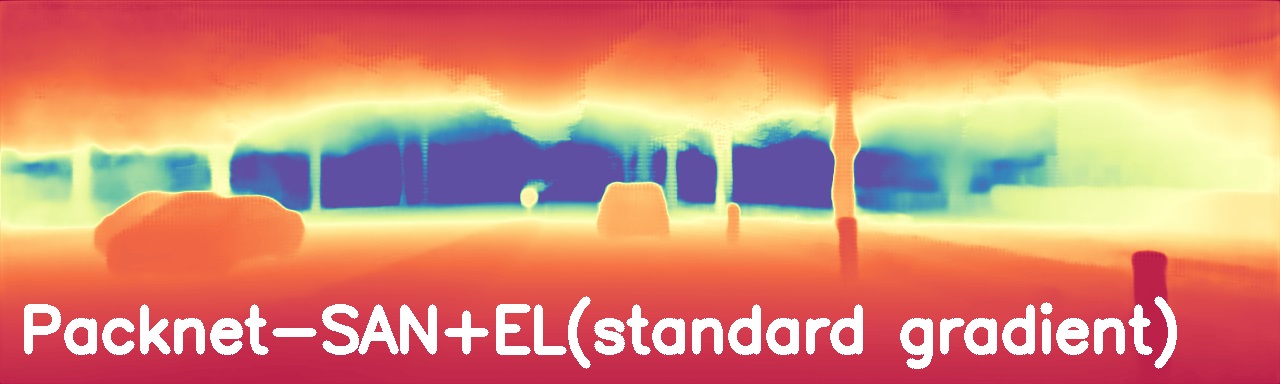} 
   \includegraphics[width=0.48\linewidth]{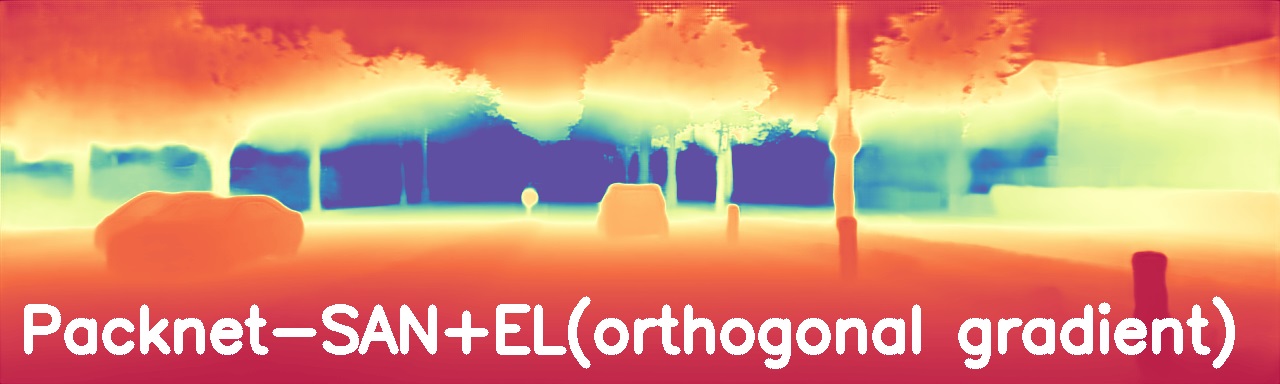} 
\end{tabular}
\caption{The standard spatial image gradient (named 'standard') and our proposed orthogonal to the edges spatial gradient (named 'orthogonal'). See Sec.~3.2 in the main paper for details. }
\label{fig:depth_edge_loss}
\end{figure*}

\begin{figure*}[t]
\centering
\begin{tabular}{cc}
KITTI-DE & DDAD-DE \\
\includegraphics[width=0.48\linewidth]{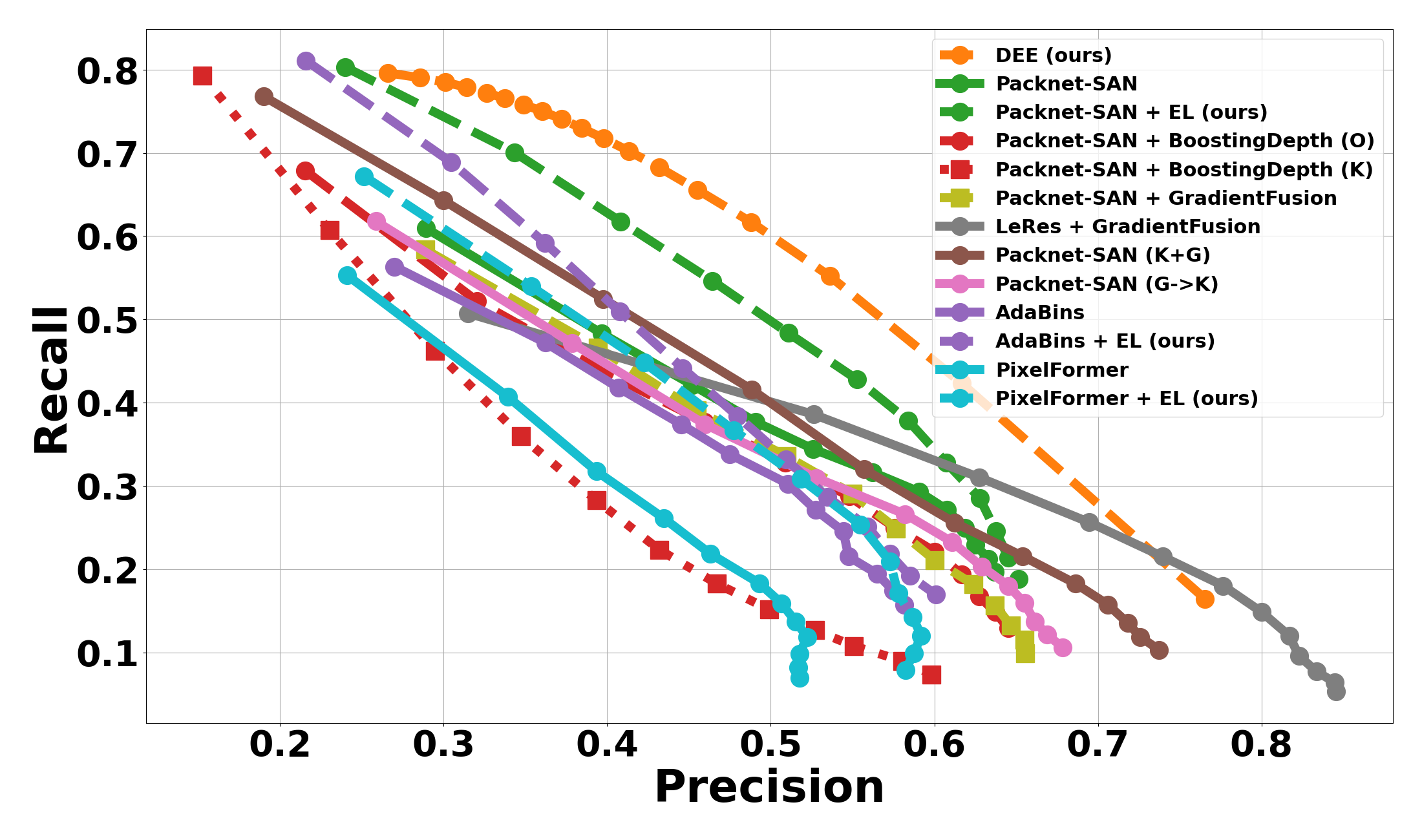} &
\includegraphics[height=4.95cm, width=0.48\linewidth]{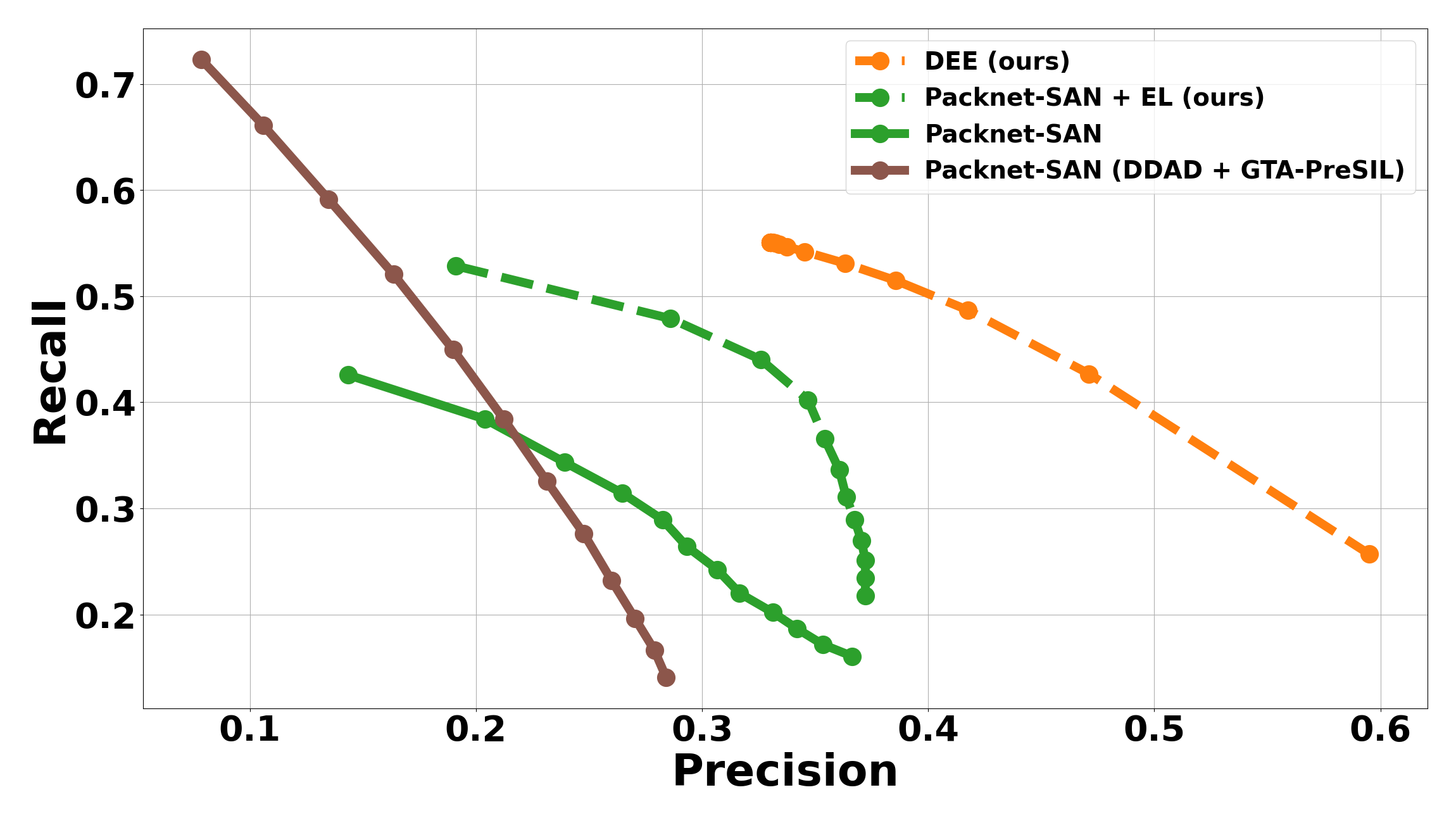}
\end{tabular}
\caption{\textbf{Precision and recall of the depth edges on the KITTI-DE and DDAD-DE evaluation sets}. Each of the points on the graphs that correspond to an MDE method is generated with different parameters of the Canny edge detector. Each of the points on the graphs that correspond to the DEE method is generated by thresholding the depth edge probability in the range $(0,1)$.}
\label{fig:prec_recall}
\end{figure*}

\subsection{Simultaneous Training on Source and Target}
In Fig.~\ref{fig:prec_recall}, Tab.~\ref{tab:results_kitti} and Tab.~\ref{tab:results_ddad} we report an experiment where we trained Packnet-SAN simoultaneously on the source (GTA-PreSIL) and the target (KITTI or DDAD) with the depth loss. Both in KITTI (denoted as Packnet-SAN (K+G)) and in DDAD (denoted as Packnet-SAN (D+G)) the edge metric (AUC) is similar or slightly better than the baseline trained on KITTI or DDAD alone. However, the depth metrics (ARE) are significantly worse, often twice as worse. We conclude that our method is a significantly better alternative than simoultaneous training. Furthermore, we note that simoultaneous training is also more time consuming since for each target dataset the source has to be trained as well. In addition, training on GTA-PreSIL, followed by training on KITTI does not yield good performance, where the edge and ARE metrics are slightly worse than the Packnet-SAN baseline.

\subsection{Using RGB Edges Instead of Depth Edges}
\begin{figure}[h]
\centering
\begin{tabular}{c}
  \includegraphics[width=0.90\linewidth]{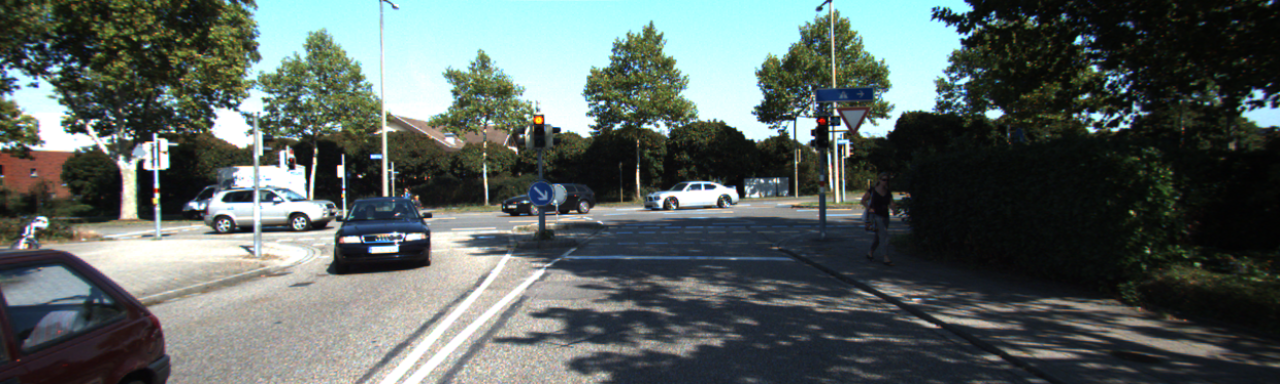}  \\
  \includegraphics[width=0.90\linewidth]{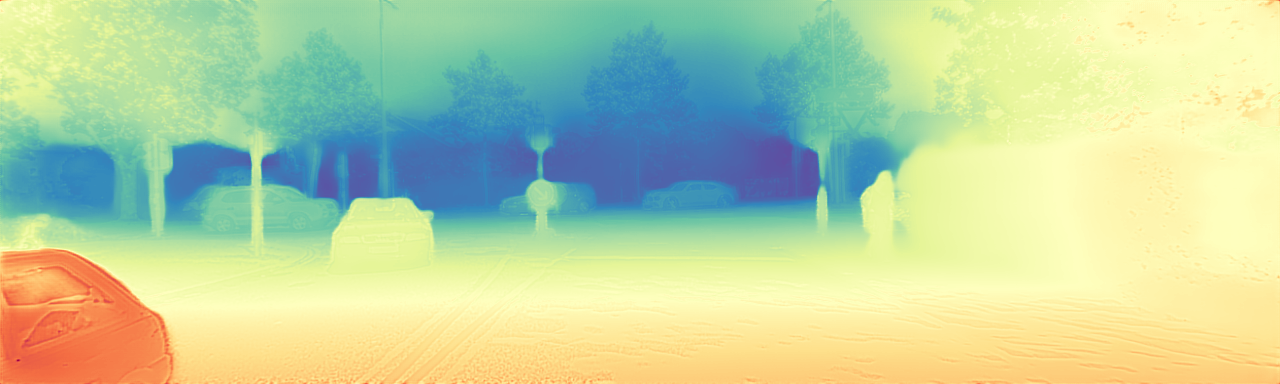}
\end{tabular}
\vspace{-0.3cm}
\caption{Training on RGB edges instead of depth edges.}
\label{fig:rgb_edges}
\vspace{-0.3cm}
\end{figure}

A potential simple alternative for the use of the depth edges, obtained from the DEE network, is to use image edges, obtained using an edge detector running directly on the RGB images. We used Canny edge detector (empirically set thresholds to 120 and 160) to extract multiscale edges and normals, which were used to train Packnet on KITTI with edge loss, instead of using our DEE output. The quantitative results using Packnet are 4.61\% ARE (worse than baseline and ours) and 49.28\% (42.09\%) AUC (slightly better than the baseline but significantly worse than ours) on the KDE dataset. These results emphasize the importance of the DEE network, as edges from the RGB are highly noisy and their correlation to the depth edges is not high enough (see false edges in Fig.~\ref{fig:rgb_edges}).

\vspace{-0.2cm}

\subsection{Virtual Human Insertion for Data Aug.} 
\begin{figure}[h]
\centering
\begin{tabular}{c}
  \includegraphics[width=0.48\linewidth]{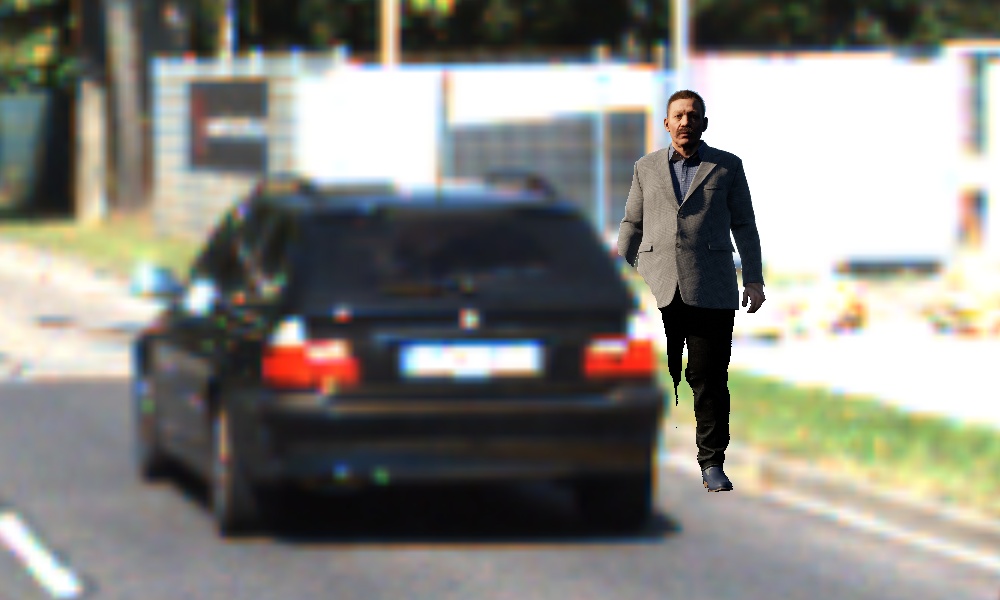} 
  \includegraphics[width=0.48\linewidth]{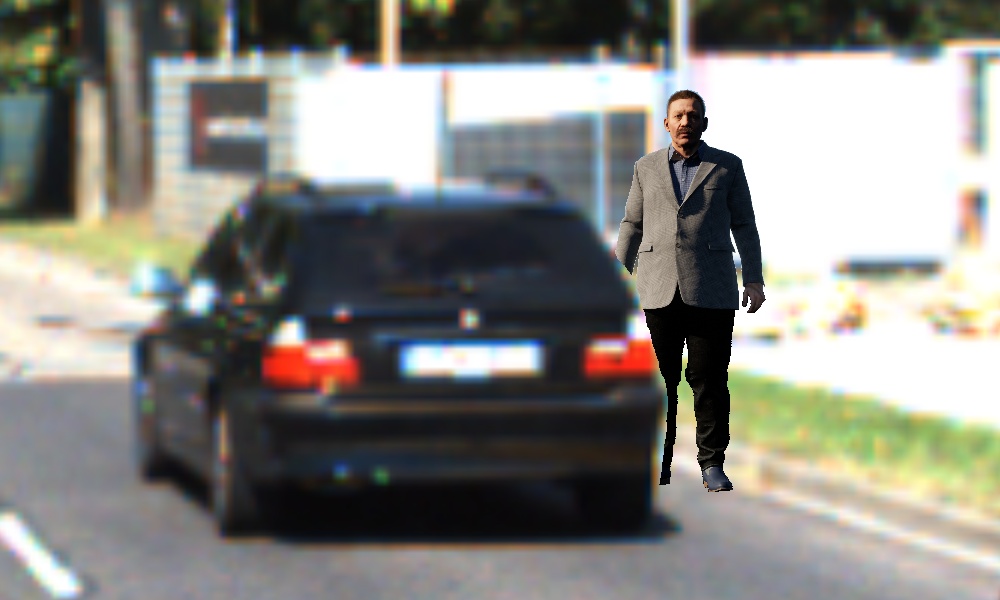}
\end{tabular}
\vspace{-0.3cm}
\caption{Virtual human insertion. Left: Packnet-SAN (baseline), Right: Packnet-SAN + EL (ours).}
\label{fig:human}
\vspace{-0.3cm}
\end{figure}

We demonstrate another potential use for our method - inserting virtual humans to automotive images, potentially for data augmentation. In Fig.~\ref{fig:human} we present a virtual human inserted to the scene, where in our method the occlusion is significantly more realistic than the baseline.

\vspace{-0.2cm}

\renewcommand{\tabcolsep}{4pt}
{\begin{table*}[t]
	\begin{center}
    		\begin{tabular}{ | l | c | c | c | c | c | c | c |}
    		\hline
				 \textbf{Method} & \multicolumn{4}{c|}{\textbf{KITTI-DE}} & \multicolumn{3}{c|}{\textbf{KITTI test}} \\ \hline
			  & AUC (edges) $\uparrow$ & ORD $\downarrow$ & ARE $\downarrow$ & $\delta < 1.25 \uparrow$  & ORD $\downarrow$ & ARE $\downarrow$ & $\delta < 1.25 \uparrow $ \\ \hline \hline	
			\textbf{Packnet-SAN} & 47.56\% (39.40\%) & 7.68\% & 3.45\% & 98.66\% & 12.40\% & 6.17\% & 95.39\% \\ \hline
			\textbf{Packnet-SAN (K+G)} & 53.55\% (41.83\%) & 8.81\% & 6.01\% & 95.83\% & 11.78\% & 8.90\% & 91.04\% \\ \hline
			\textbf{Packnet-SAN (G $\rightarrow$ K)} & 45.56\% (35.39\%) & 8.52\% & 5.26\% & 96.39\% & 11.29\% & 8.02\% & 92.28\% \\ \hline
			\textbf{Packnet-SAN + BoostingDepth (O)} & 46.04\% (37.07\%) & 10.35\% & 9.32\% & 88.90\% & 12.63\% & 11.10\% & 86.41\% \\  \hline
			\textbf{Packnet-SAN + BoostingDepth (K)} & 36.19\% (31.27\%) & 9.47\% & 7.24\% & 93.62\% & 11.45\% & 8.33\% & 91.99\% \\  \hline 
			\textbf{Packnet-SAN + GradientFusion} & 44.51\% (34.10\%) & 9.18\% & 5.93\% & 95.66\% & 11.15\% & 7.18\% & 94.17\% \\  \hline 
			\textbf{LeRes + GradientFusion} & 44.59\% (34.53\%) & 12.02\% & 17.26\% & 71.26\% & 12.80\% & 15.76\% & 76.12\% \\  \hline
			\textbf{Packnet-SAN + EL (ours)} & \textbf{61.87\% (49.02\%)} & 7.75\% & 3.61\% & 98.53\% & 12.48\% & 6.50\% & 95.06\% \\\hline \hline 
			\textbf{AdaBins} & 41.23\% (34.11\%) & 7.69\% & 3.14\% & 98.78\% & 10.14\% & 6.28\% & 95.85\% \\ \hline 
			\textbf{AdaBins + EL (ours)} & \textbf{53.47\% (44.00\%)} & 7.64\% & 3.11\% & 98.79\% & 10.13\% & 6.21\% & 95.87\% \\  \hline \hline
			\textbf{PixelFormer} & 32.79\% (26.44\%) & 7.47\% & 3.00\% & 98.79\% & 7.56\% & 5.45\% & 96.98\% \\ \hline 
			\textbf{PixelFormer + EL (ours)} & \textbf{46.23\% (35.33\%)} & 7.53\% & 2.94\% & 98.80\% & 7.58\% & 5.59\% &  96.72\% \\  \hline
    		\end{tabular}
	\end{center}
\caption{\textbf{Results on the KITTI dataset.} The AUC is given for the range where at least one MDE method has valid measurement: [0.12,0.65]. In parentheses we also report the AUC of the full [0,1] range. In BoostingDepth, O is for the original training (dense data) by the authors, and K is for our training (KITTI data). K+G and G $\rightarrow$ K stand for simultaneous training on KITTI and GTA-PreSIL, and training on GTA-PreSIL followed by training on KITTI.}
\label{tab:results_kitti}
\end{table*}
}

\renewcommand{\tabcolsep}{4pt}
{\begin{table*}[t]
	\begin{center}
    		\begin{tabular}{ | l | c | c | c | c | c | c | c |}
    		\hline
				 \textbf{Method} & \multicolumn{4}{c|}{\textbf{DDAD-DE}} & \multicolumn{3}{c|}{\textbf{DDAD test}} \\ \hline
			  & AUC (edges) $\uparrow$ & ORD $\downarrow$ & ARE $\downarrow$ & $\delta < 1.25 \uparrow$  & ORD $\downarrow$ & ARE $\downarrow$ & $\delta < 1.25 \uparrow $ \\ \hline \hline	
			\textbf{Packnet-SAN} & 31.52\% (23.32\%) & 8.03\% & 8.89\% & 91.62\% & 8.95\% & 9.49\% & 90.7\% \\ \hline
			\textbf{Packnet-SAN (D + G)} & 29.89\% (25.49\%) & 10.37\% & 14.09\% & 83.14\% & 12.29\% & 16.74\% & 78.09\% \\ \hline
			\textbf{Packnet-SAN + EL (ours)} & \textbf{48.32\% (32.29\%)} & 8.38\% & 8.99\% & 91.44\% & 9.43\% & 10.0\% & 89.5\% \\\hline 
    		\end{tabular}
	\end{center}
	\vspace{-0.1cm}
\caption{\textbf{Results on the DDAD dataset.} The AUC is given for the range where at least one MDE method has valid measurement: [0.14,0.37]. In parentheses we also report the AUC of the full [0,1] range. D+G stands for simultaneous training on DDAD and GTA-PreSIL.}
\label{tab:results_ddad}
\vspace{-0.1cm}	
\end{table*}
}

\begin{figure}
\centering
\includegraphics[width=0.98\linewidth]{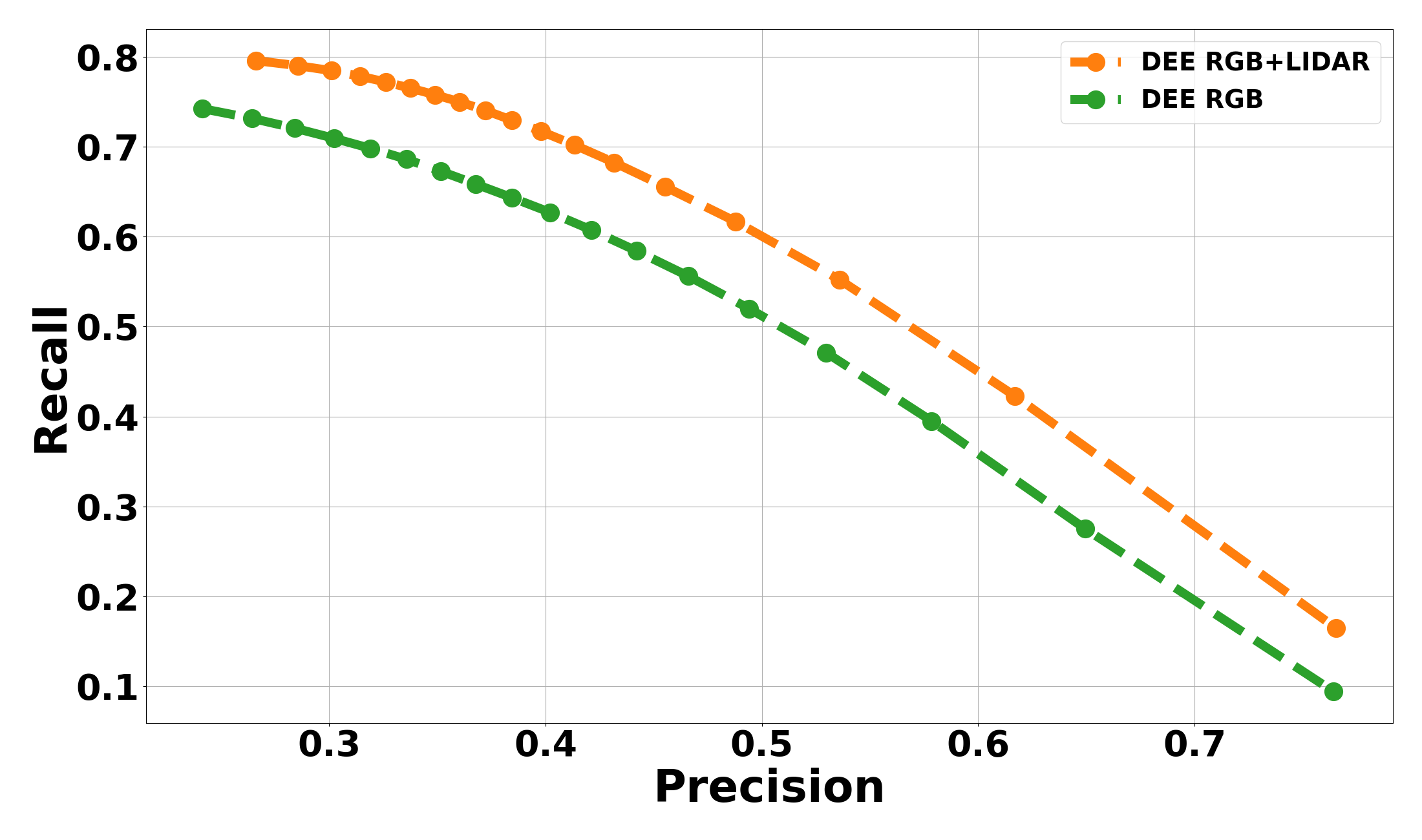}
\caption{Precision and recall of the DEE model with RGB only and RGB+LIDAR inputs, inffered on the KITTI-DE dataset.}
\label{fig:prec_recall_rgb_rgb_lidar}
\end{figure}

\begin{figure}
\centering
\includegraphics[width=0.98\linewidth]{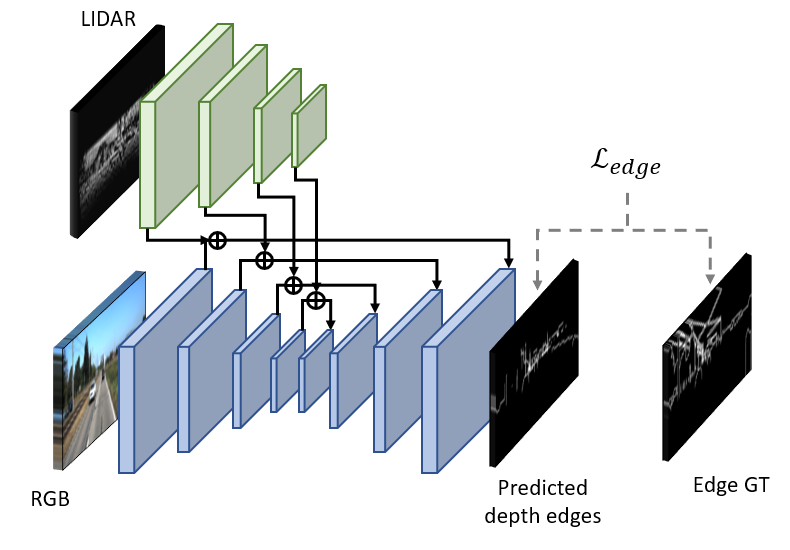}
\caption{The architecture of the DEE model.}
\label{fig:dee}
\end{figure}

\begin{figure*}[t]
\centering
\begin{tabular}{ccc}
~~~~~~~~~~~~~~~~~~~~~~~~~~~~~~~~~~~~~~~~~~~~~~~~~~~~~~~~~~~~ RGB ~~~~~~~~~~~~~~~~~~~~~~~~~~~~~~ Baseline ~~~~~~~~~~~~~~~~~~~~~~~~~~~ Ours \\
   \includegraphics[width=0.32\linewidth]{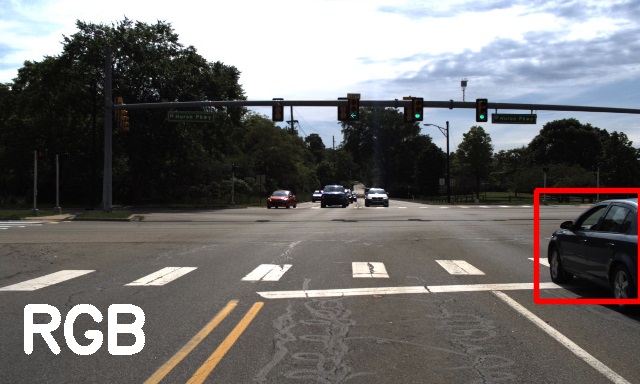} 
   \includegraphics[width=0.2\linewidth]{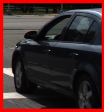} 
   \includegraphics[width=0.2\linewidth]{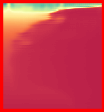} 
   \includegraphics[width=0.2\linewidth]{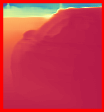} \\
   \includegraphics[width=0.465\linewidth]{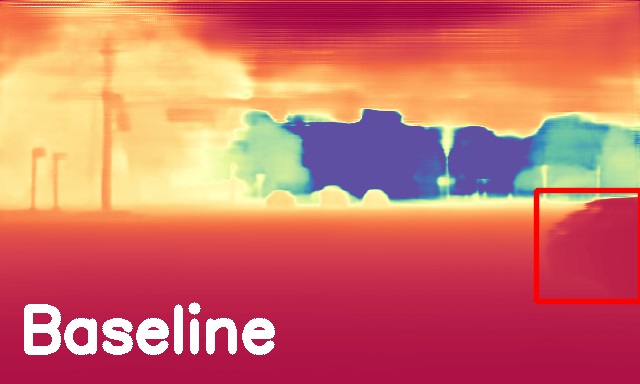} 
   \includegraphics[width=0.465\linewidth]{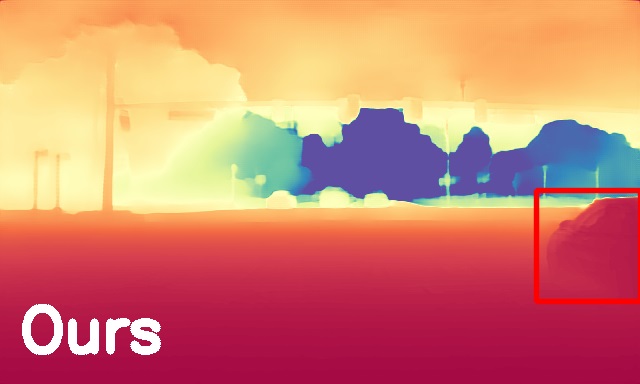} \\
   \includegraphics[width=0.465\linewidth]{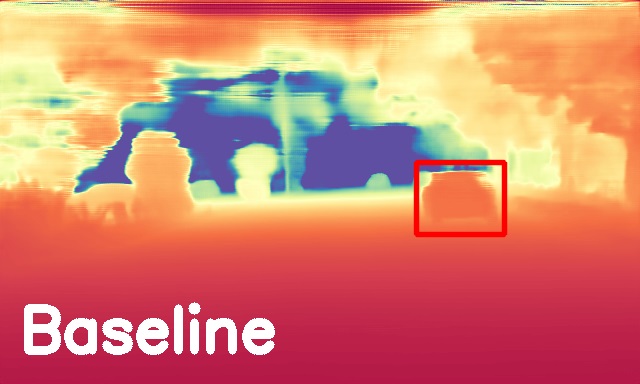} 
   \includegraphics[width=0.465\linewidth]{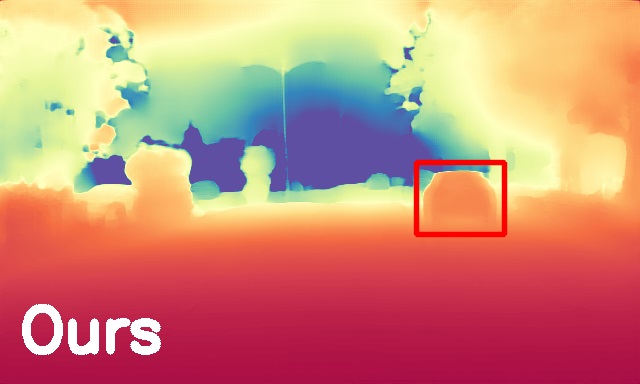} \\
   \includegraphics[width=0.287\linewidth]{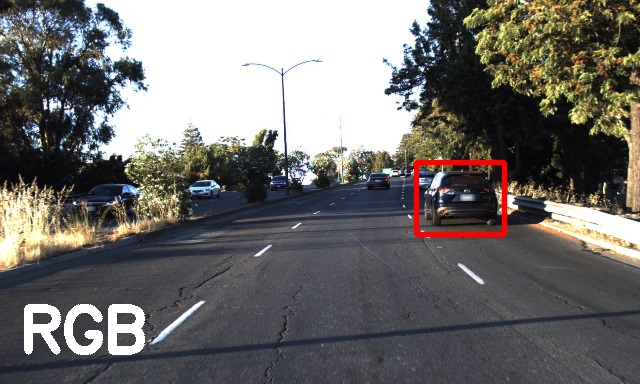} 
   \includegraphics[width=0.211\linewidth]{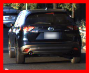} 
   \includegraphics[width=0.211\linewidth]{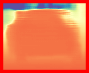} 
   \includegraphics[width=0.211\linewidth]{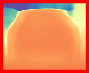} \\
~~~~~~~~~~~~~~~~~~~~~~~~~~~~~~~~~~~~~~~~~~~~~~~~~~~~~~~~~~~ RGB ~~~~~~~~~~~~~~~~~~~~~~~~~~~~~~ Baseline ~~~~~~~~~~~~~~~~~~~~~~~~~~~~~~ Ours
\end{tabular}
\caption{Examples of depth predictions of Packnet-SAN and Packnet-SAN + EL (ours) of images from the DDAD-DE dataset - Part I.}
\label{fig:qualitive_ddad_1}
\end{figure*}

\begin{figure*}[t]
\centering
\begin{tabular}{ccc}
~~~~~~~~~~~~~~~~~~~~~~~~~~~~~~~~~~~~~~~~~~~~~~~~~~~~~~~~~~~~ RGB ~~~~~~~~~~~~~~~~~~~~~~~~~~~~~~ Baseline ~~~~~~~~~~~~~~~~~~~~~~~~~~~ Ours \\
   \includegraphics[width=0.32\linewidth]{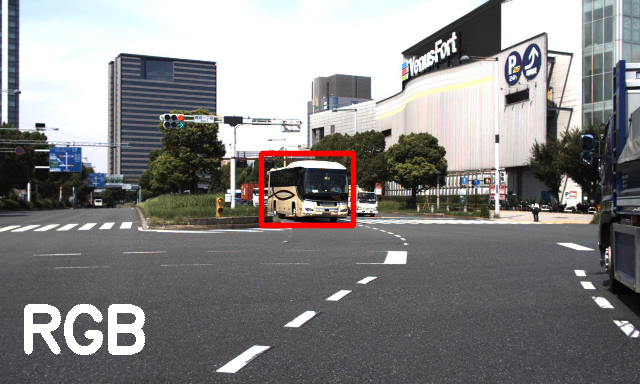} 
   \includegraphics[height=3.43cm,width=0.2\linewidth]{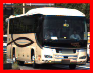} 
   \includegraphics[height=3.43cm,width=0.2\linewidth]{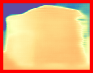} 
   \includegraphics[height=3.43cm,width=0.2\linewidth]{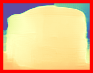} \\
   \includegraphics[width=0.465\linewidth]{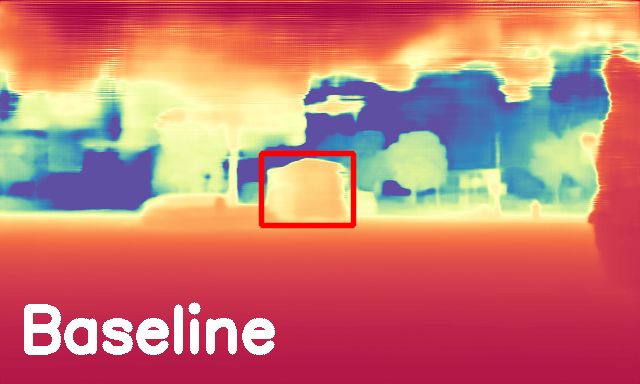} 
   \includegraphics[width=0.465\linewidth]{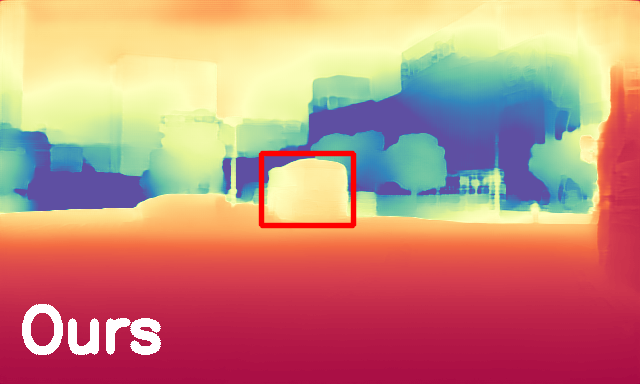} \\
   \includegraphics[width=0.465\linewidth]{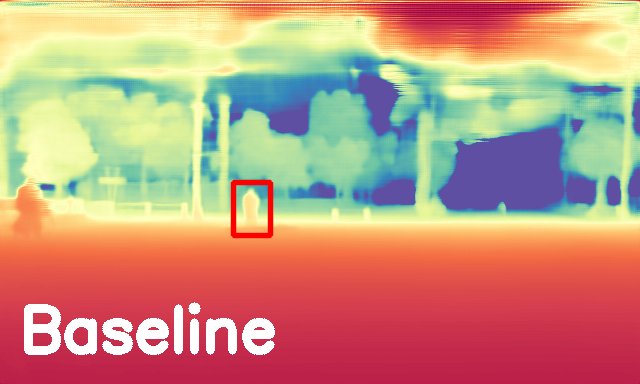} 
   \includegraphics[width=0.465\linewidth]{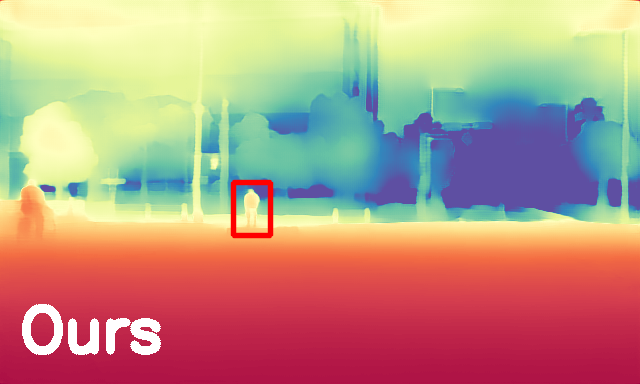} \\
   \includegraphics[width=0.4\linewidth]{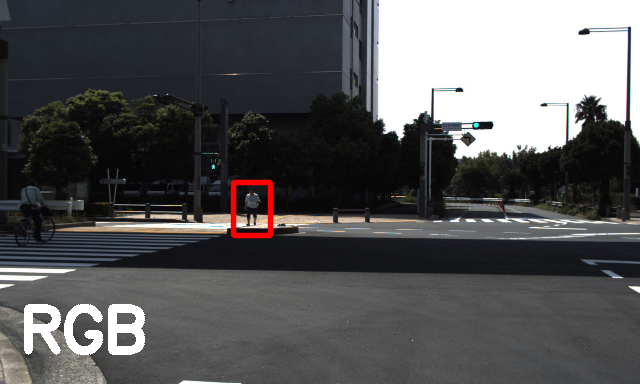} 
   \includegraphics[width=0.173\linewidth]{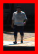} 
   \includegraphics[width=0.173\linewidth]{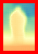} 
   \includegraphics[width=0.173\linewidth]{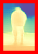} \\
~~~~~~~~~~~~~~~~~~~~~~~~~~~~~~~~~~~~~~~~~~~~~~~~~~~~~~~~~~~~~~~~~~~~~~~~~~~~~ RGB ~~~~~~~~~~~~~~~~~~~~~~ Baseline ~~~~~~~~~~~~~~~~~~~~~~~ Ours
\end{tabular}
\caption{Examples of depth predictions of Packnet-SAN and Packnet-SAN + EL (ours) of images from the DDAD-DE dataset - Part II.}
\label{fig:qualitive_ddad_2}
\end{figure*}

\begin{figure*}[t]
\centering
\begin{tabular}{ccc}
   \includegraphics[width=0.3\linewidth]{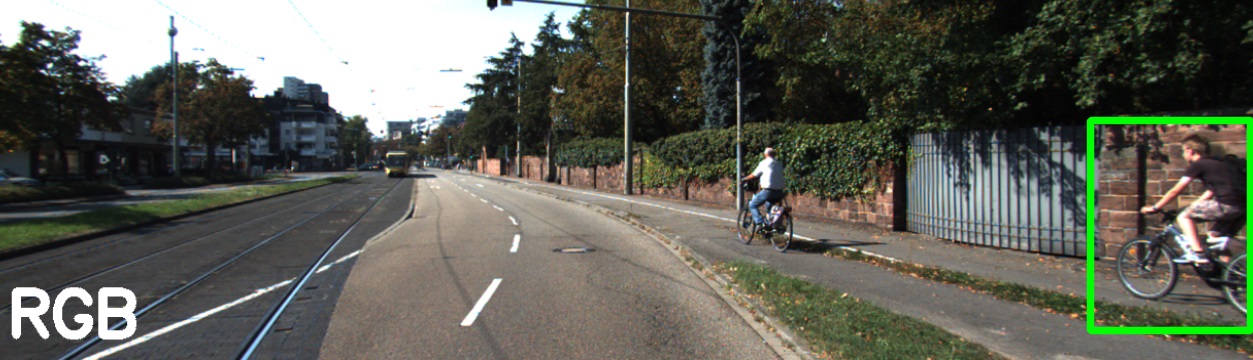} 
   \includegraphics[width=0.3\linewidth]{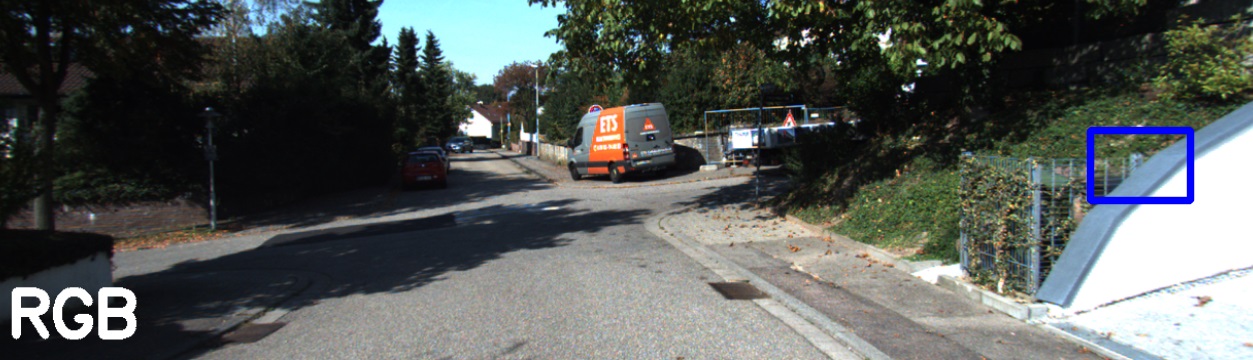} 	
   \includegraphics[width=0.3\linewidth]{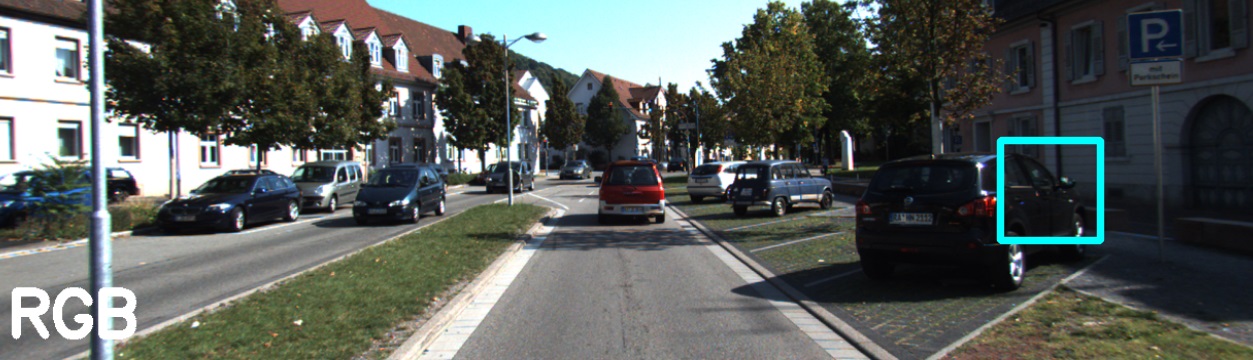} \\

   \includegraphics[width=0.3\linewidth]{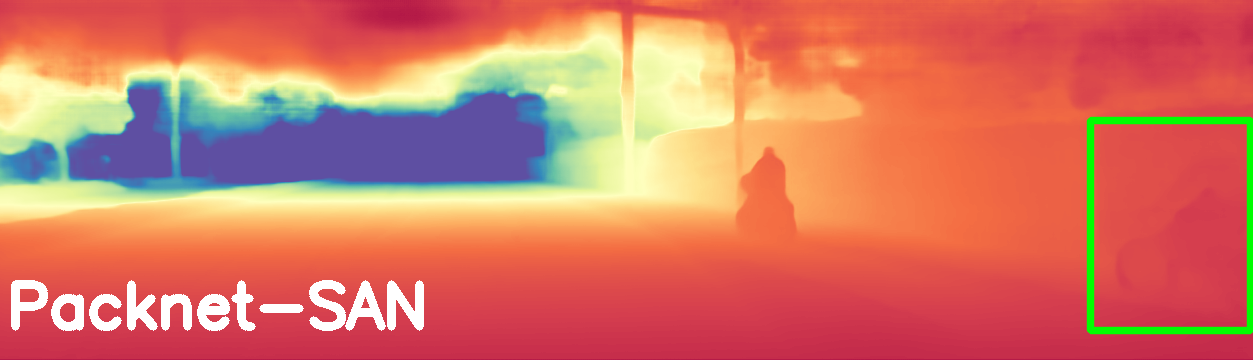} 
   \includegraphics[width=0.3\linewidth]{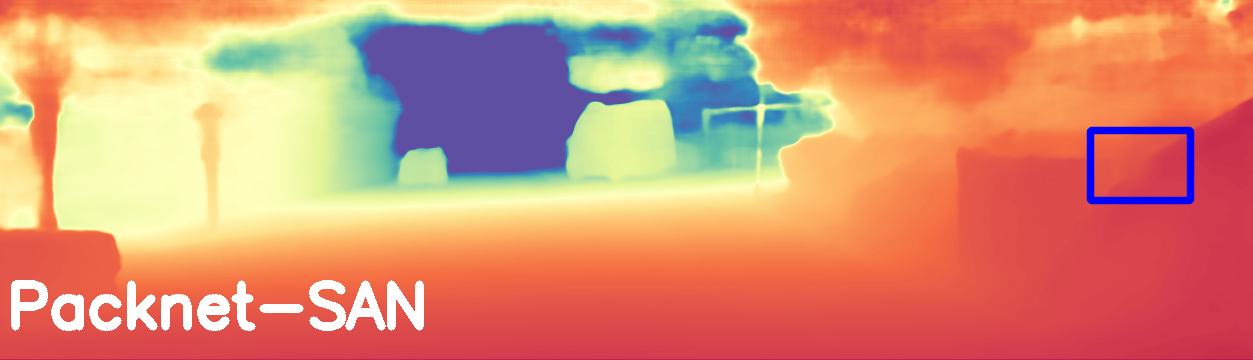}
   \includegraphics[width=0.3\linewidth]{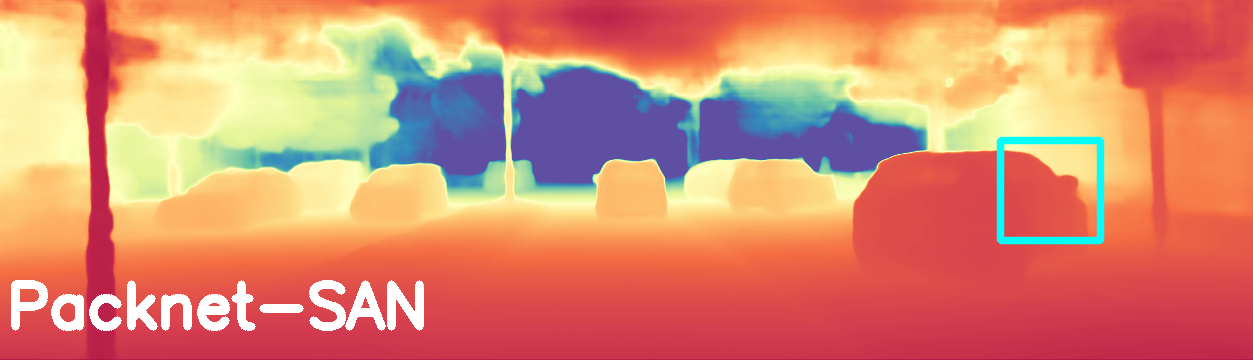} \\
   \includegraphics[width=0.3\linewidth]{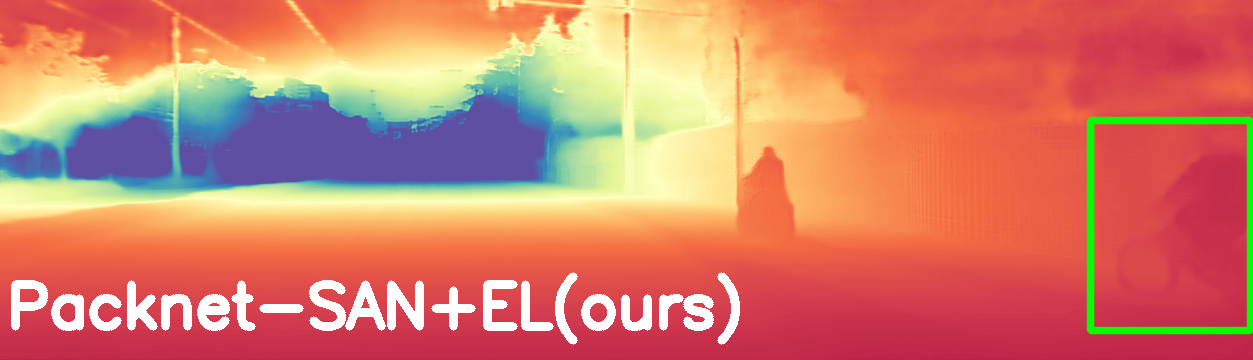} 
   \includegraphics[width=0.3\linewidth]{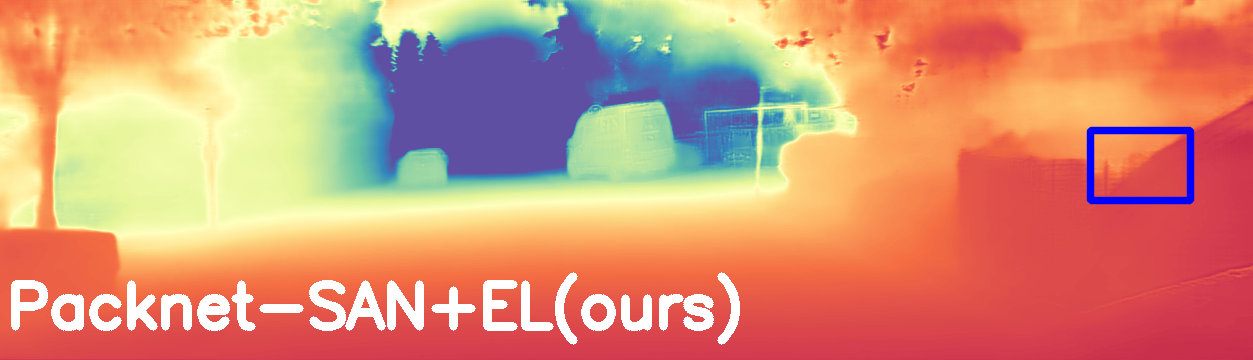}
   \includegraphics[width=0.3\linewidth]{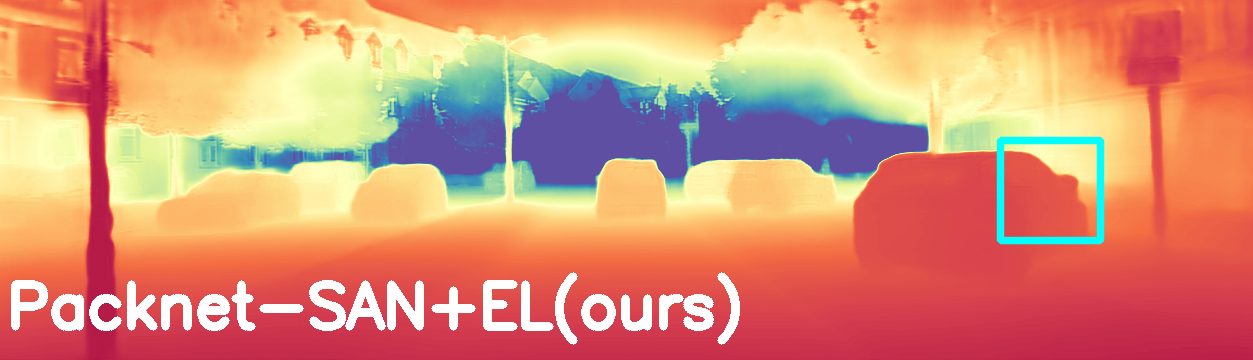} \\
   \includegraphics[width=0.3\linewidth]{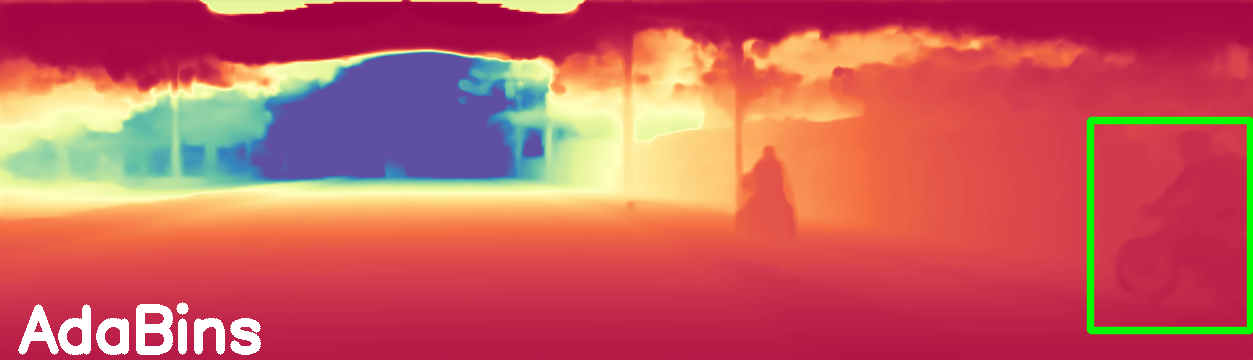} 
   \includegraphics[width=0.3\linewidth]{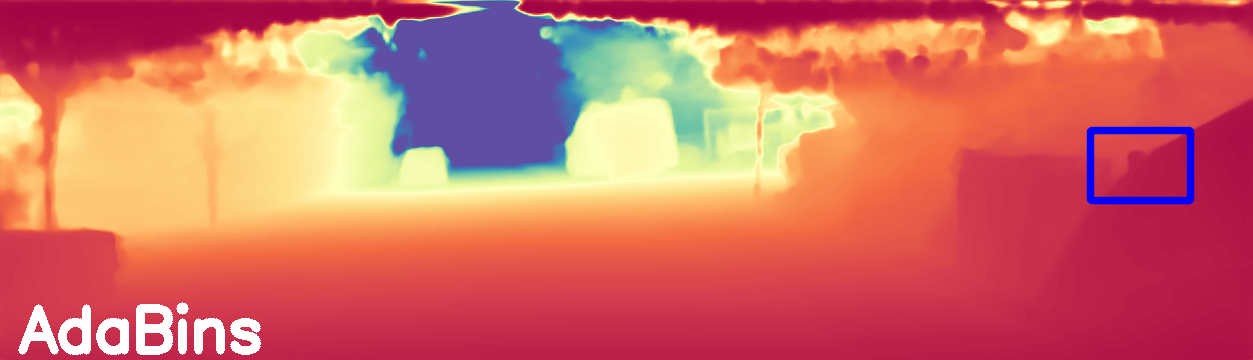} 
   \includegraphics[width=0.3\linewidth]{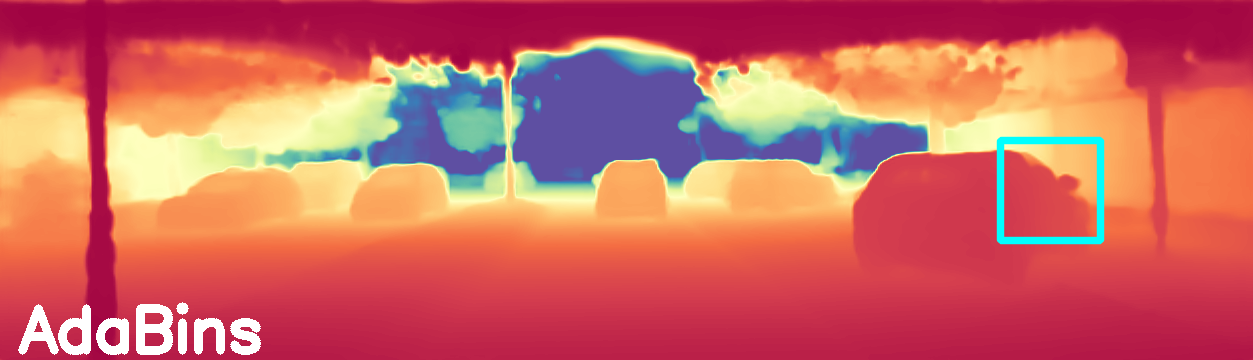} \\
   \includegraphics[width=0.3\linewidth]{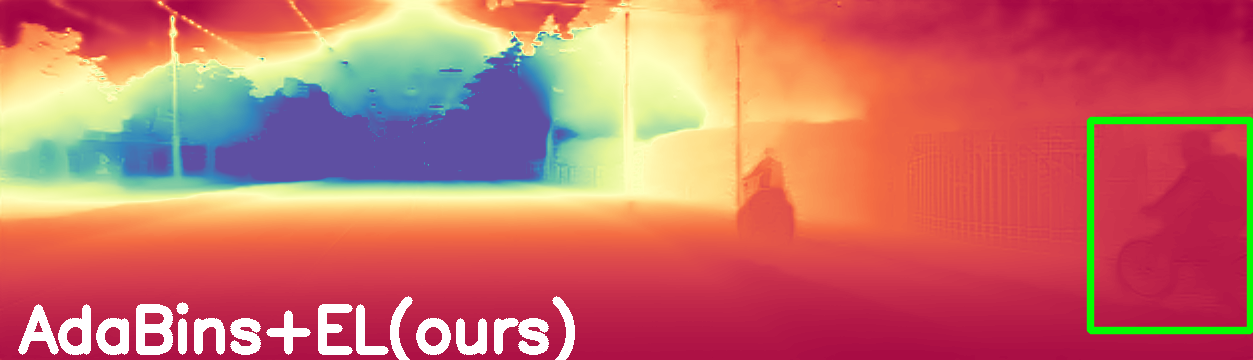}
   \includegraphics[width=0.3\linewidth]{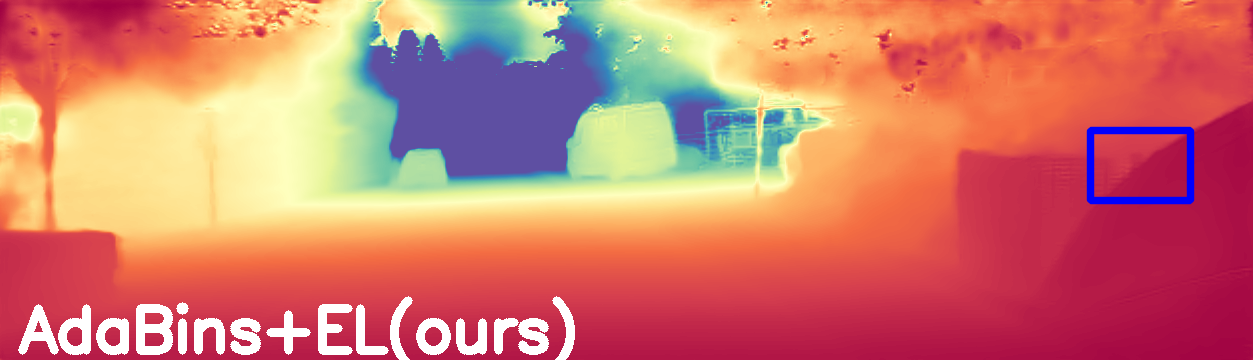}
   \includegraphics[width=0.3\linewidth]{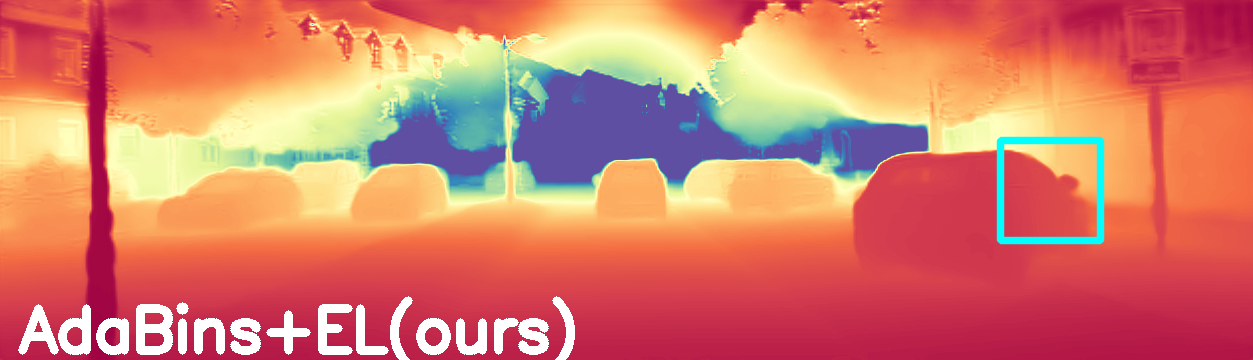} \\
   \includegraphics[width=0.3\linewidth]{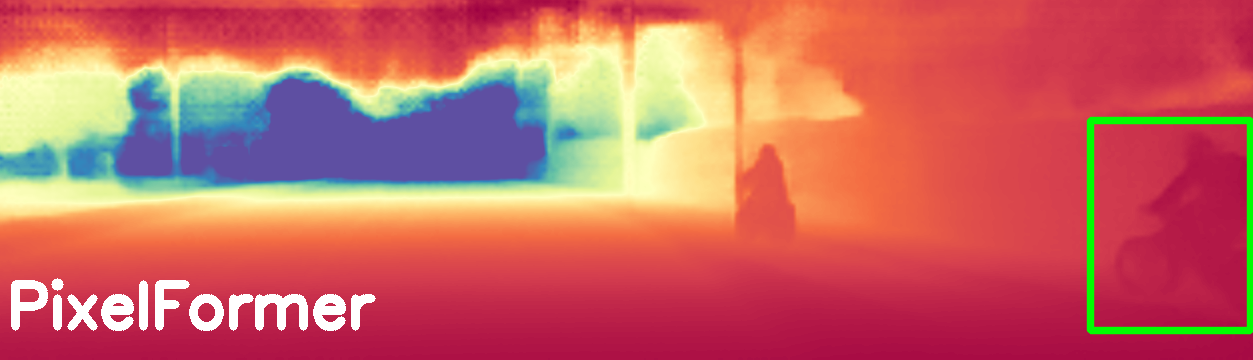}
   \includegraphics[width=0.3\linewidth]{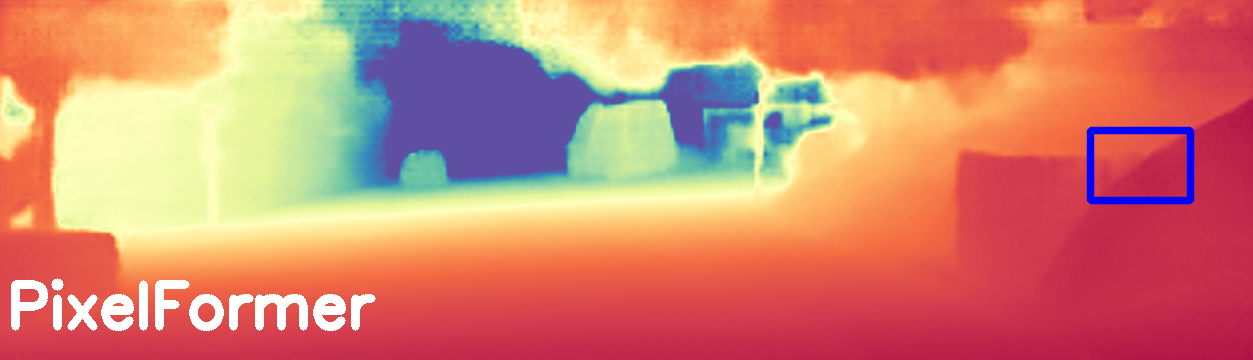}
   \includegraphics[width=0.3\linewidth]{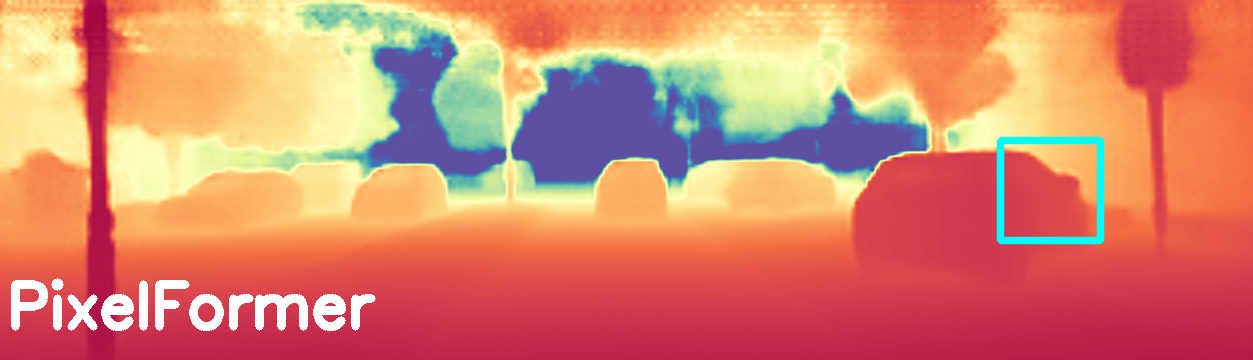} \\
   \includegraphics[width=0.3\linewidth]{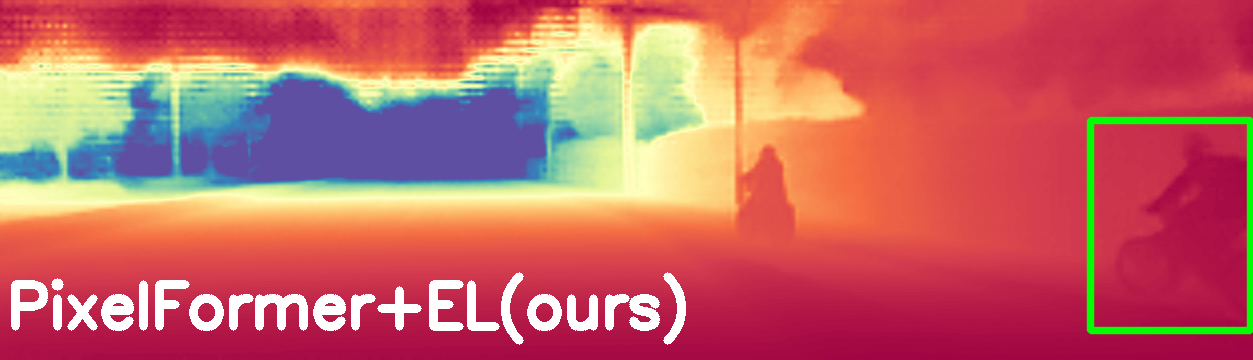}
   \includegraphics[width=0.3\linewidth]{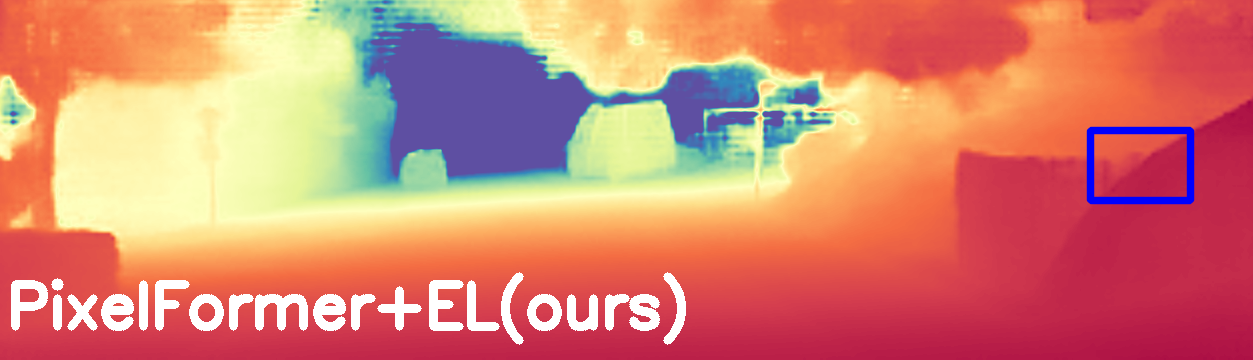}
   \includegraphics[width=0.3\linewidth]{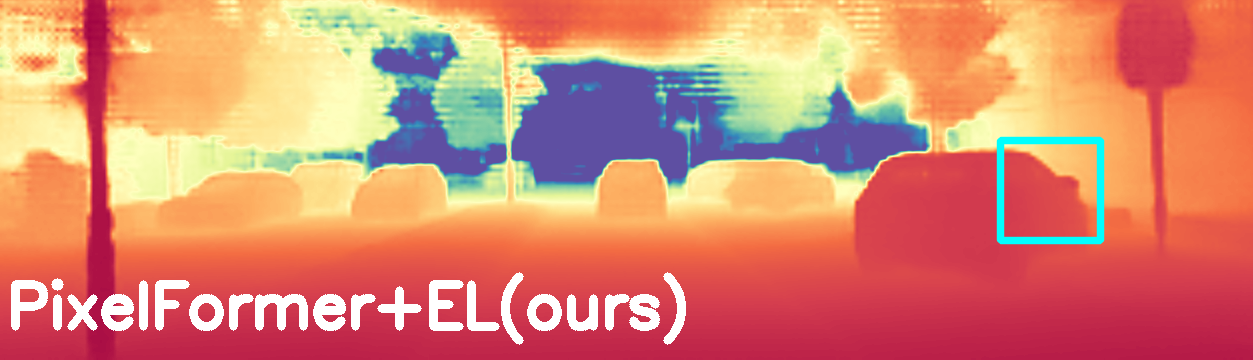} \\
   \includegraphics[width=0.3\linewidth]{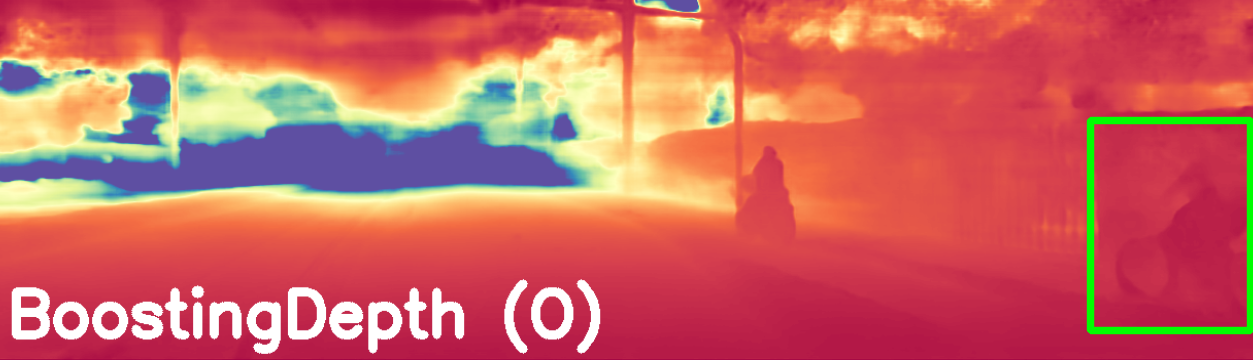}
   \includegraphics[width=0.3\linewidth]{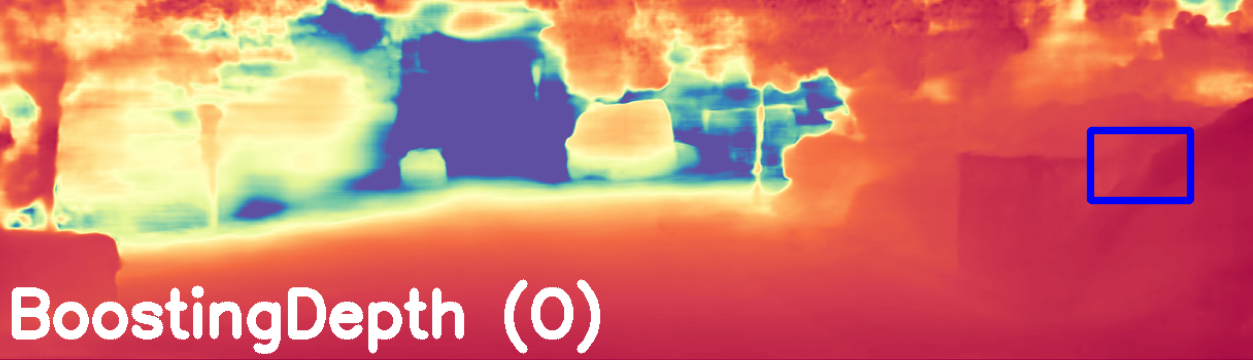}
   \includegraphics[width=0.3\linewidth]{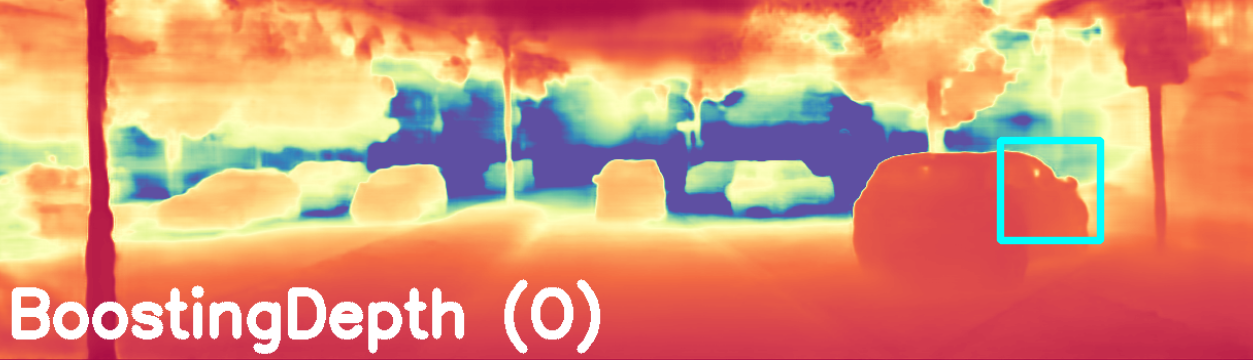} \\
   \includegraphics[width=0.3\linewidth]{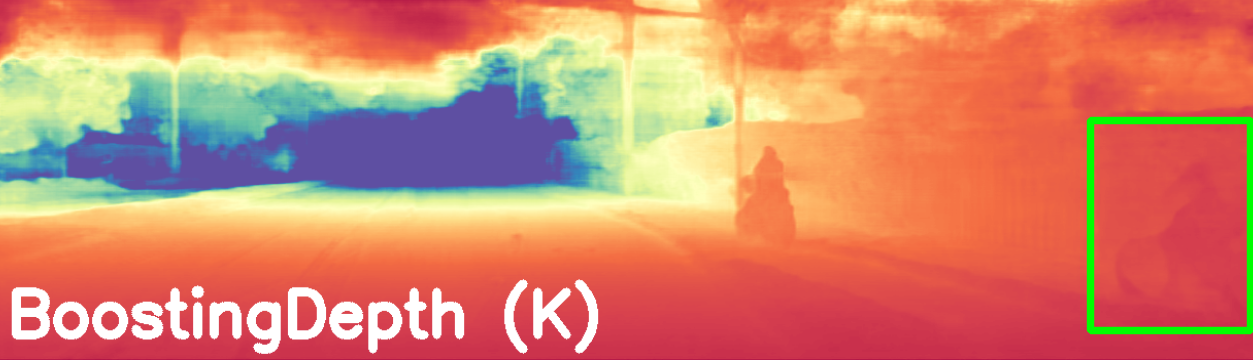}
   \includegraphics[width=0.3\linewidth]{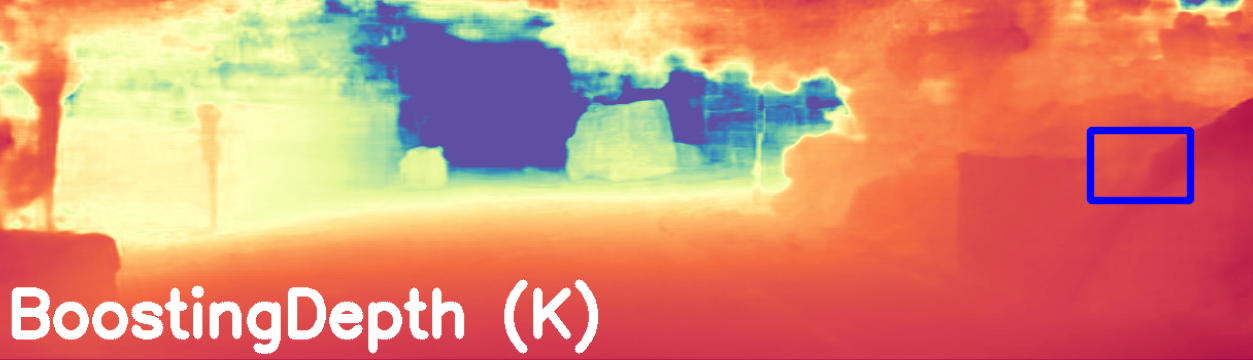}
   \includegraphics[width=0.3\linewidth]{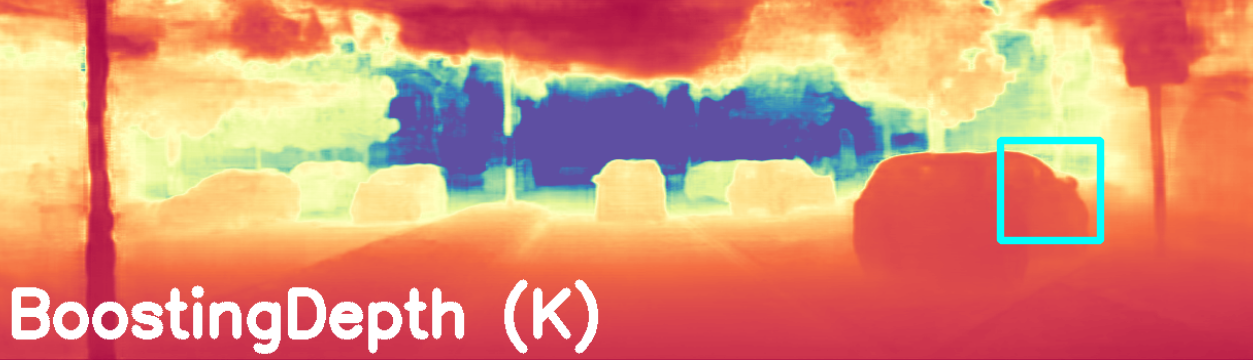} \\
   \includegraphics[width=0.3\linewidth]{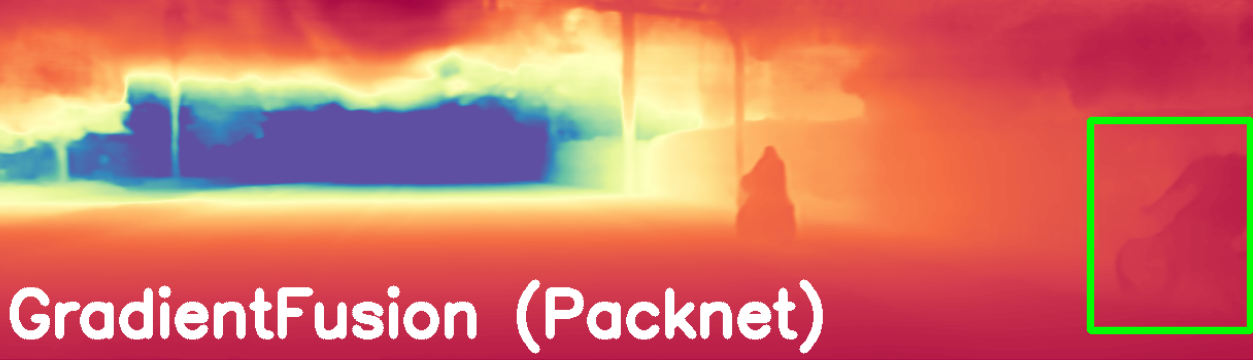}
   \includegraphics[width=0.3\linewidth]{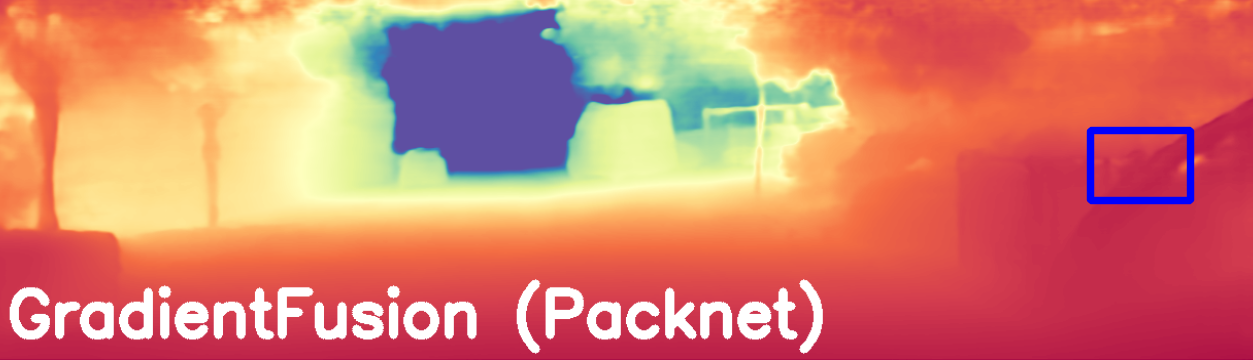}
   \includegraphics[width=0.3\linewidth]{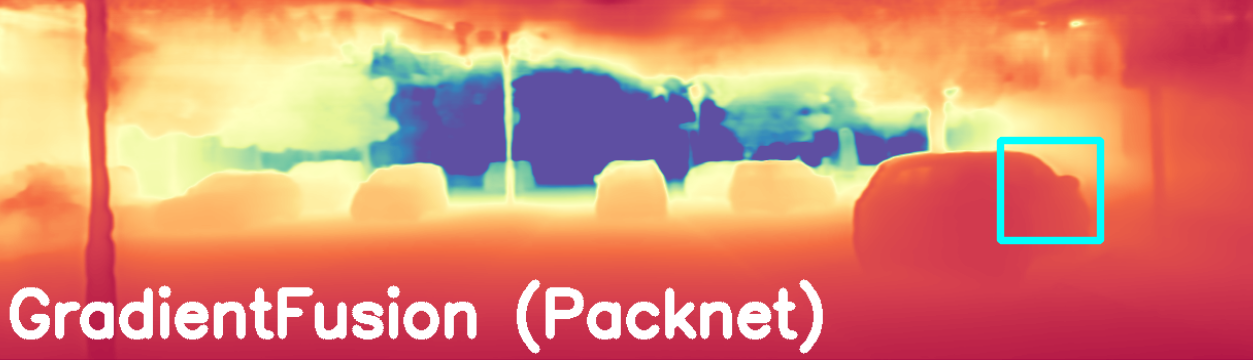} \\
   \includegraphics[width=0.3\linewidth]{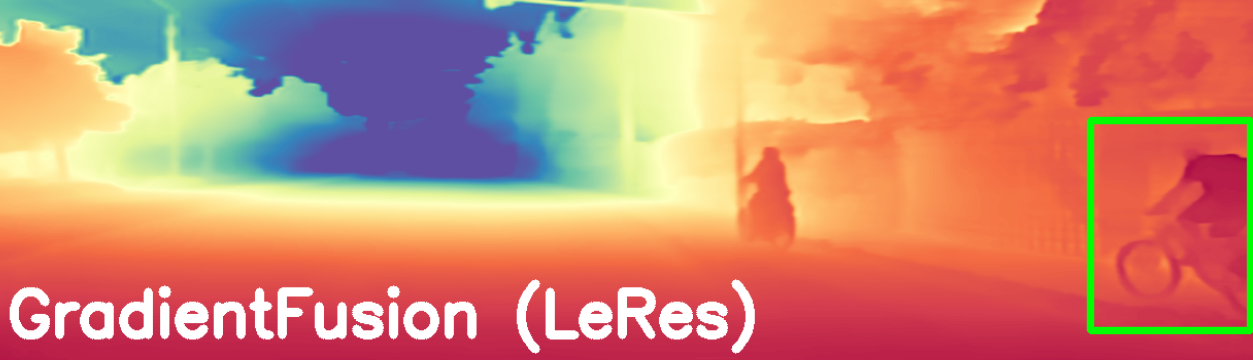}
   \includegraphics[width=0.3\linewidth]{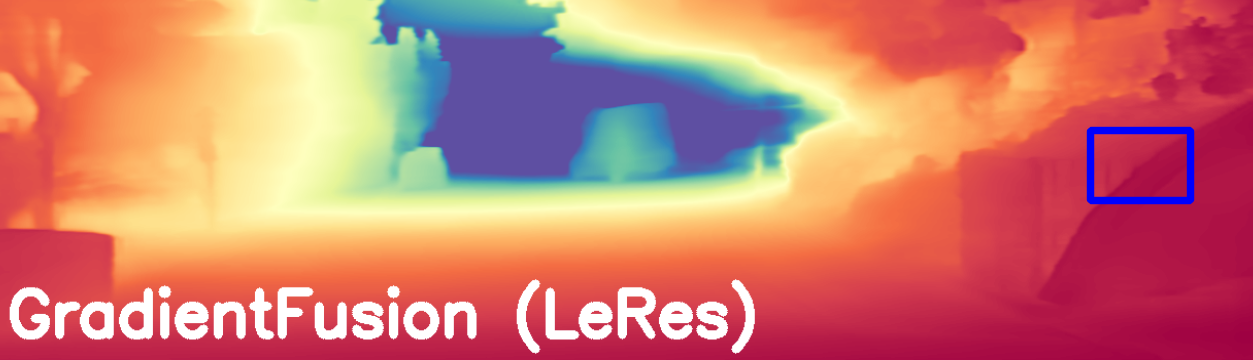}
   \includegraphics[width=0.3\linewidth]{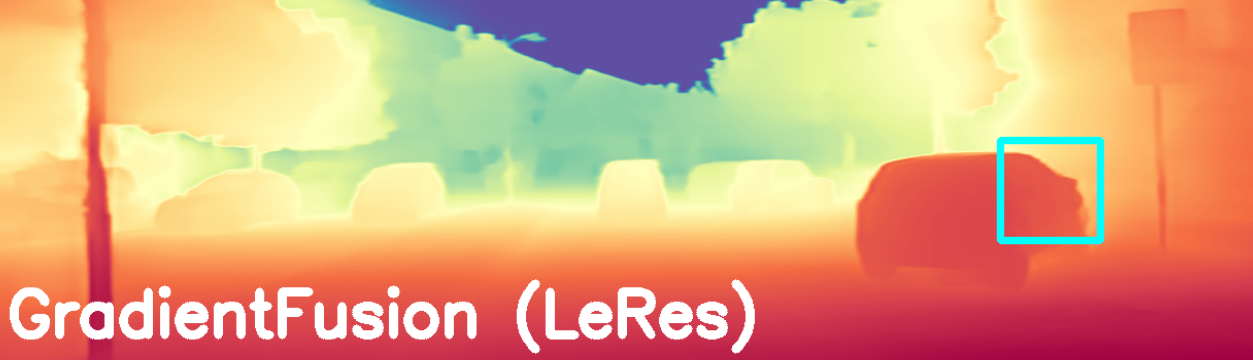} \\
   \includegraphics[width=0.3\linewidth]{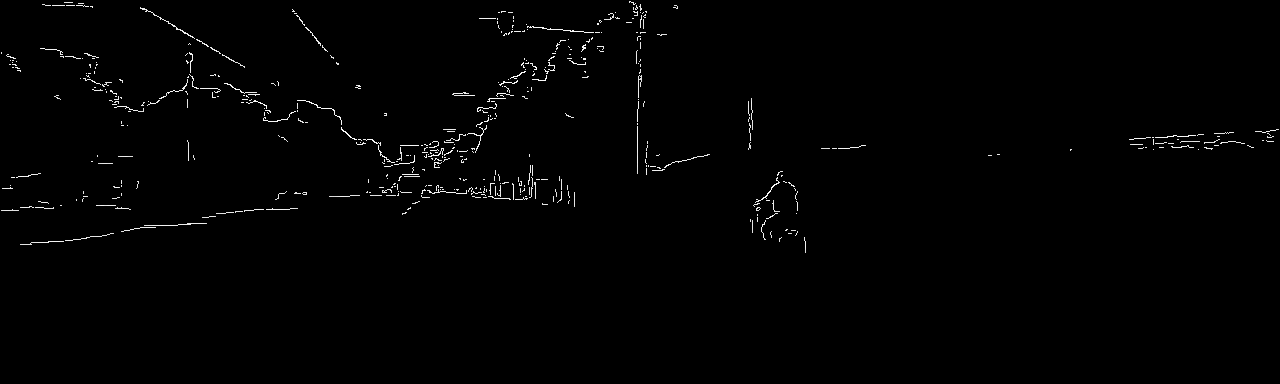}
   \includegraphics[width=0.3\linewidth]{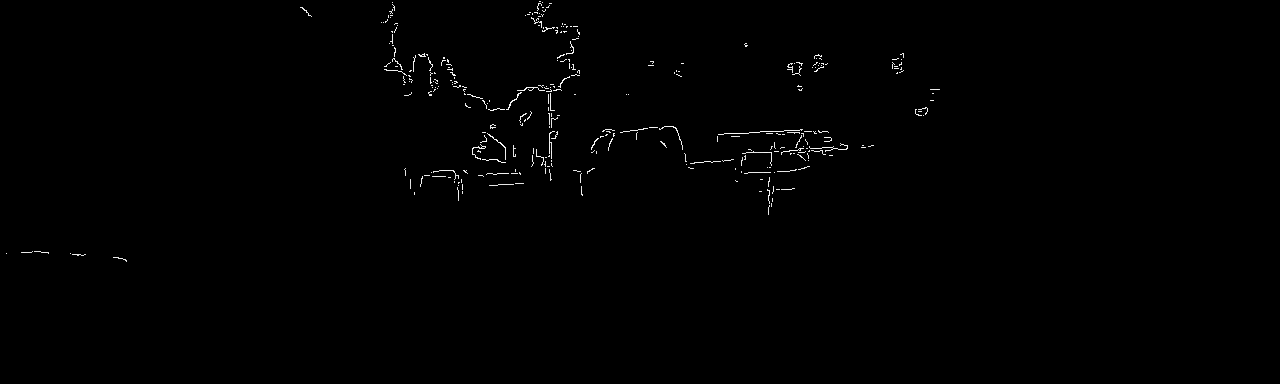}
   \includegraphics[width=0.3\linewidth]{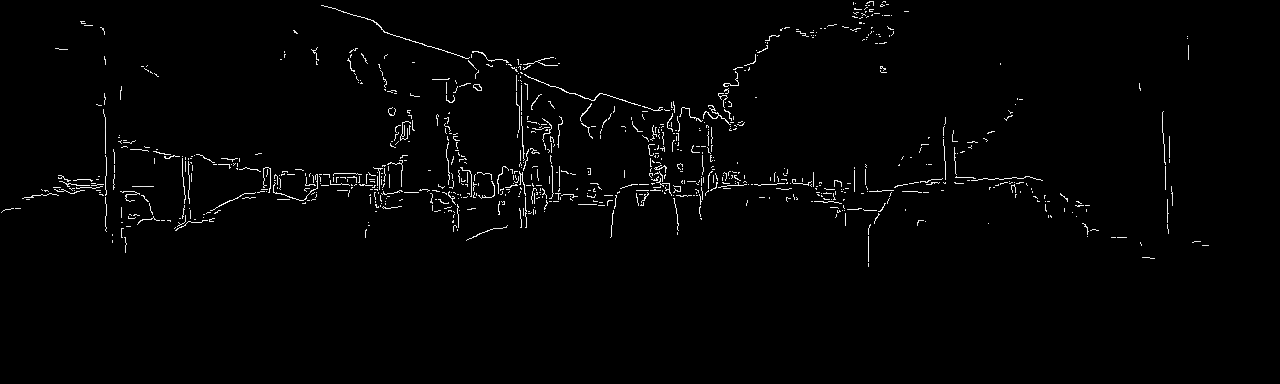}
\end{tabular}
\caption{Full Images (zoom-ins in Figure~\ref{fig:qualitive_kitti_zoom}) of depth predictions of Packnet-SAN, AdaBins, PixelFormer (both baseline and ours) and BoostingDepth and GradientFusion with the original depth merger and the version we trained on KITTI. The last row depicts the result of the DEE network on those images, which are taken from the Eigen KITTI testset.}
\label{fig:qualitive_kitti}
\end{figure*}

\begin{figure*}
    \centering
    \begin{minipage}{.28\linewidth}
            \begin{subfigure}[t]{\linewidth}
            		~~~~~~~~~~~~~~~~~~~~~~ RGB \\
                \includegraphics[width=1.00\linewidth]{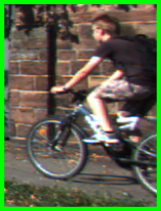}
            \end{subfigure}
        \end{minipage}
    \begin{minipage}{.7\linewidth}
        \begin{subfigure}[t]{\linewidth}
        		~~~~~~~~Packnet~~~~~~~~~~~~ AdaBins ~~~~~~~~ PixelFormer ~~~~~~~~~ BD (O) ~~~~~~~ GF (Packnet) \\
            \includegraphics[width=0.19\linewidth]{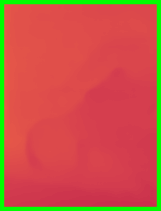}
            \includegraphics[width=0.19\linewidth]{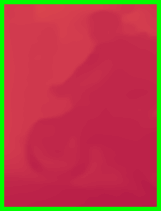}
            \includegraphics[width=0.19\linewidth]{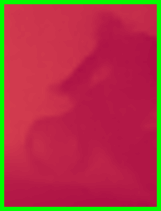}
            \includegraphics[width=0.19\linewidth]{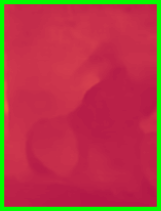}
            \includegraphics[width=0.19\linewidth]{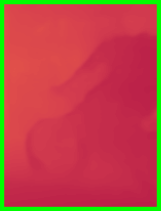}
        \end{subfigure} \\
        \begin{subfigure}[b]{\linewidth}
        ~~~ Packnet+EL ~~~~~ AdaBins+EL ~ PixelFormer+EL ~~~~~ BD (K) ~~~~~~~~~ GF (LeRes) \\
            \includegraphics[width=0.19\linewidth]{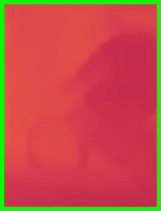}
            \includegraphics[width=0.19\linewidth]{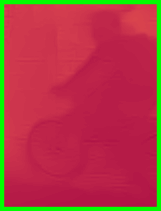}
            \includegraphics[width=0.19\linewidth]{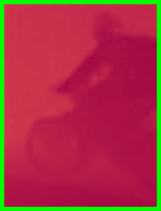}
            \includegraphics[width=0.19\linewidth]{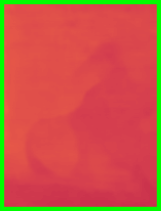}
            \includegraphics[width=0.19\linewidth]{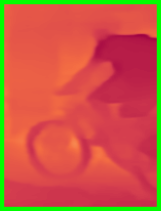}
        \end{subfigure} 
    \end{minipage} \\
    
        \begin{minipage}{.28\linewidth}
            \begin{subfigure}[t]{\linewidth}
                ~~~~~~~~~~~~~~~~~~~~~~ RGB \\
                \includegraphics[width=1.00\linewidth]{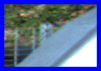}
            \end{subfigure}
        \end{minipage}
    \begin{minipage}{.69\linewidth}
        \begin{subfigure}[t]{\linewidth}
            ~~~~~~~~Packnet~~~~~~~~~~~~ AdaBins ~~~~~~~~ PixelFormer ~~~~~~~~~ BD (O) ~~~~~~~ GF (Packnet) \\
            \includegraphics[width=0.19\linewidth]{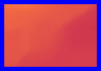}
            \includegraphics[width=0.19\linewidth]{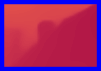}
            \includegraphics[width=0.19\linewidth]{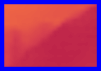}
            \includegraphics[width=0.19\linewidth]{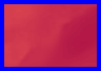}
            \includegraphics[width=0.19\linewidth]{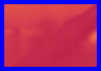}
        \end{subfigure} \\
        \begin{subfigure}[b]{\linewidth}
        ~~~ Packnet+EL ~~~~~ AdaBins+EL ~ PixelFormer+EL ~~~~~ BD (K) ~~~~~~~~~ GF (LeRes) \\
            \includegraphics[width=0.19\linewidth]{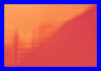}
            \includegraphics[width=0.19\linewidth]{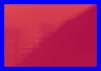}
            \includegraphics[width=0.19\linewidth]{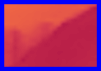}
            \includegraphics[width=0.19\linewidth]{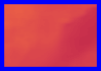}
            \includegraphics[width=0.19\linewidth]{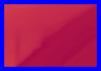}
        \end{subfigure} 
    \end{minipage} \\
    
            \begin{minipage}{.28\linewidth}
            \begin{subfigure}[t]{\linewidth}
                ~~~~~~~~~~~~~~~~~~~~~~ RGB \\
                \includegraphics[width=1.00\linewidth]{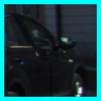}
            \end{subfigure}
        \end{minipage}
    \begin{minipage}{.695\linewidth}
        \begin{subfigure}[t]{\linewidth}
            ~~~~~~~~Packnet~~~~~~~~~~~~ AdaBins ~~~~~~~~ PixelFormer ~~~~~~~~~ BD (O) ~~~~~~~ GF (Packnet) \\
            \includegraphics[width=0.19\linewidth]{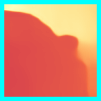}
            \includegraphics[width=0.19\linewidth]{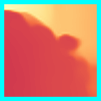}
            \includegraphics[width=0.19\linewidth]{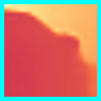}
            \includegraphics[width=0.19\linewidth]{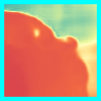}
            \includegraphics[width=0.19\linewidth]{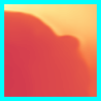}
        \end{subfigure} \\
        \begin{subfigure}[b]{\linewidth}
        ~~~ Packnet+EL ~~~~~ AdaBins+EL ~ PixelFormer+EL ~~~~~ BD (K) ~~~~~~~~~ GF (LeRes) \\
            \includegraphics[width=0.19\linewidth]{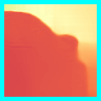}
            \includegraphics[width=0.19\linewidth]{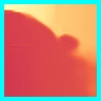}
            \includegraphics[width=0.19\linewidth]{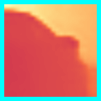}
            \includegraphics[width=0.19\linewidth]{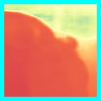}
            \includegraphics[width=0.19\linewidth]{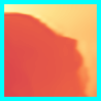}
        \end{subfigure} 
    \end{minipage}
    
    \caption{Zoom-ins of the depth predictions from Fig.~\ref{fig:qualitive_kitti}. The bottom three zoom-in rows, from top to bottom, correspond to the three columns on top, from left to right. BD and GF corresponds to BoostingDepth and GradientFusion, respectively.}
    \label{fig:qualitive_kitti_zoom}
\end{figure*}

\begin{figure*}[t]
\centering
\begin{tabular}{ccc}
~~~~~~~~~~~~~~~~~~~~~~~~~~~~~~~~~~~~~~~~~~~~~~~~~~~~~~~~~~~~~~~~~~~~~~~~~~~~~~~~~~~ RGB ~~~~~~~~~~~~~~~~~~~~~ Baseline ~~~~~~~~~~~~~~~~~~~~~ Ours \\
   \includegraphics[width=0.4268\linewidth]{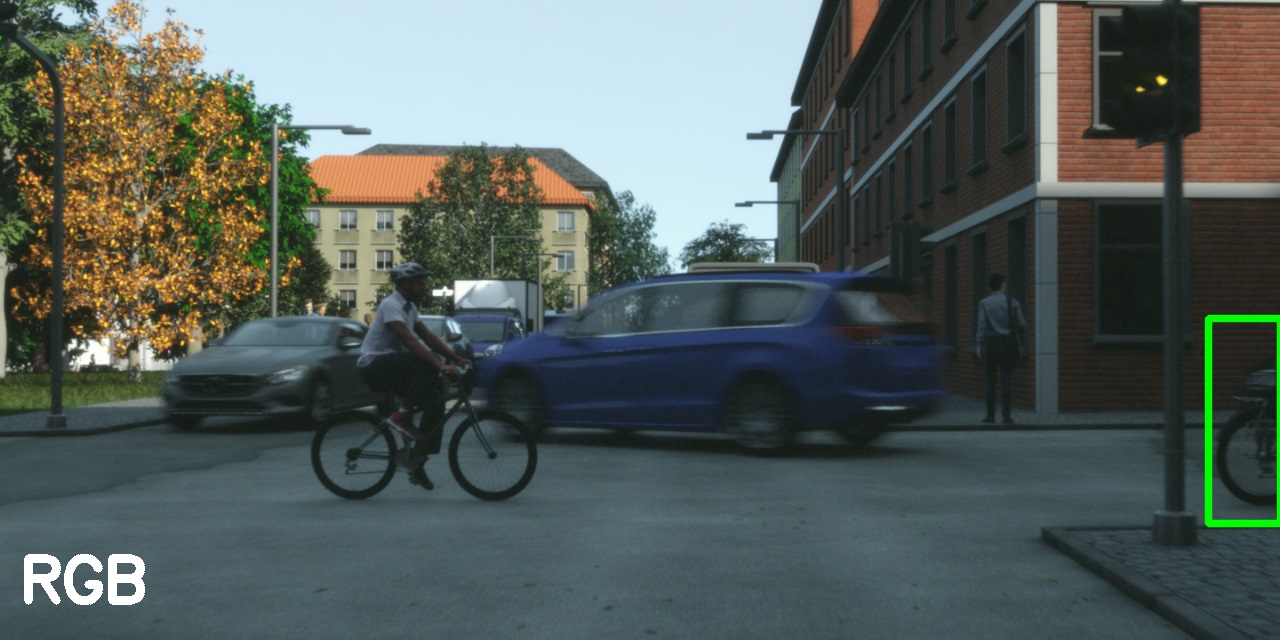} 
   \includegraphics[height=3.69cm,width=0.164\linewidth]{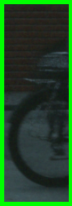} 
   \includegraphics[height=3.69cm,width=0.164\linewidth]{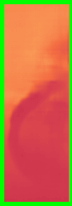} 
   \includegraphics[height=3.69cm,width=0.164\linewidth]{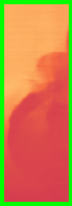} \\
   \includegraphics[width=0.465\linewidth]{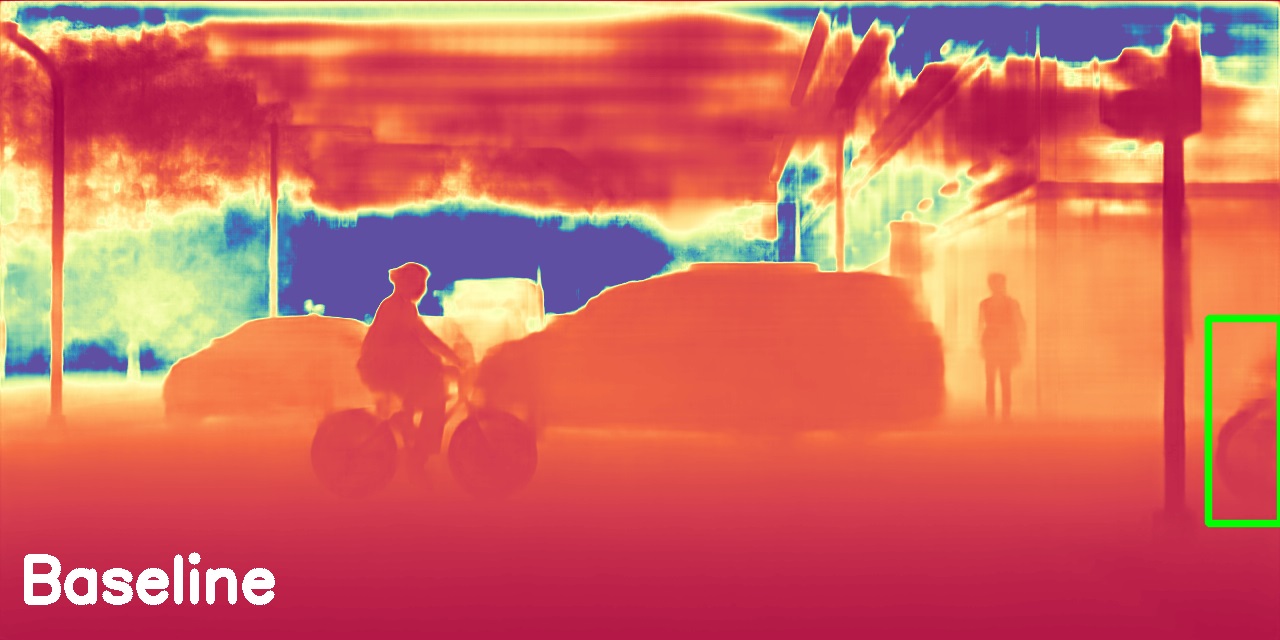} 
   \includegraphics[width=0.465\linewidth]{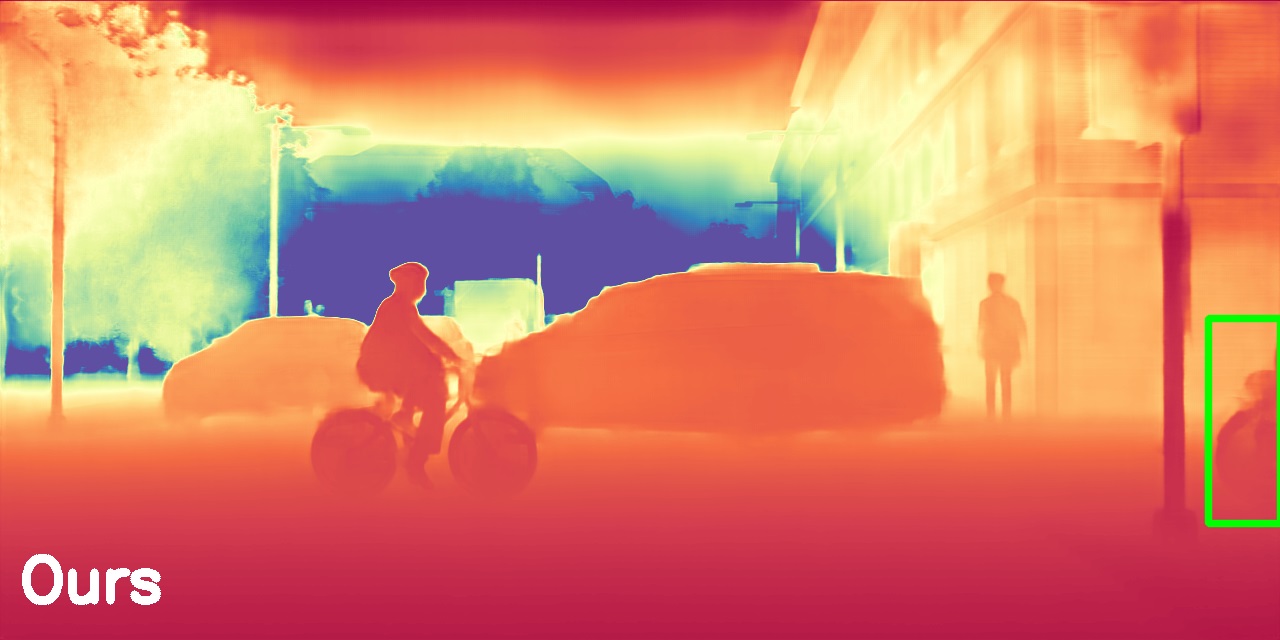} \\
   \includegraphics[width=0.465\linewidth]{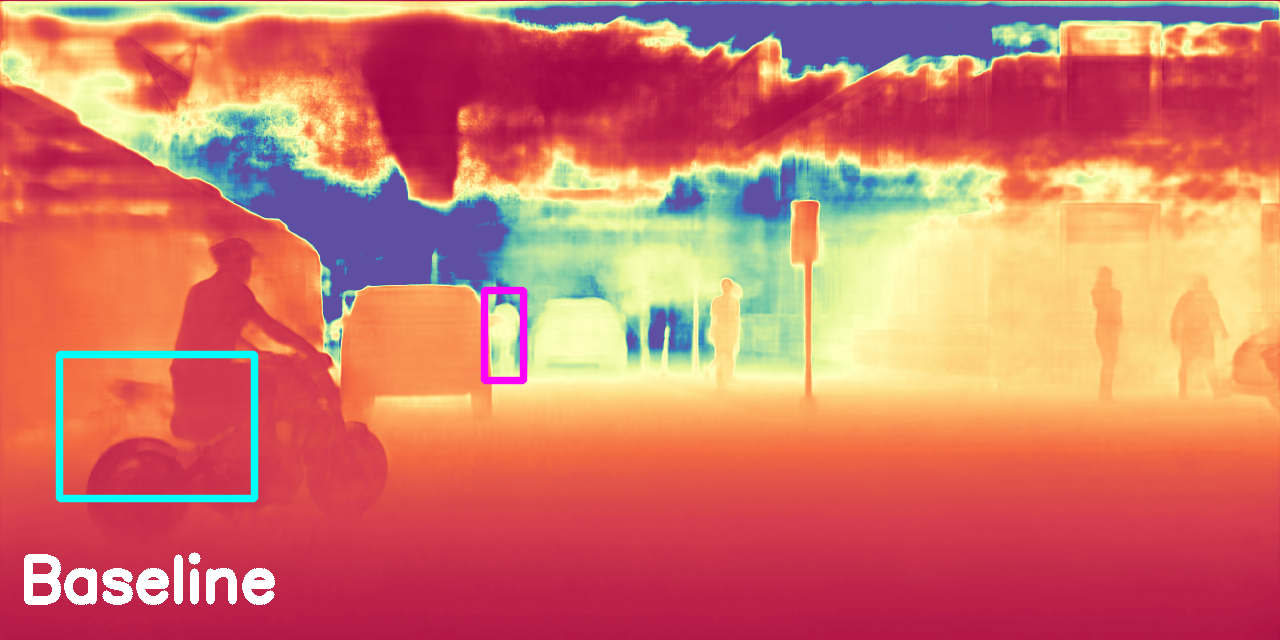} 
   \includegraphics[width=0.465\linewidth]{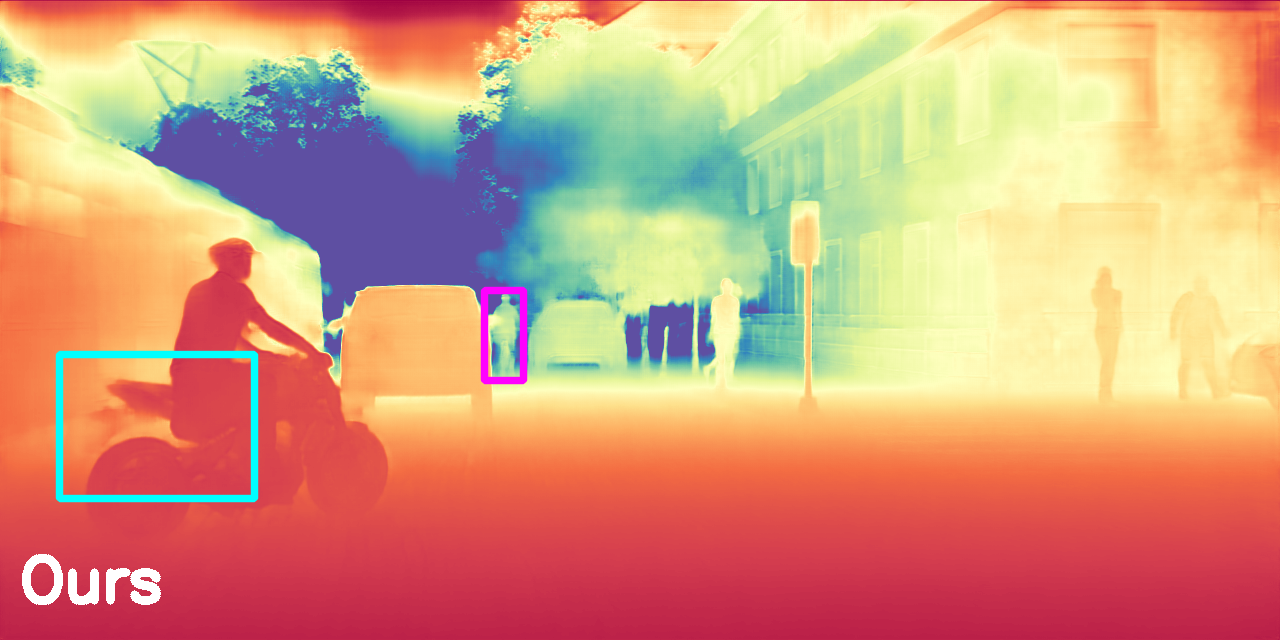} \\
   \includegraphics[width=0.332\linewidth]{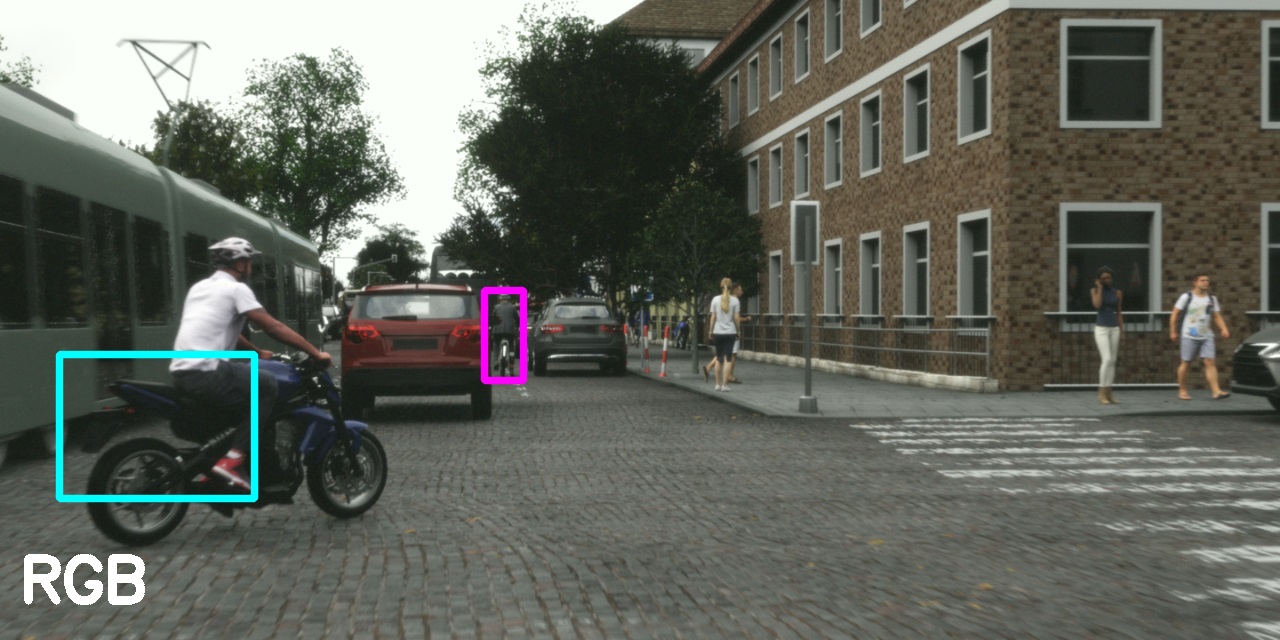} 
   \includegraphics[height=2.92cm,width=0.13\linewidth]{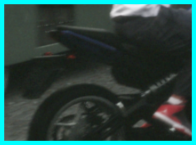} 
   \includegraphics[height=2.92cm,width=0.13\linewidth]{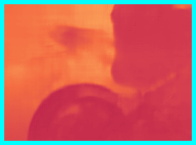} 
   \includegraphics[height=2.92cm,width=0.13\linewidth]{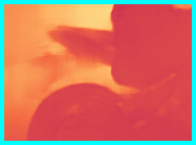}
   \includegraphics[height=2.92cm,width=0.06\linewidth]{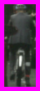} 
   \includegraphics[height=2.92cm,width=0.06\linewidth]{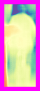} 
   \includegraphics[height=2.92cm,width=0.06\linewidth]{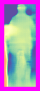} \\
~~~~~~~~~~~~~~~~~~~~~~~~~~~~~~~~~~~~~~~~~~~~~~~~~~~~~~~~~~~~~~~~~~~~~~~~ RGB ~~~~~~~~~~~~~~~ Baseline ~~~~~~~~~~~~~~ Ours ~~~~~~~~~ RGB ~ Baseline ~ Ours
\end{tabular}
\caption{Examples of depth predictions of Packnet-SAN and Packnet-SAN + EL (ours) of images from the Synscapes dataset.}
\label{fig:qualitive_synscapes}
\end{figure*}

\subsection{Comparison to BoostingDepth \cite{miangoleh2021boosting} details}
\label{sec:miangoleh}
To adjust BoostingDepth to Packnet-SAN, we computed Packnet-SAN's receptive field both theoretically and empirically and obtained a recpetive field of 1028 pixels. We utilize it as required by their method. To train the depth merger on KITTI data (called 'Packnet-SAN + BoostingDepth (K) in this paper), we generate training data as explained in BoostingDepth's Github repository, where the low-resolution patches are taken to be in the Receptive Field (RF) size of Packnet-SAN ($1028^2$) and the high-resolution paches are taken to be the entire image ($1280^2$). Note that the original weights of the depth merger were trained with the IBims-1 \cite{koch2018evaluation} and the Middlebury \cite{scharstein2014high} datasets, and used the depth maps of e.g., MIDAS. In this case the ratio between the low-resolution patches and the high-resolution patches, $384^2$ and $672^2$ is much larger than in our case, which may contribute to the difference in performance depicted between the depth mergers. This difference, which is an unchangeable constant, is a limitation of BoostingDepth which relies on the RF of the MDE and the size of the image. Furthermore, we hypothesize that the KITTI dataset does not contain sufficient depth information near depth edges (see Fig.~3 in main paper) due to the LIDAR sparsity, which results in the high resolution depth predictions of Packnet-SAN to have inaccurate edges, differently from the original depth merger trained on dense datasets. The qualitative results are presented in Fig.~\ref{fig:qualitive_kitti}.

\subsection{Additional Details}

\subsubsection{KDE and DDE Datasets Annotation Details} 
\begin{figure*}[t]
\centering
\begin{tabular}{ccc}
   \includegraphics[width=0.32\linewidth]{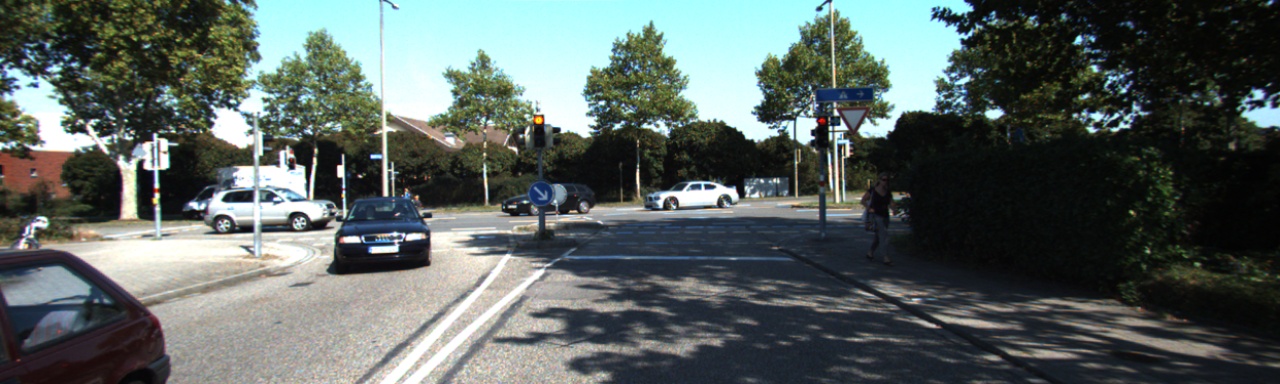} 
   \includegraphics[width=0.32\linewidth]{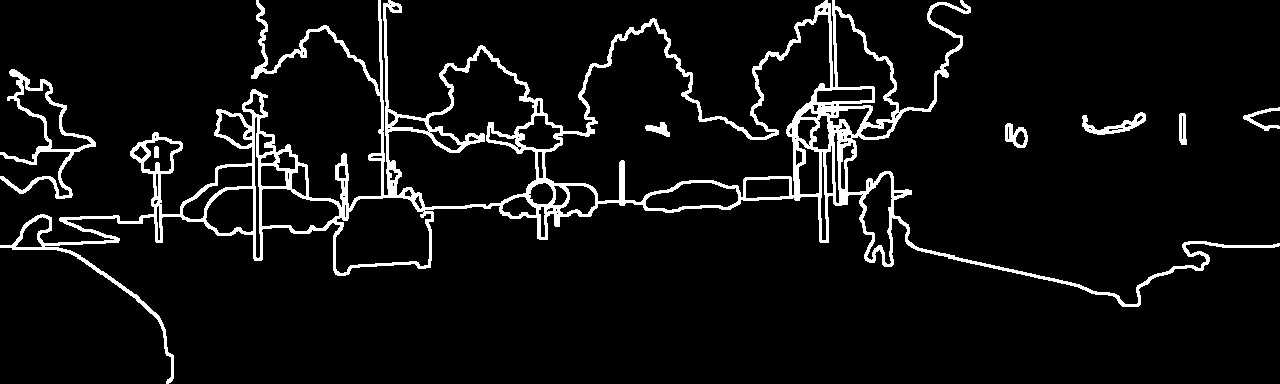} 
   \includegraphics[width=0.32\linewidth]{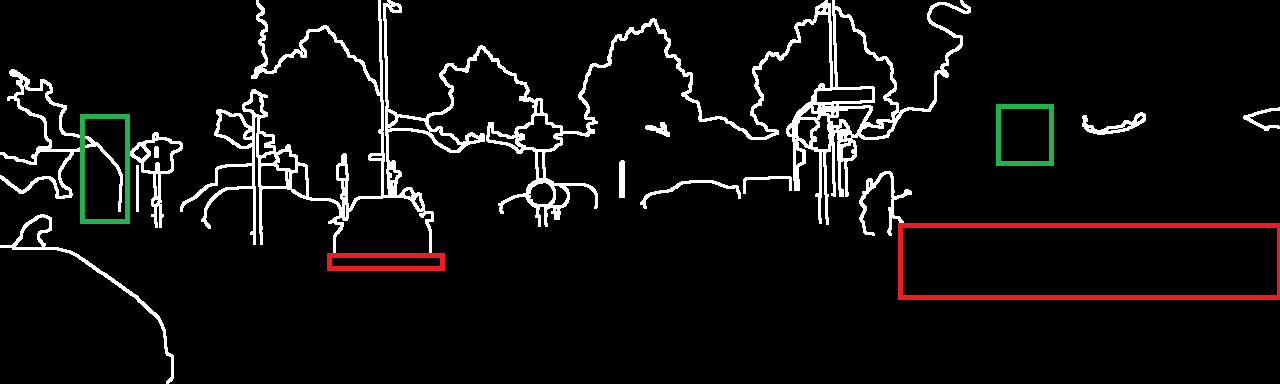} \\ 
   \includegraphics[width=0.32\linewidth]{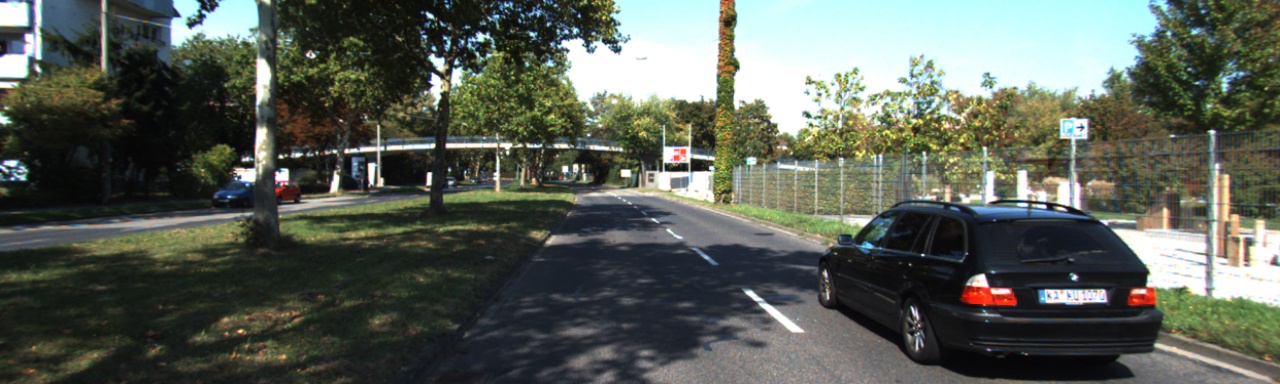} 
   \includegraphics[width=0.32\linewidth]{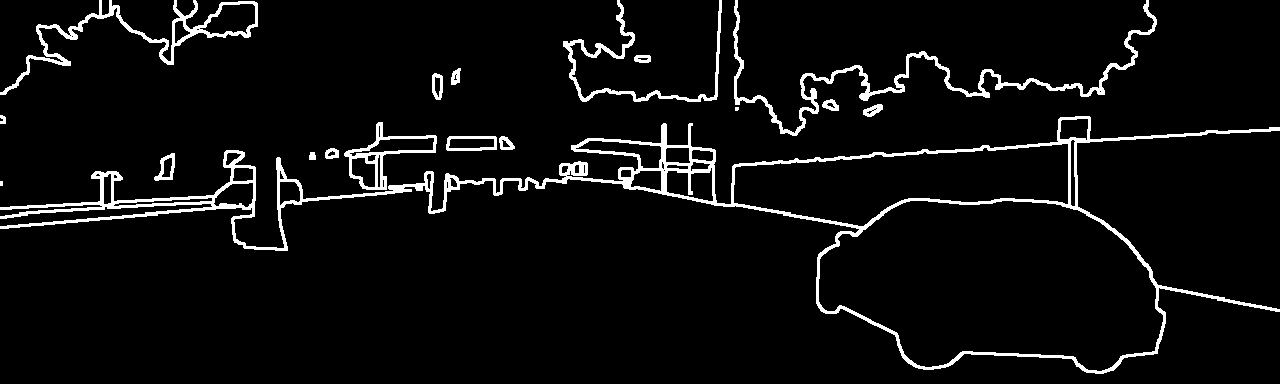} 
   \includegraphics[width=0.32\linewidth]{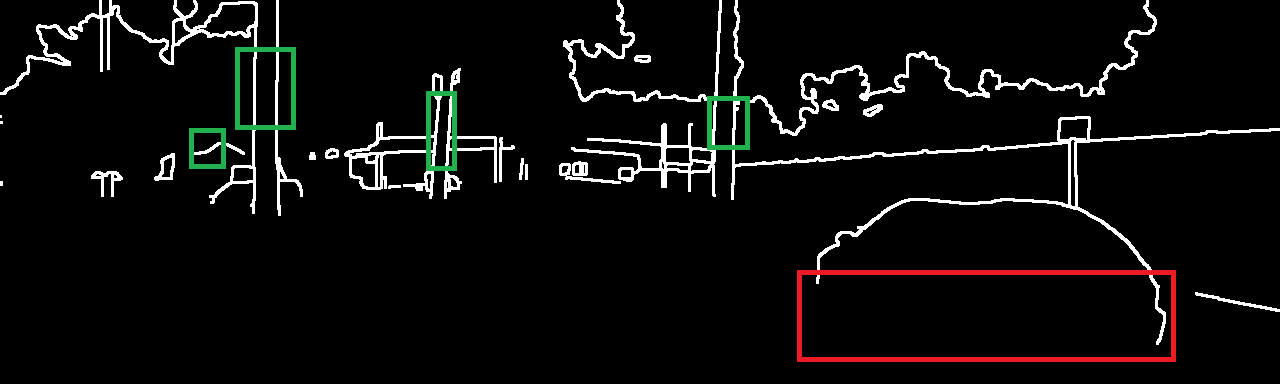}
\end{tabular}
\caption{Examples of the annotation process (the edges are dilated for visualization purposes). (a) The RGB image. (b) The edges of the instance segmentation GT. (c) The depth edges GT, where green and red rectangles exemplifies edge addition and deletion, respectively.}
\label{fig:annotation_examples}
\end{figure*}

As discussed in the paper, we manually annotate 102 images of KITTI (i.e., KDE dataset) and 50 images of DDAD (i.e., DDE dataset). To ease the annotation process, we start from edge maps of panoptic (semantic + instance) segmentation of those images, which yields an edge map which contains most of the depth edges in the scene (see Fig.~\ref{fig:annotation_examples}), but typically has the following shortcomings: (i) Some classes, e.g., building, do not have instance segmentation so potential depth edges between different instances do not exist. (ii) Some edges between different classes, e.g., (bottom of) car and road or road and sidewalk, do not necessarily reflect discontinuities in depth. In the first case, we add the relevant depth edges (see green examples in Fig.~\ref{fig:annotation_examples}), and in the second case we remove the relevant edges (see red examples in Fig.~\ref{fig:annotation_examples}). 

The annotators' guideline for adding or removing edges from the initial edge map from the panoptic segmentation is 4 meters. That is, a depth edge should be annotated if between the two sides of the depth edge there exists a depth discontinuity of at least 4 meters (this is estimated by the annotator). We also note that on average the number of annotated depth edges occupy $\sim2\%$ of the image ($\sim10K$ pixels).

\subsubsection{RGB and RGB+LIDAR in the DEE network}
As we argue in the main paper that the performance of the DEE network is significantly better when the input is RGB and LIDAR, in constrast to RGB only. In Fig.~\ref{fig:prec_recall_rgb_rgb_lidar} we present the depth edges precision-recall graph for the DEE network on the KITTI-DE dataset, where the performance gap in favor of the RGB+LIDAR input is clearly present. Moreover, examples of the output of the DEE network is presented in the bottom row of Fig.~\ref{fig:qualitive_kitti}.

\subsubsection{Depth Edge Loss}
As we argue in the main paper (L370), using the standard spatial image gradient often yield undesired artifacts. See the stripes over the car silhouette for a depiction of this phenomenon in Fig.~\ref{fig:depth_edge_loss}. The quantitative performance on the KITTI-DE dataset is also somewhat lower: ARE: 4.03\% and AUC (edges): 59.3\% (48.13\%). 

\subsubsection{Implementation Details}
The DEE network follows the architecture of the U-net like PackNet-SAN \cite{guizilini2021sparse} (Fig.~\ref{fig:dee}) which was used as an MDE network. One of the beneficial properties of PackNet-SAN is its sparse encoder for the sparse LIDAR signal, which uses only sparse layers (e.g., sparse convolutions), which is suitable for our case of depth edges estimation. The DEE network is trained for five epochs on the GTA-PreSIL dataset \cite{hurl2019precise}. For training the MDE using our edge loss, we set $\alpha=0.1$ and $\alpha=1.0$ for Packnet-SAN and AdaBins, respectively. Also, we note that AdaBins is trained only with the largest scale for both baseline and our method as in the original training.

\subsection{Additional Examples for KITTI and DDAD}
We present additional qualitative results for KITTI in Fig.~\ref{fig:qualitive_kitti} and for DDAD in Fig.~\ref{fig:qualitive_ddad_1} and Fig.~\ref{fig:qualitive_ddad_2}.


{\small
\bibliographystyle{ieee_fullname}
\bibliography{egbib}
}

\end{document}